\documentclass[journal]{IEEEtran}
\usepackage{cite}
\usepackage{multicol}
\usepackage[hidelinks]{hyperref}
\usepackage{bm}
\usepackage{rotating}
\usepackage{url}
\usepackage{graphicx} 
\usepackage[mathscr]{euscript}  
\usepackage{amsmath} 
\usepackage{amssymb}  
\usepackage{flushend}
\usepackage{multirow}\usepackage{multirow}

\usepackage{siunitx}
\usepackage{xcolor}
\usepackage[export]{adjustbox}
\usepackage{array}
\usepackage{booktabs}
\usepackage[ruled,vlined]{algorithm2e}  

\usepackage[inline]{enumitem}
\usepackage[hang,flushmargin]{footmisc}
\usepackage{microtype}
\usepackage{threeparttable}
\usepackage{pifont}
\usepackage{mathtools}

\usepackage{dsfont}

\usepackage{verbatim}
\usepackage{xcolor}
\definecolor{dark-green}{RGB}{12,80,12}

\DeclareMathOperator{\E}{\mathbb{E}}
\newcommand{\expnumber}[2]{{#1}\mathrm{e}{#2}}

\newcommand{\ours}{N$^2$M$^2$}
\newcommand{\ourslong}{Neural Navigation for Mobile Manipulation}

\DeclareSIUnit{\degree}{deg}
\DeclareSIUnit{\rad}{rad}
\pdfoutput=1

\newcommand{\secref}[1]{Sec.~\ref{#1}}
\renewcommand{\eqref}[1]{Eq.~(\ref{#1})}
\newcommand{\figref}[1]{Fig.~\ref{#1}}  
\newcommand{\tabref}[1]{Tab.~\ref{#1}}

\newcommand{\myworries}[1]{\textcolor{black}{#1}}
\newcommand\myworriestwo[1]{\textcolor{black}{#1}}

\newcolumntype{P}[1]{>{\centering\arraybackslash}p{#1}}

\usepackage{tabularx}
\newcolumntype{Y}{>{\centering\arraybackslash}X}
\newcolumntype{Z}{>{\raggedleft\arraybackslash}X}

\begin{document}

\title{\ours: Learning Navigation for Arbitrary\\Mobile Manipulation Motions in Unseen and\\Dynamic Environments}
\author{Daniel Honerkamp,
        Tim Welschehold, 
        and~Abhinav Valada%
\thanks{\hspace{8pt}© 2023 IEEE. Personal use of this material is permitted.  Permission from IEEE must be obtained for all other uses, in any current or future media, including reprinting/republishing this material for advertising or promotional purposes, creating new collective works, for resale or redistribution to servers or lists, or reuse of any copyrighted component of this work in other works.}
\thanks{\hspace{8pt}Manuscript received 15 December 2022; revised 19 April 2023; accepted May 2023. This work was supported by the European Union’s Horizon 2020 Research and Innovation Program under Grant 871449-OpenDR and in part by Toyota Motor Europe with an HSR robot for experimental evaluation. This paper was recommended for publication by Associate Editor Y. Bekiroglu and Editor F. Chaumette upon evaluation of the reviewers’ comments.}
\thanks{\hspace{8pt}All authors are with the Department of Computer Science, University of Freiburg, 79110 Germany (e-mail: honerkamp@cs.uni-freiburg.de, twelsche@cs.uni-freiburg.de, valada@cs.uni-freiburg.de).}
\thanks{\hspace{8pt}Digital Object Identifier 10.1109/TRO.2023.3284346
}}

\markboth{IEEE TRANSACTIONS ON ROBOTICS}%
{Honerkamp \MakeLowercase{\textit{et al.}}: \ours: Learning Navigation for Arbitrary\\Mobile Manipulation Motions in Unseen and\\Dynamic Environments}
%



\maketitle

\begin{abstract}
Despite its importance in both industrial and service robotics, mobile manipulation remains a significant challenge as it requires seamless integration of end-effector trajectory generation with navigation skills as well as reasoning over long-horizons. Existing methods struggle to control the large configuration space and to navigate dynamic and unknown environments. 
In previous work, we proposed to decompose mobile manipulation tasks into a simplified motion generator for the end-effector in task space and a trained reinforcement learning agent for the mobile base to account for the kinematic feasibility of the motion. In this work, we introduce \ourslong ~(\ours) which extends this decomposition to complex obstacle environments, \myworries{extends the agent's control to the torso joint and the norm of the end-effector motion velocities, \myworriestwo{uses a more general reward function}} and  \myworries{thereby enables robots} to tackle a \myworries{much} broad\myworries{er} range of tasks in real-world settings. 
The resulting approach can perform unseen, long-horizon tasks in unexplored environments while instantly reacting to dynamic obstacles and environmental changes. At the same time, it provides a simple way to define new mobile manipulation tasks. We demonstrate the capabilities of our proposed approach in extensive simulation and real-world experiments on multiple kinematically diverse mobile manipulators. Code and videos are publicly available at \url{http://mobile-rl.cs.uni-freiburg.de}.
\end{abstract}

%
\IEEEpeerreviewmaketitle


\section{Introduction}\label{sec:intro}
\IEEEPARstart{W}{hile} recent progress in control and perception has propelled the capabilities of robotic platforms to autonomously operate in unknown and unstructured environments~\cite{blomqvist2020go, hurtado2021learning, sirohi2021efficientlps, valverde2021there}, this has largely focused on pure navigation tasks~\cite{duan2022survey, younes2021catch}. In this work, we focus on autonomous mobile manipulation which combines the difficulties of navigating unstructured, human-centered environments with the complexity of jointly controlling the base and arm. Mobile Manipulation is commonly reduced to sequential base navigation followed by static arm manipulation at the goal location.
This simplification is restrictive as many tasks such as door opening require the joint use of the arm and base and is inefficient as it dismisses simultaneous movement and requires frequent repositioning.

\begin{figure}
    \footnotesize
	\centering
	\resizebox{\columnwidth}{!}{%
  		\includegraphics[width=0.32\columnwidth,trim={0.0cm 0.0cm 0.0cm 0.0cm},clip,angle =0]{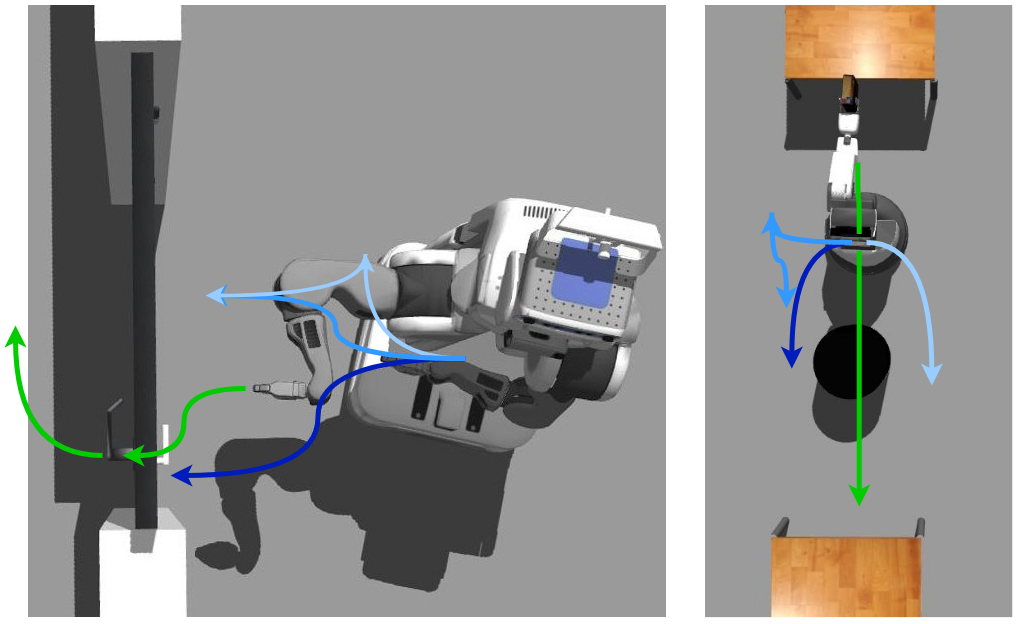}
	}
	\caption{Mobile manipulation tasks in unstructured environments typically require the simultaneous use of the robotic arm and the mobile base. While it is comparably simple to find end-effector motions to complete a task (green), defining base motions (blue) that conform to both the robot's and the environment's constraints is highly challenging. We propose an effective approach that learns feasible base motions for arbitrary end-effector motions. The resulting model is flexible, dynamic and generalizes to unseen motions and tasks.}
  	\label{fig:teaser}
\end{figure}

Mobile manipulation requires a range of capabilities including collision-aware navigation, object interactions, manipulation, exploration of unknown environments, and long-term reasoning.
As a result, approaches from a variety of areas such as planning, optimal control, and learning have been proposed. Inverse reachability map (IRM) approaches seek good base positioning for the robot base given the task constraints~\cite{vahrenkamp2013robot, paus2017combined} but often require expert knowledge and can be overly restrictive.
Planning approaches~\cite{karaman2011sampling, kuffner2000rrt-connect, burget2016bi2rrt} come with asymptotic optimality guarantees but scale unfavorably with the size of the configuration space, can have long planning times, and often require frequent re-planning in dynamic environments. Model predictive control (MPC) formulations explicitly define and optimize over a range of collision and desirability constraints, and recently achieved strong results in mobile manipulation~\cite{pankert2020perceptive, minniti2021model}. However, they are computationally expensive, often do not optimize past a limited horizon, and can struggle when objectives oppose each other. Learning-based methods efficiently learn directly from high-dimensional observations and are well suited to handle unexplored environments~\cite{xia2020relmogen, wong2022error, fabian22exploration}. Nevertheless, they either restrict the action space~\cite{xia2020relmogen, kindle2020wholebody} or rely on expert demonstrations~\cite{wong2022error} to cope with the high-dimensional action space and long-horizon nature of mobile manipulation. Furthermore, the learned behavior is often task-specific, requiring re-training for each novel task. 

In this work, we formulate mobile manipulation as a goal conditional reinforcement learning (RL) problem in which the RL agent observes the end-effector motion and goal and aims to ensure that these motions remain kinematically feasible. \figref{fig:teaser} depicts a high-level overview of our approach. We extend the formulation introduced in~\cite{honerkamp2021learning} to unstructured obstacle environments, increase the agent's freedom to control the \myworriestwo{robots' torso lift joint} and \myworries{the norm of the} velocity of the end-effector motions, \myworries{generalize the reward function, introduce a regularization to the inverse kinematics solver to address configuration jumps and develop a diverse training task}. This provides a very simple and yet effective way to define new tasks, as the RL agent takes care of all the complexities regarding collision-free navigation and kinematic feasibility. The resulting approach, which we term \ours ~(\ourslong), efficiently learns to solve \myworries{the navigation for} long-horizon \myworries{mobile manipulation} tasks, lasting thousands of steps \myworries{or up to five minutes in real-time execution}, generalizes to unseen tasks and environments in a zero-shot manner, and reacts instantaneously to changes in the environment without any planning times. Lastly, the approach is directly applicable to a wide range of kinematics including both holonomic and non-holonomic robotic bases. We show that with appropriate action regularization, our hybrid approach can be trained without a complex simulator and directly transfers to the real world.
We demonstrate these capabilities in both extensive simulation and real-world experiments on multiple mobile manipulators, across a wide range of tasks and environments \myworries{and find that each of the introduced components strongly contributes to the overall large improvements over previous work}.


This paper makes the following main contributions:
\begin{enumerate}
    \item We formulate the fulfillment of kinematic feasibility constraints for mobile manipulation tasks in the presence of obstacles as a reinforcement learning problem.
    \item We propose a reactive approach to learn complementary mobile manipulator base motions for arbitrary end-effector motions on unstructured obstacle maps.
    \item We develop a procedurally generated training task and an approach to generate end-effector motions that can be used across a variety of mobile manipulators with largely varying kinematics and driving models. 
    \item We demonstrate the capabilities of our approach in extensive simulated and real-world experiments on unseen environments, obstacles, and tasks.
    \item We make the code publicly available at \url{http://mobile-rl.cs.uni-freiburg.de}.
\end{enumerate}

The remainder of the paper is organized as follows: \secref{sec:related} discusses existing literature and approaches for mobile manipulation. \secref{sec:approach} describes the technical details of our method, \secref{sec:experiments} then evaluates its capabilities and  compares it to previous approaches. Lastly, \secref{sec:conclusion} discusses limitations and future work and concludes.

\section{Related Work}\label{sec:related}
Mobile manipulation tasks require the composition of a vast range of capabilities spanning perception, control, and exploration. In the following, we discuss previous methods from planning, optimal control, and learning to tackle these challenges.

{\parskip=5pt \noindent\textit{Sequential navigation and manipulation}:
Due to the difficulties of planning in the conjoint space of the mobile manipulator base and arm, many existing approaches restrict themselves to sequential movements of the base followed by static manipulations with the arm. This decomposition has been popular across approaches based on reachability~\cite{paus2017combined}, planning~\cite{diankov2008manipulation, blomqvist2020go, arduengo2021robust}, impedance control~\cite{liu2020garbage}, and reinforcement learning~\cite{xia2020relmogen, jauhri2022robot}.}


{\parskip=5pt \noindent\textit{Planning}:
To ensure kinematic feasibility in mobile manipulation tasks, planning-based approaches plan trajectories for the robot in joint space and as such only explore kinematically feasible paths~\cite{burget2013whole,burget2016bi2rrt}. Sampling-based approaches such as rapidly exploring random trees (RRT) and their variants have been shown to perform well in high-dimensional spaces and come with certain optimality guarantees. However, increasing their configuration spaces and environments can result in long planning times or far from optimal solutions. Operation in unobserved or dynamic environments can trigger costly re-planning if the environment changes~\cite{arslan2013rrtsharp, otte2015rrtx}. While certain constraints such as fixed orientations or task space regions~\cite{berenson2011task} can be incorporated well to plan motions such as opening a door~\cite{arduengo2021robust}, the incorporation of arbitrary (end-effector) constraints is difficult and often requires expert knowledge and task-specific adaptations. In contrast, our approach can near-instantly react to dynamic changes, offers a natural way to incorporate unexplored environments, and can be directly applied to arbitrary end-effector motions.}

{\parskip=5pt \noindent\textit{Inverse Reachability}
maps~\cite{vahrenkamp2013robot} can be used to seek good positioning for the robot base given the task constraints~\cite{paus2017combined}, subsequently the task is solved stationary. Combinations with planning methods exist~\cite{leidner2014object}, but it remains a hard problem to integrate kinematic feasibility constraints in task space mobile motion planning. Welschehold~\textit{et~al.}~\cite{twelsche18coupling} treat the kinematic feasibility of arbitrary gripper trajectories as an obstacle avoidance problem based on a geometric description of the inverse reachability. They analytically modulate the base velocity such that the base stays within feasible regions and orientations relative to the end-effector.}

{\parskip=5pt \noindent\textit{Optimal Control}
%
approaches have demonstrated promising results on complex manipulation tasks that require conjoint movements of the arm and base. Model predictive control (MPC) based approaches have demonstrated strong performance on whole-body control tasks such as door opening~\cite{sleiman2021aunified, minniti2021model}, obstacle avoidance~\cite{pankert2020perceptive}, and articulated objects~\cite{mittal2021articulated}. Constraints are explicitly incorporated into the objective function and optimized over a (usually fixed) rollout horizon. While efficient implementations can incorporate horizons of up to several seconds in near real-time, these approaches still require significant compute. Moreover, they also require simplified collision objects to achieve this and often do not take into account consequences beyond the rollout horizon. In contrast, our approach learns a value function reflective of the full episode horizon and can perform fast inference with a single forward pass, without reliance on highly optimized implementations. Furthermore, it does not rely on explicit representations of the environment and offers direct extensions to arbitrary input modalities and partially observed environments.}

Haviland~\textit{et~al.}~\cite{haviland2022holistic} propose a reactive controller for both holonomic and non-holonomic bases, however, they do not demonstrate any collision avoidance.
Redundancy resolution methods specify desirable aspects of different solutions through additional constraints or parameters~\cite{ancona2017redundancy}. In contrast, our approach directly learns to resolve redundancies with regard to long-term optimality, removing the need to specify the explicit desirability of different choices.

{\parskip=5pt \noindent\textit{Task and motion planning} (TAMP) combines low-level motion planning and high-level task reasoning and has resulted in approaches that generalize to many robots and environments~\cite{garrett2021integrated, toussaint2018differentiable, garrett2018ffrob}. At the same time, it can be computationally demanding, depends on the specification of symbolic actions, and does not generally incorporate uncertainties or partial observability as it requires full information about the 3D structure of the environment. In contrast, we assume that we can generate the desired end-effector motions, but may act in unexplored or dynamically changing environments.}

{\parskip=5pt \noindent\textit{Obstacle Avoidance}:
%
Learning-based methods have been shown to be effective for learning obstacle avoidance from high-dimensional sensor inputs such as color images~\cite{bansal2020combining, el2013visual}, LiDAR~\cite{kollmitz2020predicting, guldenring2020learning}, depth images~\cite{hoeller2021learning} or a combination thereof~\cite{everett2018motion}. These approaches have demonstrated the ability to avoid dynamic obstacles such as pedestrians~\cite{everett2018motion, guldenring2020learning, patel2021dynamically}. Sensors such as LiDAR and depth sensors have also shown good transfer from the simulation into the real world~\cite{hoeller2021learning, patel2021dynamically}. Several approaches have explored the use of 2D obstacle maps to learn a local planner component to avoid pedestrians based on a 2D LiDAR~\cite{guldenring2020learning}, to predict a collision map from a 2D obstacle map based on a laser scan and binary collision events~\cite{kollmitz2020predicting} \myworries{or to directly map from 2D maps to dense Q-values~\cite{wu2020spatial}. LiDAR and depth modalities are commonly used to construct local or global (occupancy) maps. These are then often transformed into egocentric perspective~\cite{chen2018learning, wu2020spatial, guldenring2020learning} or into different resolutions~\cite{chen2018learning, goel2022predicting, fabian22exploration} before being processed by neural networks}.
Optimal control approaches have successfully used voxel map representations~\cite{oleynikova2017voxblox} and signed distance fields~\cite{han2019fiesta, pankert2020perceptive, mittal2021articulated}.}


{\parskip=5pt\noindent\textit{Reinforcement Learning}:
Recently mobile manipulation has also been tackled as a reinforcement learning task. Relmogen~\cite{xia2020relmogen} learns subgoals for arm and base, but relies on sequential execution of each and pre-specified pushing motions. Kindle~\textit{et~al.}~\cite{kindle2020wholebody} learn to directly control both arm and base, however, they restrict the arm movements to a plane. Jauhri~\textit{et~al.}~\cite{jauhri2022robot} learn a base positioning and a discrete decision on whether to use the arm.
While these approaches learn effective policies for specific tasks, they cannot easily be applied to novel tasks. In contrast, we address the problem of enabling arbitrary end-effector trajectories, providing a straightforward technique to define novel tasks with arbitrary motion constraints in task space and the capability for zero-shot generalization to such unseen tasks at test time.}

Recently several approaches and benchmarks for complex long-horizon tasks have been proposed. Most similar to our tasks is MoMaRT, which uses imitation learning to control a Fetch robot in simulation and in mostly seen environments~\cite{wong2022error}. In contrast, we focus on generalization to completely unseen environments as well as the real-world. ManipulaTHOR focus on the higher-level task aspects, using a strongly simplified robot arm and magic grasping action~\cite{ehsani2021manipulathor}. Rearrangement-style benchmarks define tasks as moving the environment from an initial to a goal state~\cite{batra2020rearrangement}. BEHAVIOR~\cite{srivastava2022behavior} and Habitat-2.0~\cite{szot2021habitat} instantiate a large number of tasks in simulation. These approaches often abstract from the actual kinematics with actions such as "magical" grasping actions that immediately succeed~\cite{srivastava2022behavior, szot2021habitat, ehsani2021manipulathor} or explicitly provided lower-level motion primitives~\cite{srivastava2022behavior, gan2021threedworld}. In contrast, we assume that the agent is aware of the high-level task goals in form of access to an end-effector motion, but has to learn to achieve them with the actual robotic kinematics. Secondly, these approaches achieve limited success even when training separate agents specific to each task. We provide an approach that generalizes to unseen tasks without any re-training or finetuning. A very interesting aspect will be to incorporate and combine our work with the higher-level task focus of these benchmarks.

We formulate our approach as goal conditional RL problem~\cite{kaelbling1993learningto, schaul2015universal} which conditions its policy on a goal state to arrive at this goal.
Hierarchical methods~\cite{sutton1999between, bacon2017option, kaelbling1993hierarchical} abstract complex long-horizon tasks into a high-level policy proposing subgoals and goal-conditional low-level policies. While adapting these methods for mobile manipulation can improve sample efficiency~\cite{li2019HRL4INHR}, it still has to deal with the complexity and non-stationarity of learning hierarchical policies.

{\parskip=5pt\noindent\textit{Articulated Objects} have been a particular focus for mobile manipulation. Mittal~\textit{et~al.}~\cite{mittal2021articulated} estimate the articulation parameters of unseen objects, then use this to generate keyframes for the end-effector. Kineverse constructs articulation models for complex kinematics and demonstrates how these can be used to generate motions~\cite{rofer2022kineverse}. These works provide simple ways to generate end-effector motions for a wide variety of objects, complementing our approach and making it even more broadly applicable.}

\section{\ourslong}\label{sec:approach}

\begin{figure*}
\centering
	    \footnotesize
    \setlength{\tabcolsep}{0.3cm}
    {\renewcommand{\arraystretch}{1}
    \resizebox{\textwidth}{!}{%
    \begin{tabular}{cc}
\includegraphics[width=0.485\textwidth,trim={0.0cm 0cm 0cm 0cm},clip,angle =0]{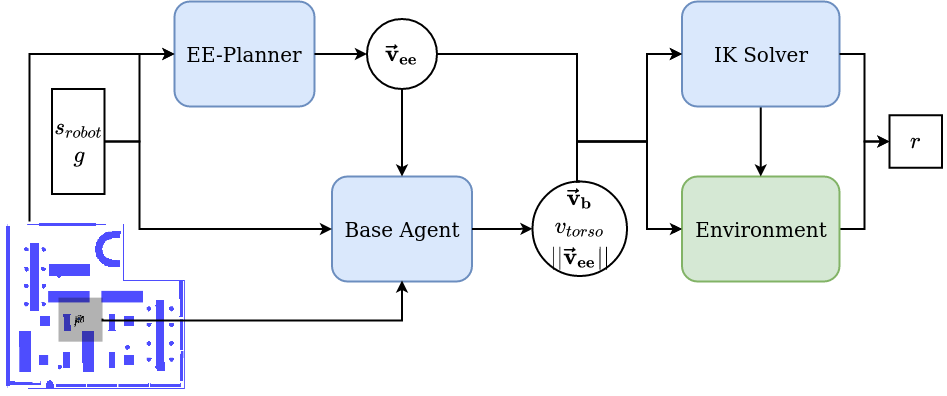} &
\includegraphics[width=0.485\textwidth,trim={0.0cm 0cm 0cm 0cm},clip,angle =0]{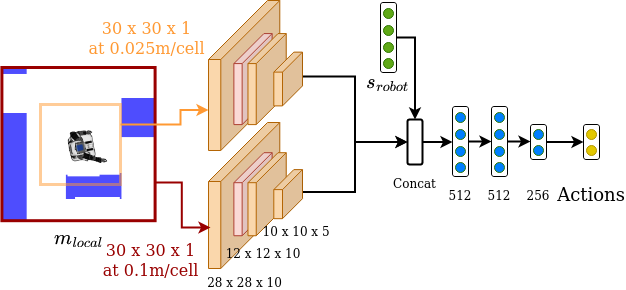}
        \\
        {(a)} & {(b)}
    \end{tabular}}
    }
    \vspace{-0.3cm}
\caption{(a) \ours: We decompose mobile manipulation tasks into two components: an end-effector (EE) motion generation and a conditional RL agent that controls the base \myworries{velocity $\vec{v}_b$}, the torso lift joint \myworries{velocity $v_{torso}$} and the \myworries{norm of the} velocity of the end-effector motion, \myworries{$||\vec{v}_{ee}||$}. The agent receives a local map of the environment (shown in grey) together with the robot state \myworries{$s_{robot}$}, an end-effector goal $g$ and the next end-effector motions \myworries{$\vec{v}_{ee}$}. Given the agent's actions, an inverse kinematics solver then solves for the remaining arm joints \myworries{and together with the environment returns the reward $r$}. 
\myworries{(b) Network architecture of the RL agent, illustrated for the actor. The critic uses the same architecture but  additionally conditions on the actions by concatenating them with the robot state and outputs a value estimate instead of a vector of actions. In orange are convolutional layers with kernel size (3, 3) and stride 1. Red are max pool layers with stride 2. All non-final layers are followed by a ReLU activation. The networks consist of 935,036 parameters for the actor and 935,283 for each of the critic's two Q-Networks.}
}
\label{fig:rl_scheme}
\end{figure*}

Mobile manipulation in real-world environments is a complex long-horizon task that combines navigation and control in a high-dimensional action space while respecting constraints imposed by the task, the hardware, and the environment. At the core, we decompose the mobile manipulation problem into an end-effector motion and base velocities learned by a reinforcement learning agent with the goal to ensure that this concurrent end-effector and base motions are kinematically feasible. We extend the formulation introduced by \cite{honerkamp2021learning} to enable the reinforcement learning agent to simultaneously avoid collisions with the environment and further expand its control to the velocity of the end-effector motions and the height adjustment of the robot torso.
We then generalize the objective and training scheme to learn policies that avoid configuration jumps and generalize to unseen environments while remaining applicable to a diverse set of mobile manipulation platforms. The resulting approach enables it to solve a very broad range of tasks in unstructured real-world environments. An overview of the proposed approach is depicted in \figref{fig:rl_scheme}.

\subsection{End-effector Motions}

Defining tasks such that they can be achieved by algorithms while remaining generally applicable yet simple to specify for humans is a hard problem. Much of the recent work has focused on addressing this challenge through imitation learning or the definition of task-specific goal states. In the domain of mobile manipulation, outcomes often depend on specific end-effector operations. As such we define tasks by end-effector motions. This is on one hand very expressive: a wide variety of tasks can be expressed this way. On the other hand, it provides a simple interface for humans to define and generate novel tasks quickly and unambiguously, as it requires only to reason about motions in simple Cartesian space, instead of a whole-body control problem. We provide a wide variety of example tasks in \secref{sec:experiments}. Further, this definition is not overly restrictive in terms of methodology as various types of motion generators for the end-effector can be employed, such as dynamic system-based imitation learning, planning-based systems, or even a reinforcement learning system.  

In particular, we only assume access to an arbitrary end-effector motion generator $f_{ee}$ that, given a current (partial or fully known) map $m_{global}$ of the environment, the current end-effector pose $ee_t\myworries{\in SE(3)}$ and velocities $v_{ee, t}\myworries{\in \mathbb{R}^6}$, and the current end-effector goal $g$, outputs next-step velocities $v_{ee, t+1}$ for the end-effector. 
\myworries{This motion generator does not handle the kinematic feasibility of the whole-body motion. Its only constraints are \myworriestwo{to avoid} collisions of the end-effector with the environment and not to violate fundamental affordances of the robot's hardware, e.g. not creating end-effector poses outside the reachable space such as poses that are too high to reach regardless of the base positioning. Collision constraints for the robot base as well as kinematic feasibility constraints will be handled by the learned base agent described bellow.}
We can define this \myworries{end-effector motion generation} as a function
\begin{equation}
    v_{ee, t+1} = f_{ee}(ee_t, v_{ee, t}, m_{global}, g).
\end{equation}


\subsection{Learning Base Control for Kinematic Feasibility}

We formulate mobile manipulation tasks as a goal conditional reinforcement learning problem in which the RL agent observes end-effector motion and goal, and aims to ensure that these motions remain kinematically feasible~\cite{honerkamp2021learning}. The agent controlling the robot's base is operating in a Partially Observable Markov Decision Process (POMDP) $\mathcal{M} = (\mathcal{S}, \mathcal{A}, \mathcal{O}, T(s' | s, a), P(o | s), r(s, a))$ where $\mathcal{S}, \mathcal{O}$ and $\mathcal{A}$ are the state, observation and action spaces, $T$ and $P$ describe the transition and observation probabilities, \myworries{$s$, $s'$ are the current and next state, $o$ is the current observation, $a$ is the current action} and $r$ and $\gamma$ are the reward and discount factor. The agent's objective is to learn a policy $\pi(a | o)$ that maximises the expected return $\E_\pi[\sum_{t=1}^{T} \gamma^t r(s_t, a_t)]$. In the goal conditional setting~\cite{kaelbling1993learningto}, the agent then learns a policy $\pi(a | o, g)$ that maximises the expected return $r(s, a, g)$ under a goal distribution $g \sim \mathcal{G}$ as $\E_{\pi, \mathcal{G}}[\sum_{t=1}^{T} \gamma^t r(s_t, a_t, g)]$. At each step, an arbitrary end-effector motion generator produces the next step velocities $v_{ee}$ for the end-effector. The reinforcement learning agent receives an observation $o \myworries{= [v_{ee}, ee, \hat{ee}, g, s_{robot}, a_{t-1}, m_{local}]}$ consisting of these velocities $v_{ee}$, the current and resulting desired EE poses \myworries{$ee$ and $\hat{ee}$}, and an end-effector goal pose $g$ in the robot's base frame together with the current robot state consisting of the joint \myworries{angle} configurations, the agent's previous actions $a_{t-1}$ and a binary local obstacle map $m_{local}$ centered and oriented around the base and outputs \myworries{actions} $a \sim \pi(a | o, g)$ \myworries{consisting of base velocities $\mathbf{v_b}$, the torso lift joint velocities $v_{torso}$ and the norm of the end-effector velocities $n_{ee}$}. In practice, we do not let the agent observe the final end-effector goal which can often be far away, but we repeatedly apply the end-effector motion generator $f_{ee}$ to generate an intermediate goal roughly \SI{1.5}{\meter} ahead of the agent.

The objective of the agent is to ensure that these end-effector motions remain feasible. Given the desired end-effector motion and the agent's base commands, we use inverse kinematics (IK) to solve for the remaining degrees of freedom in the arm. \myworries{The IK solver is restricted to solutions that adhere to joint limit and self-collision constraints. \myworriestwo{This means} the agent has to produce base motions such that the end-effector poses can be achieved while staying within these constraints, failure to do so will result in large deviations from the desired end-effector poses.} To achieve this objective, we extend our previous work~\cite{honerkamp2021learning} in several dimensions. First, we generalize the kinematic feasibility reward as
\begin{equation}
    r_{ik} = - ||\hat{ee}_{xyz} - ee_{xyz}||^2 - c_{rot} * d_{rot}(\hat{ee}, ee),
\end{equation}
where $ee$ and $\hat{ee}$ are the achieved and desired end-effector poses with position and orientation components $ee_{xyz}$ and $ee_{o}$, $d_{rot}$ calculates a rotational distance of the quaternions $d_{rot} = 1.0 - \langle\hat{ee_o}, ee_o\rangle^2$ and $c_{rot}$ is a constant scaling both components into a similar range.
This allows us to explicitly trade-off precision in the achieved pose with adhering to executable arm motions within the IK solver to address the configuration jumps observed in \cite{honerkamp2021learning}. We use the flexible BioIK solver and regularise the IK solver with a minimum displacement goal, consisting of the squared distances of each joint, weighted by the reciprocals of the maximum joint velocities~\cite{ruppel2018cost}. While this restricts the arm joints to stay close to the last time step, for robots with flexible arms there may still be several possible configurations for a given end-effector motion. Moreover, the selected arm configuration can be important to fulfill the remaining motions. To increase the RL agent's control over what configuration the arm goes into, we extend its control to the torso lift joint. This joint is often very slow and as a consequence requires a substantial look-ahead to change significantly in height. We extend the agent's action space with a learned torso velocity and use the IK solver only for the remaining arm joints.

An episode terminates early if the gripper deviates too much from the desired pose for more than 20 steps. We optimize this as an infinite horizon task, correcting for the (non-markovian) early terminations~\cite{pardo2018time} and taking into account that robots are expected to act in the real world in which they will have to continue reaching new goals after fulfilling the current task. To do this we continue to bootstrap the value functions in the final episode states. Lastly, we replace the regularization of the agent's actions' norm with a regularisation of the acceleration, incentivizing smooth instead of small actions:
\begin{equation}
    r_{acc} = ||a_{t} - a_{t-1}||^2.
\end{equation}
To ensure the environment remains Markovian, we extend the observation space with the agent's previous actions. In our experiments, we find this regularisation to be essential for the transfer to the real world.




\subsection{Obstacle Avoidance}

To perceive the environment, we equip the agent with a local occupancy map of its surroundings. Occupancy maps provide several advantages:
\begin{enumerate*}[label=(\roman*)]
\item They can be constructed from and integrate several sensors, such as LiDAR, RGB-D, and range sensors, ensuring applicability across different robots with different sensors.
\item They offer a geometric abstraction over sensors such as RGB images, which we hypothesize is easier to generalize to unseen objects and environments at test time.
\item They can be constructed in either 2D or 3D voxel maps.
\end{enumerate*}
In the following, we rely on a 2D representation which as we will show in the experiments is sufficient for a large range of real-world tasks while keeping computational costs low and during training can be generated without having to rely on a complex simulator.


The agent receives a robot-centered map $m_{local}$ with a side length of three meters. This map is processed by a map encoder before being concatenated with the robot's proprioceptive state and the goal observation. The map encoder receives two versions of the local map: a coarse one at a resolution of \SI{0.1}{\meter} and a fine-grained version with a side length of \SI{0.75}{\meter} at a resolution of \SI{0.025}{\meter}. \myworries{While the low resolution enables learning of larger spatial contexts, the fine-grained map reduces the impact of the grid discretization and enables the precise perception of available space in narrow corridors while keeping computational costs minimal}. The concatenation of the two maps is passed through three convolutional layers, the first followed by a maxpool, with ReLU activations, and then flattened into a vector. \myworries{The network is illustrated in \figref{fig:rl_scheme} (b)}.
We then extend the reward with a collisions penalty of 
\begin{equation}
    r_{coll} = - 10 \myworries{* \mathds{1}_{coll}},
\end{equation} \myworries{where $\mathds{1}_{coll}$ is a binary collision indicator for the base of the robot. As the RL agent directly learns a mapping from observations to actions, the resulting approach can handle both convex and non-convex obstacles \myworriestwo{(see \figref{fig:train_task} and video)}}.


\setlength{\tabcolsep}{1pt}
\renewcommand{\arraystretch}{1}
\begin{figure*}
	\centering
    \footnotesize
    \setlength{\tabcolsep}{0.0cm}
    {\renewcommand{\arraystretch}{1}
    \resizebox{\textwidth}{!}{%
    \begin{tabular}{ccccc}
  		\includegraphics[width=0.205\textwidth,trim={7.0cm 1.5cm 7.0cm 1.0cm},clip,angle=0]{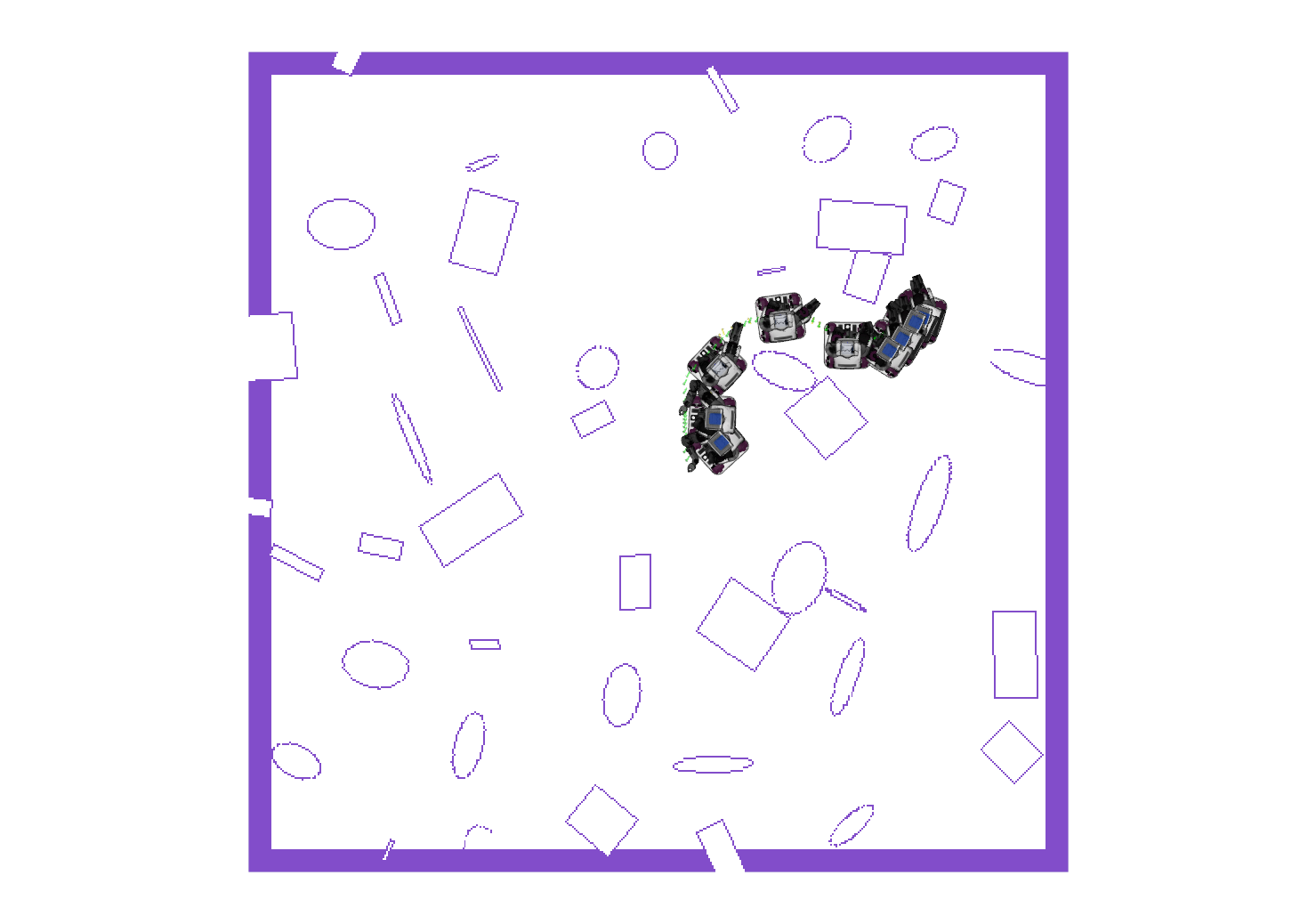} &
  		\includegraphics[width=0.20\textwidth,trim={0.0cm 0.0cm 0.0cm 0.0cm},clip,angle=0]{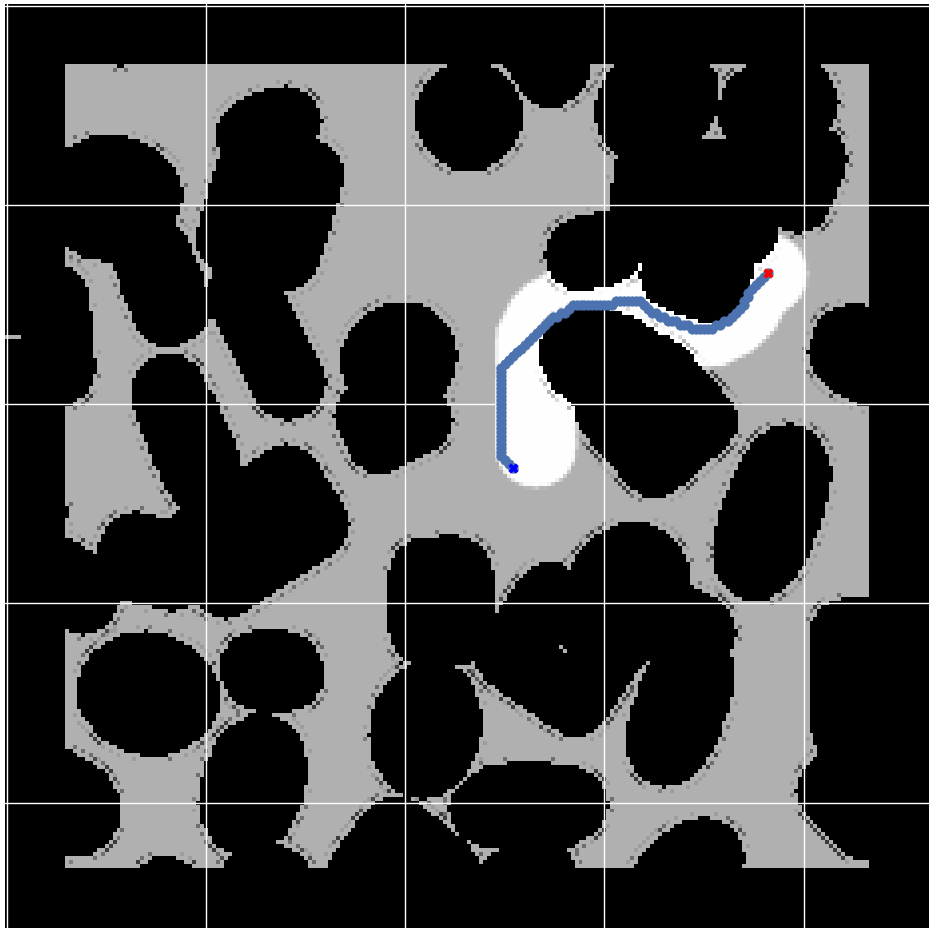} &
        \includegraphics[width=0.195\textwidth,trim={13cm 2.0cm 13cm 2.0cm},clip,angle=0]{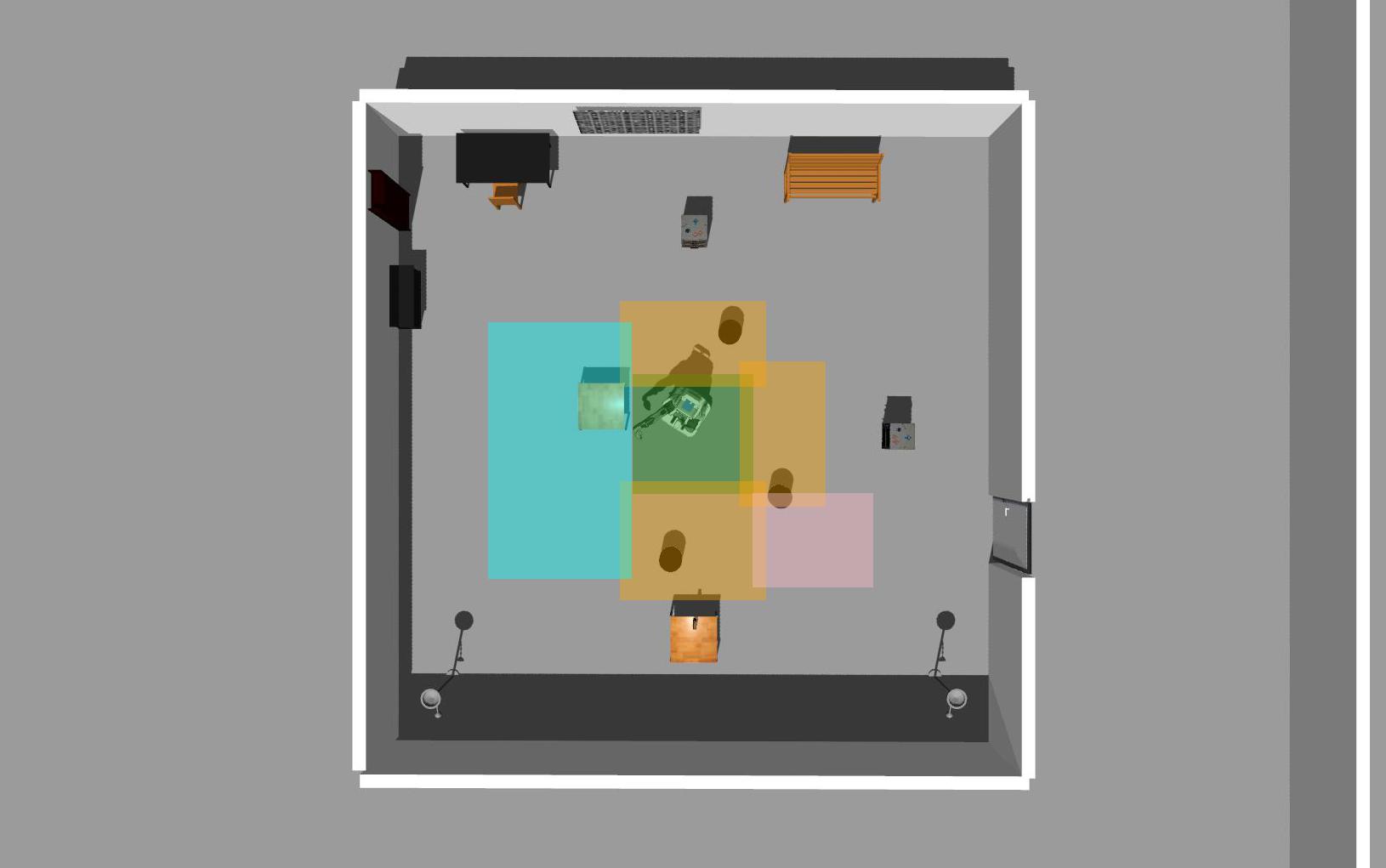}	&
        \includegraphics[width=0.25\textwidth,trim={17cm 0.0cm 19cm 0.0cm},clip,angle=0]{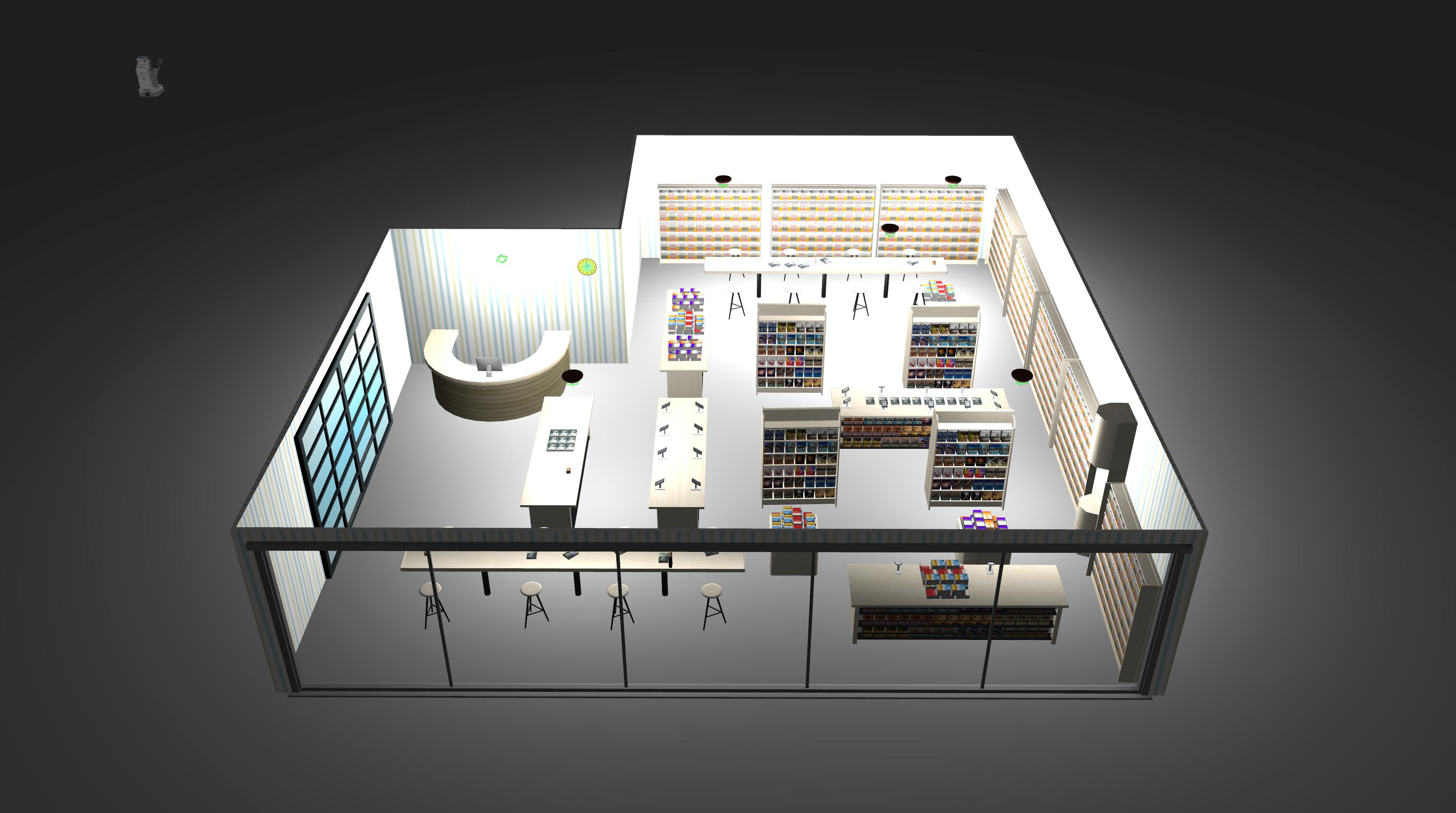} &
        \includegraphics[width=0.24\textwidth,trim={7cm 0.0cm 7cm 0.0cm},clip,angle=0]{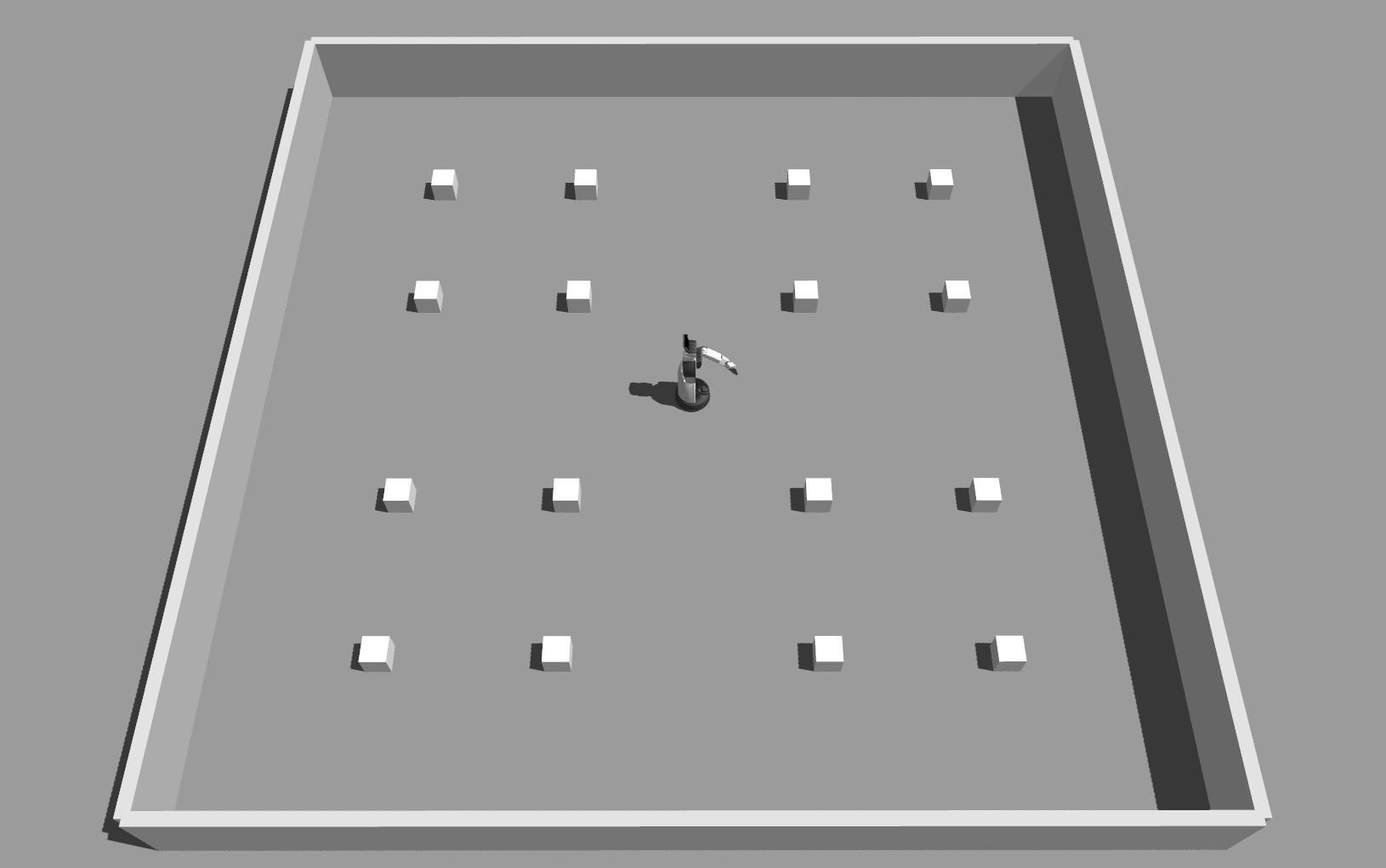}	
        \\
        {(a)} & {(b)} & {(c)} & {(d)} & {(e)}
    \end{tabular}}
    }
    \vspace{-0.3cm}
	\caption{Depiction of the training and testing environments. (a) procedurally generated training task. (b) generated weights and path of the end-effector planner according to \eqref{eq:weights}. The darker the image, the higher the cost. (c) articulated object environment. The starting area of the robot is shown in green, the spawn area of the obstacles is shown in orange (except for door and spline tasks), the starting area for the door task is shown in pink. For the p\&p task, the robot has to grasp a cereal box on the bottom table and place it onto a table randomly spawned in the cyan area. (d) bookstore map. (e) dynamic obstacle task.}
  	\label{fig:train_task}
\end{figure*}
\setlength{\tabcolsep}{6pt}
\renewcommand{\arraystretch}{1}

\subsection{Controlling Motion Velocities}

While we can express discretized motions as a time-stamped sequence of poses in SE(3), many real-world tasks do not depend on a particular execution speed. Furthermore, in cluttered environments, the robot base has to frequently evade obstacles, potentially requiring significantly longer paths than the end-effector. As a consequence, we propose to also learn to control the norm of the velocity of the end-effector motions. We extend the agent's action space by an additional action $n_{ee}$ that controls the norm of the end-effector velocity after observing the desired next motion by scaling it to 
\begin{equation}
\myworriestwo{v'_{ee} = n_{ee} * \frac{v_{ee}}{||v_{ee}||}.}
\end{equation}
To incentivize fast motions whenever possible, we add the following reward:
\begin{equation}
    r_{vel} = - (v_{ee, max} - n_{ee})^2,
\end{equation}
where $v_{ee, max}$ is the maximum velocity the end-effector is allowed, which we set equal to the maximum base velocity.

To prevent the agent from finding fast movement through difficult poses as a valid strategy to minimize collisions or ik failures, we furthermore transform the penalties for collisions and ik failures from a reward per time step into a reward per distance by multiplying them with the normalization term
\begin{equation}
    n_{vel} = \frac{n_{ee}}{v_{ee, max}}.
\end{equation}

\subsection{Overall Objective}

Combining the objectives, the overall reward function is
\begin{equation}
    r = n_{vel} (\lambda_{ik} r_{ik} + r_{coll}) + \lambda_{vel} r_{vel} + \lambda_{acc} r_{acc},
\end{equation}
\myworries{where $\lambda_{ik}$, $\lambda_{vel}$ and $\lambda_{acc}$} are constants to balance the objective (see \tabref{tab:hyperparams}). We optimize this objective with model-free reinforcement learning, namely soft-actor critic (SAC) which has been shown to effectively learn robust policies for continuous control tasks~\cite{haarnoja2018soft}.

\subsection{Training Environment}\label{sec:training}

To generalize to unseen motions and obstacles, the agent has to be exposed to a wide variety of motions, poses, and obstacles. We introduce a procedurally generated environment that uses elementary shapes to generate obstacles and a simple yet effective obstacle-avoiding end-effector motion generator. As we will demonstrate in our experiments, the resulting policy generalizes to unseen geometries and motions at test time.

{\parskip=5pt\noindent\textit{Task}:}
In each episode, we randomly arrange elementary shapes, namely rectangles and ellipses, on an empty map. Each obstacle is placed on a regular grid, offset by a random number drawn from a normal distribution, and placed in a random orientation. The obstacle's width, breadth, and height are drawn from a uniform distribution. An example of such a map is depicted in \figref{fig:train_task}. Given the obstacle map, we set a goal reaching task by sampling a random start pose, random initial joint values, and a random goal position within a distance of \SI{0.5}{\meter} to \SI{5} {\meter} from the start. 

While procedurally generated environments have proven very powerful to learn robust policies, they come with their challenges. In particular, we have to ensure that the generated task is solvable. For this, we follow a simple heuristic and reject any goal for which no valid path can be found in a map with an inflation radius of \SI{0.4}{\meter}. While more elaborate methods such as curricula~\cite{klink2020self} and adversarial environment generation~\cite{dennis2020emergent} have shown promising results, we found it not just simpler but also more effective to directly randomize and maximize the diversity of the environment and goals. 
During training, we simply integrate the robot kinematics over time \myworries{and directly use the inverse kinematics solutions as next joint values for the arm}, allowing us to generate data very quickly and without relying on a complex simulator. We will refer to this as \textit{analytical environment} throughout the remainder of the paper. \myworries{In all other environments the inverse kinematics solutions are used as a goal for the low-level arm controllers.}

{\parskip=5pt\noindent\textit{End-Effector Motions}:}
For training, we propose an instance of a simple obstacle-aware \myworries{end-effector} motion generator that remains agnostic to the robot. In \secref{sec:experiments}, we evaluate how well the behavior learned on these motions generalizes to arbitrary, unseen motions.
To ensure that the desired motions are in general feasible, we use two heuristics and develop a refined A*-based planner to implement them as soft constraints:
\begin{enumerate}[label=(\roman*)]
    \item Avoid end-effector collisions: A minimum distance $d_{ee}$ of the end-effector from tall obstacles.
    This can be implemented by inflating the obstacle map, as is common practice in robot navigation. We ignore obstacles that are smaller than the maximum height the end-effector can reach (minus a small margin) $max\_z$, meaning the end-effector motion is allowed to pass over obstacles where kinematically possible.
    \item Ensure the base can follow the end-effector motions: we first plan a path from the initial base pose to the end-effector goal in a map that is inflated by the radius of the robot base.
    We then define a maximum distance $d_{base}$ that the end-effector is allowed to deviate from this base path. We set this value to the range of the robot's arm. 
    We implement this by inflating the base path by $d_{base}$ and adding the inverse of this to the weights. 
\end{enumerate}

Together, this results in the following weights $w$ of the A*-planner, consisting of the inflated end-effector path and the inverse of the inflated base path:
\begin{align}\label{eq:weights}
    w = c * \mathds{1}[&inflate(m_{global,z+}, d_{ee}) \nonumber \\
        & + (1 - inflate(m_{basepath}, d_{base}))],
\end{align}
where $c$ is a constant, $m_{global,z+}$ is a map with only the obstacles larger than $max\_z$ over which the end-effector cannot pass, and $m_{basepath}$ is a map consisting only of the shortest path from the initial base pose to end-effector goal obtained by A*. An example of the resulting weights is shown in \figref{fig:train_task}.

We then generate dense waypoints with A* and interpolate them with a linear dynamic system to produce the actual end-effector motions. We linearly interpolate the height over the found path from the current height towards $\max(next\_obstacle\_z + height\_margin, goal\_z)$. Rotations are obtained via spherical linear interpolation (slerp) from start to goal orientation.
As we demonstrate in the experiments, this results in an effective and yet simple method capable of generating motions for complex, real-world floor plans. Note that while more sophisticated methods could generate easier to achieve motions in crowded maps, our main interest here is to generate diverse motions for the RL agent. 
The task is deemed successful if the agent gets within \SI{2.5}{\centi\meter} and a rotational distance of 0.05 of the end-effector goal, has no base collisions or self-collisions and does not deviate more than \SI{10}{\centi\meter} or a rotational distance of 0.05 from the desired motion. We early terminate the episode if the agent violates any of these constraints for more than 20 steps.

\section{Experimental Evaluations}\label{sec:experiments}

We evaluate our approach across different robotic platforms, different environments, and a wide variety of tasks. In these experiments, we aim to answer the following questions:
\begin{itemize}
    \item Do agents trained with our proposed approach generalize to unseen end-effector motions?
    \item Does our approach generalize across robotic platforms with different kinematic abilities?
    \item Does the agent learn robust obstacle avoidance that generalizes to (i) occupancy maps generated by real-world LiDAR scans and (ii) unseen obstacles?
    \item Does our approach scale to complex, real-world based obstacle environments?
    \item Can the agent react to dynamic obstacles and changing environments?
    \item Does our approach transfer to direct execution in the real world?
\end{itemize}

\subsection{Experimental Robotic Platforms}

We train agents for three different robotic platforms differing considerably in their kinematic structure and base motion abilities. The \textit{PR2} robot consists of an omnidirectional base, a height-adjustable torso, and a $7$-DOF arm, giving it high mobility and kinematic flexibility. \myworries{With} a base diagonal of $\SI{0.91}{m}$ it is comparably large, limiting its maneuverability in narrow spaces. The \textit{Toyota HSR} robot has an omnidirectional base as well, but the arm is limited to $5$-DOF including the height-adjustable torso. As a result, it cannot reach all poses in SE3 unless it coordinates movements with its base. The \textit{TIAGo} robot is equipped with a height-adjustable torso similar to the PR2 as well as a flexible $7$-DOF arm. \myworriestwo{However,} it uses a differential drive, restricting its mobility compared to the PR2. All three robots are equipped with a base LiDAR sensor, covering 270, 240, and 220 degrees on the PR2, HSR, and TIAGo, respectively. The TIAGo additionally has three range sensors pointing backward, each with a field of view of $\SI{29 }{\degree}$ and a range of one meter. In the simulation, we also equip the PR2 and HSR with the same range sensors. The action space for these platforms is continuous, consisting of either one (diff-drive) or two (omni) directional velocities $\mathbf{v_{b, \{x, y\}}}$, and an angular velocity $\mathbf{v_{b, \theta}}$. This action space is completed with two more actions, one controlling the velocity of the end-effector motion and the other controlling the torso lift joint. \myworries{The only exception to this is} the HSR for which we do not find a benefit in learning the torso over solving for it with the IK as it only has limited DoF \myworries{and therefore do not learn the torso velocities}. \tabref{tab:constraints} lists the constraints we set across the different platforms in the analytical environment.

\begin{table}
    \centering
    \caption{Velocity and pose constraints.}
    \begin{tabularx}{\columnwidth}{l|YcYY}
      \toprule
        Parameter & EE-Motion & PR2 & TIAGo & HSR \\
      \midrule
        Max. velocity (m/s) & 0.2 & 0.2 & 0.2 & 0.2 \\
        Max. rotation (rad/s) & 0.3 & 1.0 & 1.0 & 1.0 \\
        Goal height (m) & - & [0.2, 1.55] & [0.2, 1.5] & [0.2, 1.4]\\
        Restr. height (m) & - & [0.4, 1.0] & [0.4, 1.1] & [0.4, 1.1]\\
      \bottomrule
    \end{tabularx}
    \label{tab:constraints}
\end{table}

\subsection{Baselines}

We compare our proposed approach against a range of methods from planning, optimal control, and machine learning.

{\noindent\textit{MPC}:}
As the main baseline, we compare against a recent model-predictive control approach~\cite{mittal2021articulated}. The method uses the SLQ algorithm to minimize the deviation to the desired end-effector trajectory with additional soft constraints for self-collision, joint velocity, and joint position limits. To ensure a fair comparison, we change all robot collision meshes to simplistic geometries, reduce the number of self-collision constraints to a bare minimum (avoiding end-effector -- torso and end-effector -- head collisions), and set all unused joints as fixed.
Collision avoidance is incorporated as a hinge-loss function based on a signed distance field. We provide it with a groundtruth 2D signed distance field at a resolution of \SI{2.5}{\centi\meter} from the obstacle map and model the base collisions as a cylinder centered on the robot base. For the rectangular base of the PR2, we define four more small cylinders at the corners of the base and implement the derivatives with regard to position and orientation. We follow the authors and run the algorithm at a control frequency of \SI{30}{\hertz} with a time horizon of four seconds. The agent has to follow the same end-effector motions as our approach. As the original code is not publicly available, we reimplement it based on the same optimization library\footnote{\url{https://github.com/leggedrobotics/ocs2}}.


{\noindent\textit{E2E}:}
An end-to-end reinforcement learning agent which does not rely on an IK solver but directly learns to control the whole robot by extending the action space to include the joint velocities for the arm. It receives the same desired end-effector motions and reward signal but directly acts in the full configuration space of the robot to achieve these motions. \myworries{In case of self-collisions, the previous joint values are kept. This is indicated to the agent via a binary variable that is true if there would have been a self-collision.}

{\noindent\textit{LKF}:}
The original learning kinematic feasibility approach~\cite{honerkamp2021learning} uses a binary indicator as a reward and does not take into account obstacles. In a range of ablation studies, we extend it until we reach our approach.

{\noindent\textit{RRTConnect}:}
For comparison with classical planning approaches, we evaluate RRTConnect~\cite{kuffner2000rrt-connect}. However, as general motion constraints over time cannot easily be defined in the sampling space, we omit these constraints. The planning algorithm only receives the same start and goal poses as the learning agent and is allowed to freely move to the target pose. As such it has much more freedom to achieve the goal, but at the same time is limited in applicability to goal reaching and pick\&place tasks. We use the planner implementations provided by the MoveIt motion planning library \cite{coleman2014reducing}. While samplers can in principle support arbitrary constraints such as non-holonomic bases, current ROS1 implementations do not support it. As Tiago is not yet ROS2 compatible, we do not evaluate it on this platform.

{\noindent\textit{Bi2RRT$^*$}:} A bidirectional RRT$^*$ algorithm developed for mobile manipulation with the PR2 which achieves efficient sampling by using a two-tree variant of Informed RRT$^*$~\cite{burget2016bi2rrt}. As for RRTConnect, we omit the motion constraints. We use the authors' implementation.

\subsection{Evaluation Setup}

As the main focus of this work is the base motions, we provide the EE-planner with a groundtruth map of the environment for all tasks except those involving dynamic environments. However, the RL agent is always restricted to the local map (groundtruth in the analytical environment, based on its sensors in all other environments). The planner baselines receive access to the full 3D groundtruth planning scene. Only for the planner baselines, we additionally simplify all complex collision meshes in the bookstore map with simple geometries to reduce the impact of collision checking on the planning times. We tune the MPC and planner baselines for each robot via grid search on the rnd obstacle task, then evaluate the best set of parameters on all tasks. For our approach, we use a single set of parameters across all robots without any further platform-specific tuning. Hyperparameters for all methods are reported in \tabref{tab:hyperparams}. The approaches were evaluated on an AMD Ryzen 9 5900X or AMD EPYC 7452 CPU. As the PR2 Gazebo physics no longer matches the real robot\footnote{\url{https://github.com/PR2/pr2_controllers/issues/402}}, we update it to roughly fit the actual hardware by setting the wheel, and rotation joint efforts to 25 and the wheel controller's proportional gains to 200.

{\parskip=5pt\noindent\textit{Metrics}:}
A task is deemed successful if the end-effector reaches the goal without any collisions and never deviates more than \SI{10}{cm} or a rotational distance $d_{rot}$ of 0.05 from the desired motion. In the analytical environment, we also count configuration jumps as a failure, where we define a jump as exceeding any of the robot's maximum joint velocities. In the analytical environment, we only track base collisions, in the Gazebo simulator we check all collisions with the robot body. We evaluate each task over 50 episodes. For learning-based methods, we additionally average over five models trained with different random seeds.


\setlength{\tabcolsep}{2pt}
\begin{table}
    \centering
    \caption{Hyperparameters for the different approaches.}
    \begin{threeparttable}
    \begin{tabularx}{0.485\textwidth}{lY|lY}
      \toprule
        \multicolumn{4}{c}{RL} \\
      \midrule
      tau                                   & $\expnumber{1}{-3}$   & buffer size                   & $\expnumber{10}{5}$\\
      lr                                    & $\expnumber{1}{-4}$   & lr end                        & $\expnumber{1}{-6}$ \\
      gamma                                 & 0.99                  & steps                         & $\expnumber{10}{6}$\\
      ent coef                              & learned               & train intensity               & 0.1\\
      batch size                            & 256                   & hidden layers                 & [512, 512, 256] \\
      action noise                          & 0.04                  & frame skip    &  8 (Tiago)   \\
      $\lambda_{ik}$                        & 50                    & $c_{rot}$                     & 2    \\
      $\lambda_{vel}$                       & 0.1                   & $\lambda_{acc}$               & 0.05 \\
      \midrule
        \multicolumn{4}{c}{MPC} \\
      \midrule
        time horizon                     & \{4 sec, 6 sec\}                             & frequency          & 30 Hz \\
        $\mu$ collision                     & \{500, 1000, 5000\}                         & $\delta$ collision    & \{0.001, 0.01, 0.1\} \\
        $\mu$ joint limit                  & \{0.01, 0.1\}                                & $\delta$ joint limit & \{0.001, 0.01\} \\
        min step                         & \{0.01, 0.03\} \\
      \midrule
        \multicolumn{4}{c}{RRTConnect} \\
      \midrule
        max waypoint    & \{0.1, 0.2, 0.4\} & range & \{default, 1, 10\} \\
        plan attempts   & \{1, 10\} & time limit & 200 sec \\
      \midrule
        \multicolumn{4}{c}{Bi2RRT$^*$} \\
      \midrule
      extend step  & \{0.1, 0.2, 0.4\} & time limit & 200 sec \\
      \bottomrule
    \end{tabularx}
  \begin{tablenotes}[para,flushleft]
       \footnotesize      
       Notes: For the planner and MPC baselines we perform a grid search on the obstacle task over these values and then report the results of the best parameters on all tasks.
     \end{tablenotes}
   \end{threeparttable}
    \label{tab:hyperparams}
\end{table}
\setlength{\tabcolsep}{6pt}


\begin{table*}
    \centering
    \caption{Success rates for experiments in the analytical environment.} 
    \begin{threeparttable}
    \begin{tabularx}{\textwidth}{cl|YYYYY|YYY|Y|Y}
    \toprule
           & & \multicolumn{5}{c|}{$A^*$-slerp} & \multicolumn{3}{c|}{Imitation Learning} & \multicolumn{1}{c|}{Spline} & \multicolumn{1}{c}{Avg} \\ 
    \cmidrule{3-12}
      & Agent                & rnd obstacle & p\&p & bookstore p\&p &  dynamic obstacle & dynamic p\&p & cabinet & drawer & door & spline & \\ 
    \midrule
        \parbox[t]{1mm}{\multirow{3}{*}{\rotatebox[origin=c]{90}{PR2}}}
        & MPC                &   54.0       & 70.0 &           ~8.0 & 64.0 & 54.0 & 62.0 &   76.0 &     32.0 & 96.0 & 57.3          \\ 
        & E2E                &   ~0.0       & ~0.0 &           ~0.0 & ~0.0 & ~0.0 & ~0.0 &   ~0.0 &     ~0.0 & ~0.0 & ~0.0          \\ 
        & \textbf{\ours}      &   86.0       & 97.6 &           70.0 & 84.4 & 81.6 & 97.2 &   97.6 &     98.4 & 80.8 & \textbf{88.2} \\ 
        \cmidrule{1-12}
        \parbox[t]{1mm}{\multirow{3}{*}{\rotatebox[origin=c]{90}{HSR}}}
        & MPC                &   64.0       & 72.0 &           48.0 & 60.0 & 60.0 & 68.0 &   68.0 &     66.0 & 98.0 & 67.1          \\ 
        & E2E                &   25.2       & 38.8 &           10.4 & 17.6 & 13.2 & 52.4 &   14.4 &     43.2 & 38.0 & 28.1          \\ 
        & \textbf{\ours}      &   63.2       & 92.0 &           68.0 & 82.0 & 88.0 & 93.2 &   87.2 &     88.0 & 84.0 & \textbf{82.4} \\ 
        \cmidrule{1-12}
        \parbox[t]{1mm}{\multirow{3}{*}{\rotatebox[origin=c]{90}{Tiago}}}
        & MPC                &   46.0       & 68.0 &           ~0.8 & 40.0 & 30.0 & 62.0 &   52.0 &    32.0  & 86.0 & 46.3          \\ 
        & E2E                &   ~0.0       & ~0.0 &           ~0.0 & ~0.0 & ~0.0 & ~0.0 &   ~0.0 &     ~0.0 & ~0.0 & ~0.0          \\ 
        & \textbf{Ours}      &   61.2       & 91.6 &           42.0 & 54.4 & 50.4 & 81.6 &   89.2 &     62.4 & 56.0 & \textbf{65.4} \\ 
    \bottomrule
    \end{tabularx}
        \begin{tablenotes}[para,flushleft]
       \footnotesize      
       Notes: Evaluation on unseen tasks from three different motions systems, an $A^*$-based system, an imitation system learned from human demonstrations and spline interpolation of random waypoints. The last column reports the average across all tasks. We evaluate all models on three different robotic platforms, the PR2, HSR, and TIAGo.
     \end{tablenotes}
   \end{threeparttable}
    \label{tab:analytical}
\end{table*}

\subsection{Generalization to Arbitrary Motions}\label{sec:arbitrary}

To evaluate the generalization to arbitrary motions, we generate a range of tasks from different motion systems:

{\noindent\textit{A$^*$-slerp}} refers to the A$^*$-based motions used during training, as described in \secref{sec:training}.
{\noindent\textit{A$^*$-fwd}} uses the same end-effector path but generates different desired end-effector orientations: Instead of spherical interpolation, it generates motions in which the end-effector always points in the direction of movement. Only when close to the goal, it uses slerp to achieve the desired final end-effector orientation. From these motions, we construct the training task, labeled \textit{rnd obstacle}, and a pick\&place (\textit{p\&p}) task in which the agent has to grasp a cereal box from one table and place it down on another table.

{\noindent An \textit{imitation learning system}} learned from a human teacher~\cite{twelsche2017learning}. The motions are encoded in a dynamic system following a demonstrated hand trajectory to manipulate a certain object. In particular, we use motions to grasp and open a cabinet, grasp and open a drawer and grasp a door handle, push it down and open the door inwards while driving through it. This task is particularly challenging as it requires avoiding both the door frame and the moving door, resulting in a highly constrained and narrow navigation task. With \SI{0.89}{m}, the door frame is more narrow than the PR2 base-diagonal of \SI{0.91}{m}. The moving door makes the environment dynamic as it changes the free space. \myworries{As these motions encode knowledge on how to interact with the objects, we can use them to enable object manipulations that the agent has not seen in training.}

{\noindent\textit{Spline interpolation}} draws five random waypoints in SE(3) with a distance of \SI{1}{\meter} to \SI{3}{\meter} from the previous and connects them using cubic splines and spherical interpolation. This results in arbitrary motions over a total distance of \SI{5}{\meter} to \SI{15}{\meter}.

\setlength{\tabcolsep}{1pt}
\renewcommand{\arraystretch}{1}
\begin{figure*}
	\centering
	\resizebox{.8\textwidth}{!}{%
	\begin{tabular}{ccc}
	    \includegraphics[width=0.2\textwidth,trim={0.0cm 0.0cm 0.0cm 0.0cm},clip,angle=0]{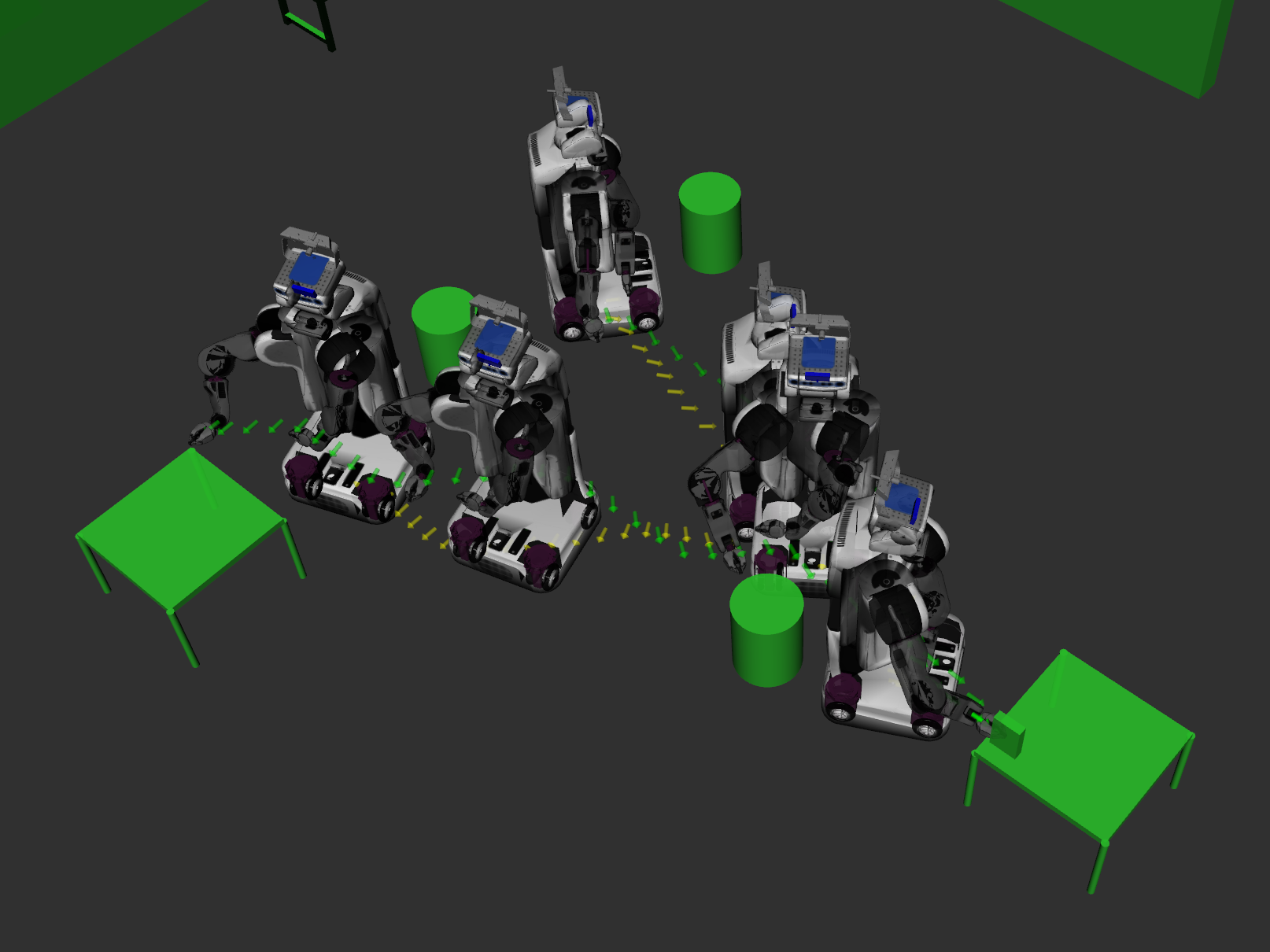} &
	    \includegraphics[width=0.2\textwidth,trim={0.0cm 0.0cm 0.0cm 0.0cm},clip,angle=0]{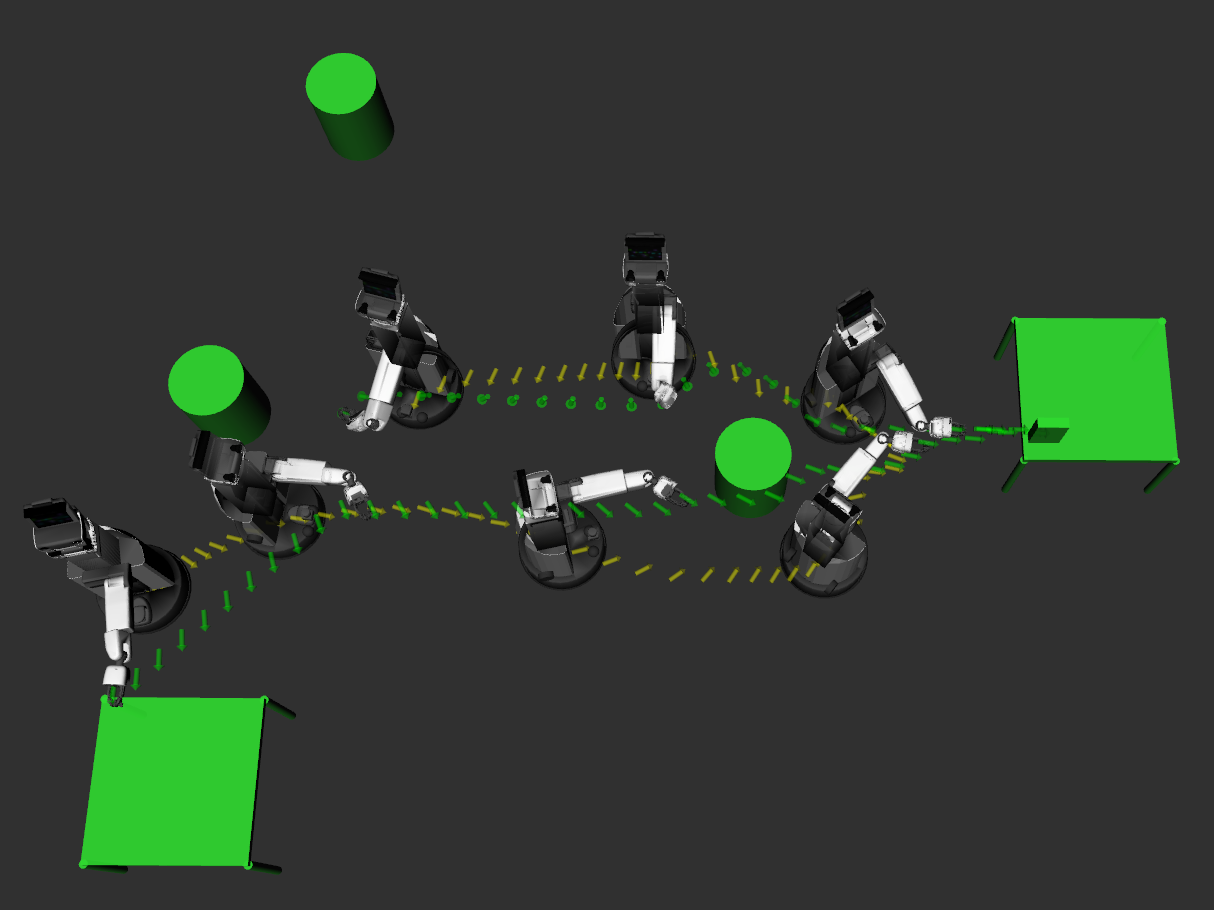} &
	    \includegraphics[width=0.2\textwidth,trim={0.0cm 0.0cm 0.0cm 0.0cm},clip,angle=0]{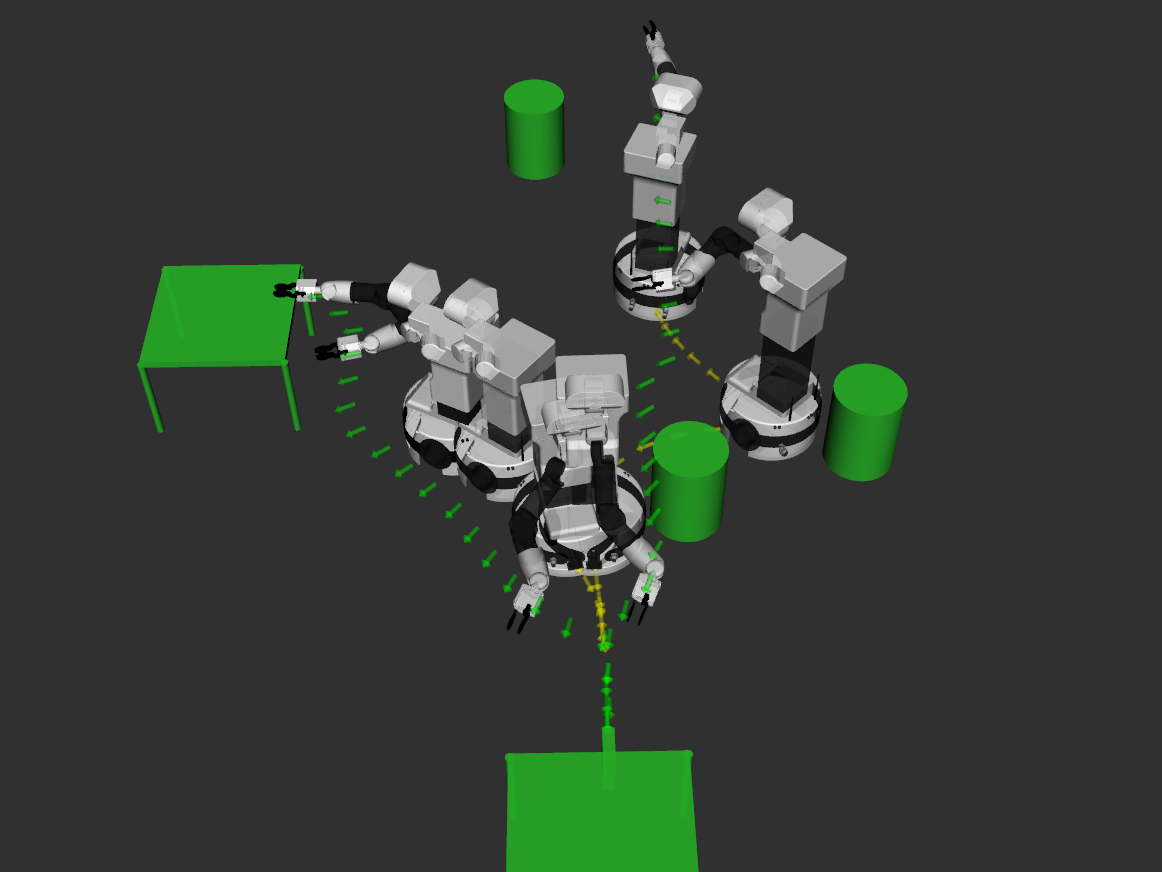}\\
	    \includegraphics[width=0.2\textwidth,trim={0.0cm 0.0cm 0.0cm 0.0cm},clip,angle=0]{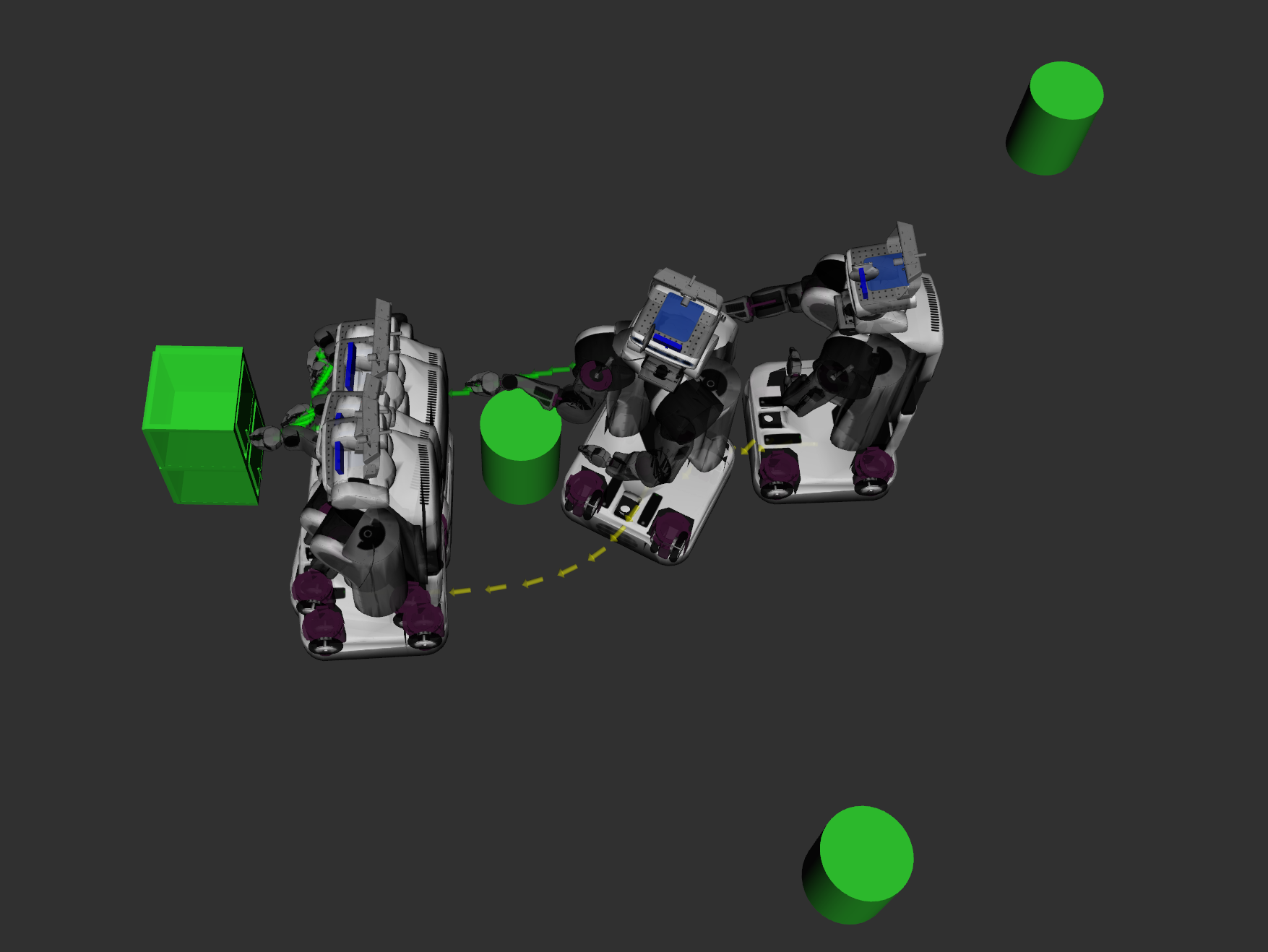} & 
	    \includegraphics[width=0.2\textwidth,trim={0.0cm 0.0cm 0.0cm 0.0cm},clip,angle=0]{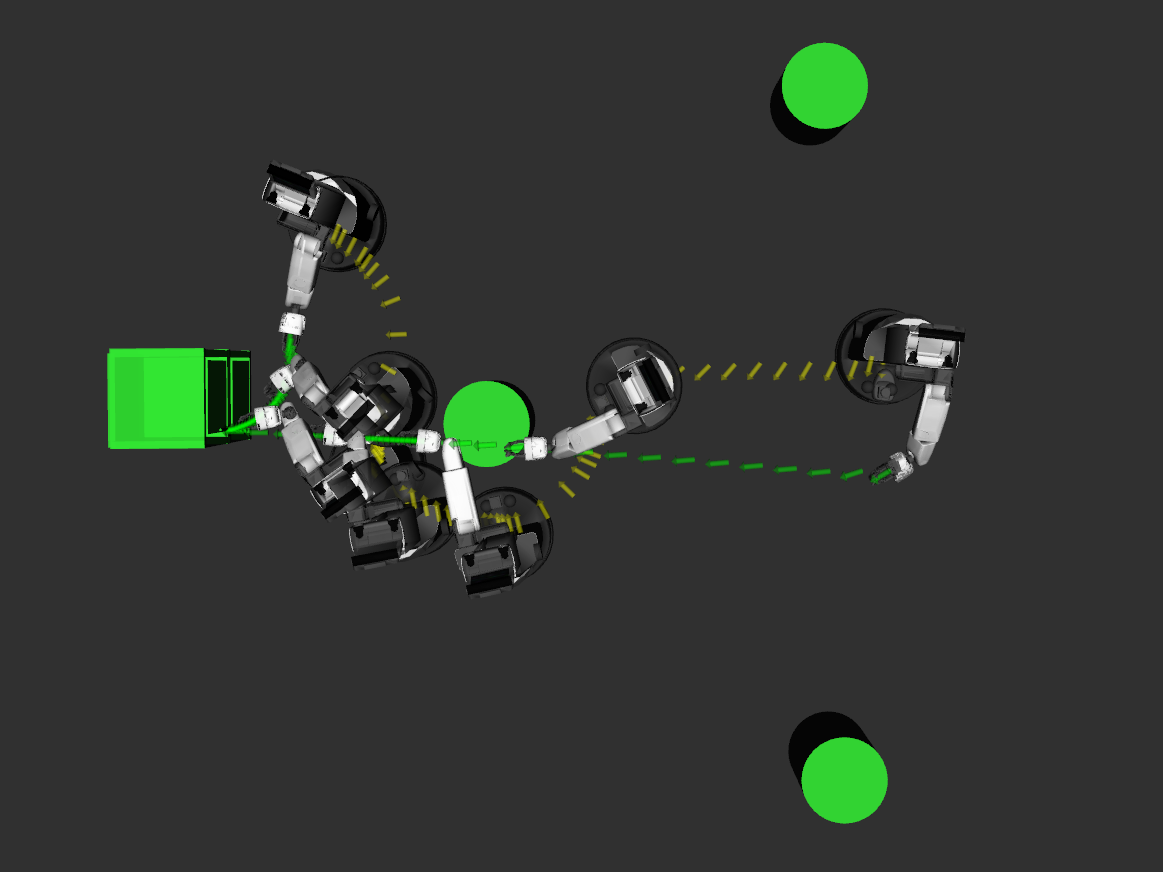} & 
	    \includegraphics[width=0.2\textwidth,trim={0.0cm 0.0cm 0.0cm 0.0cm},clip,angle=0]{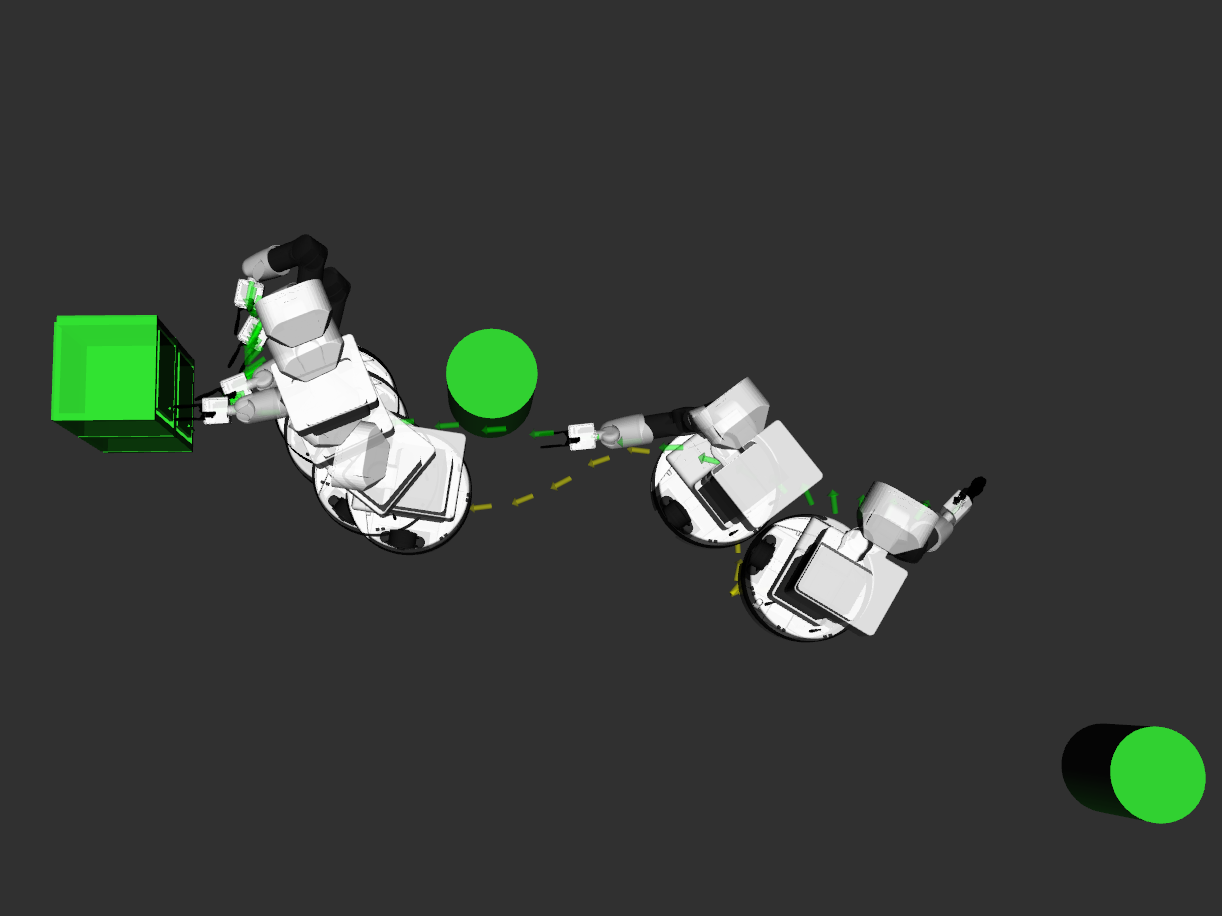}\\
	    \includegraphics[width=0.2\textwidth,trim={0.0cm 0.0cm 0.0cm 0.0cm},clip,angle=0]{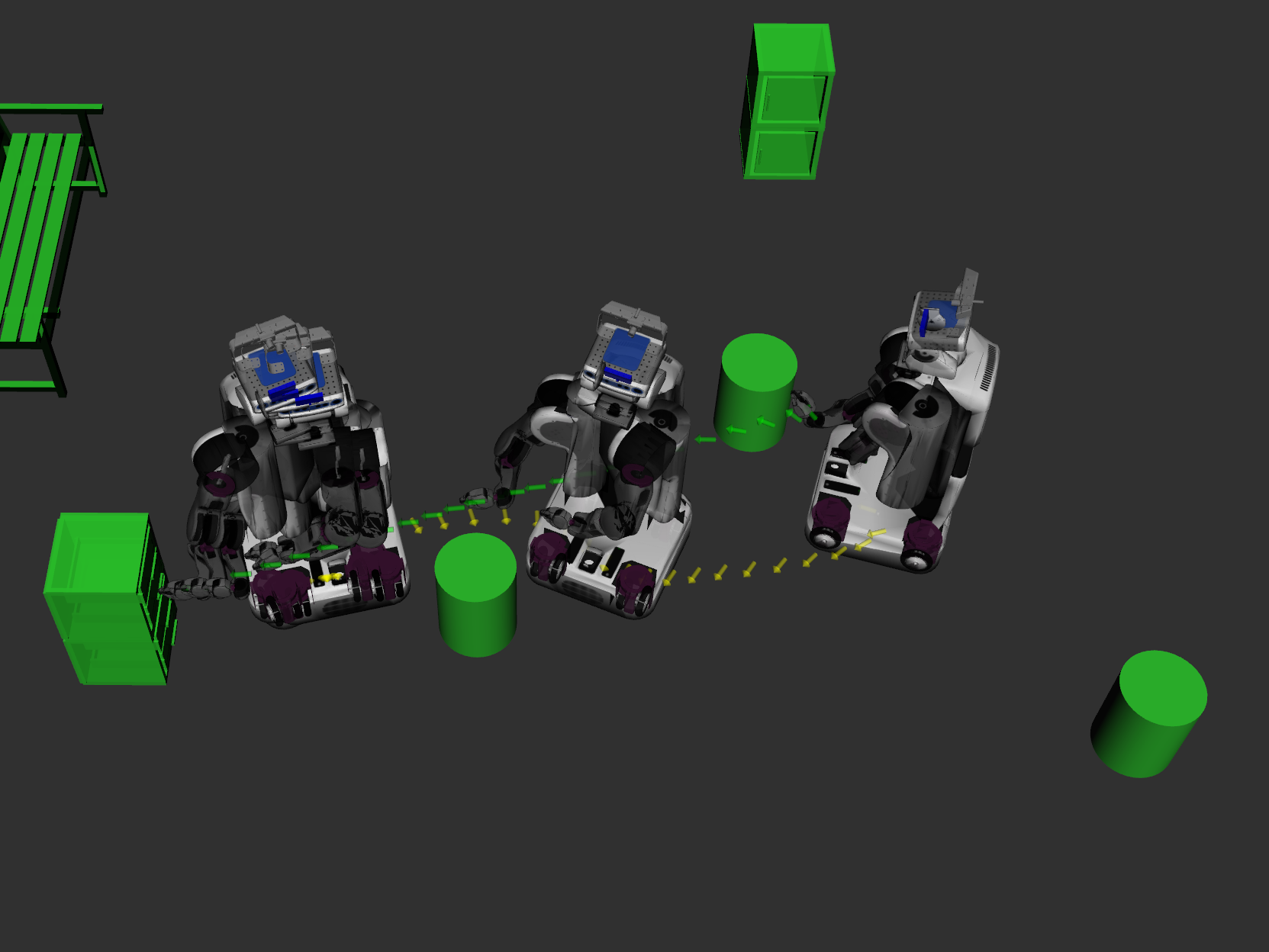} & 
	    \includegraphics[width=0.2\textwidth,trim={0.0cm 0.0cm 0.0cm 0.0cm},clip,angle=0]{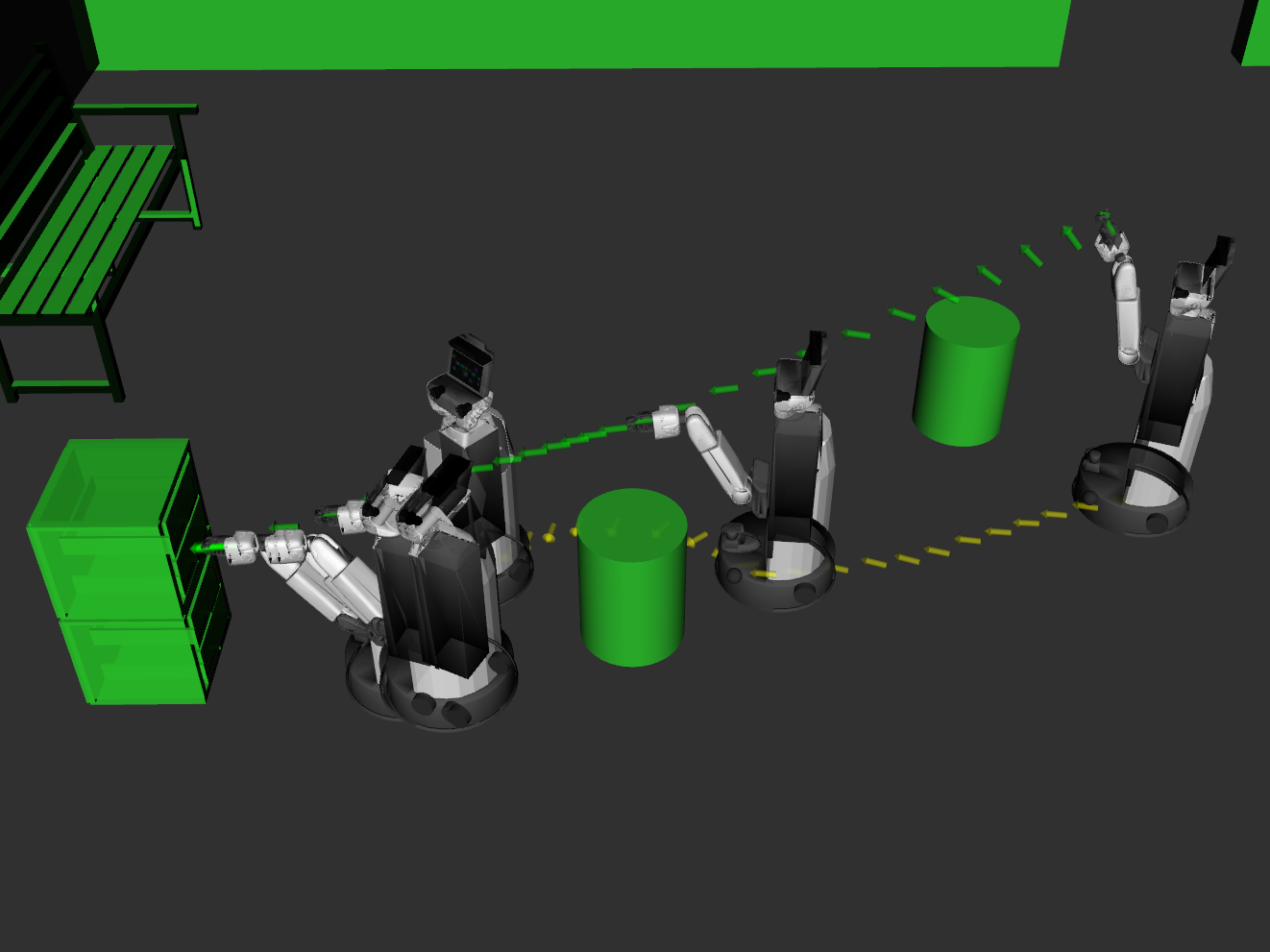} & 
	    \includegraphics[width=0.2\textwidth,trim={0.0cm 0.0cm 0.0cm 0.0cm},clip,angle=0]{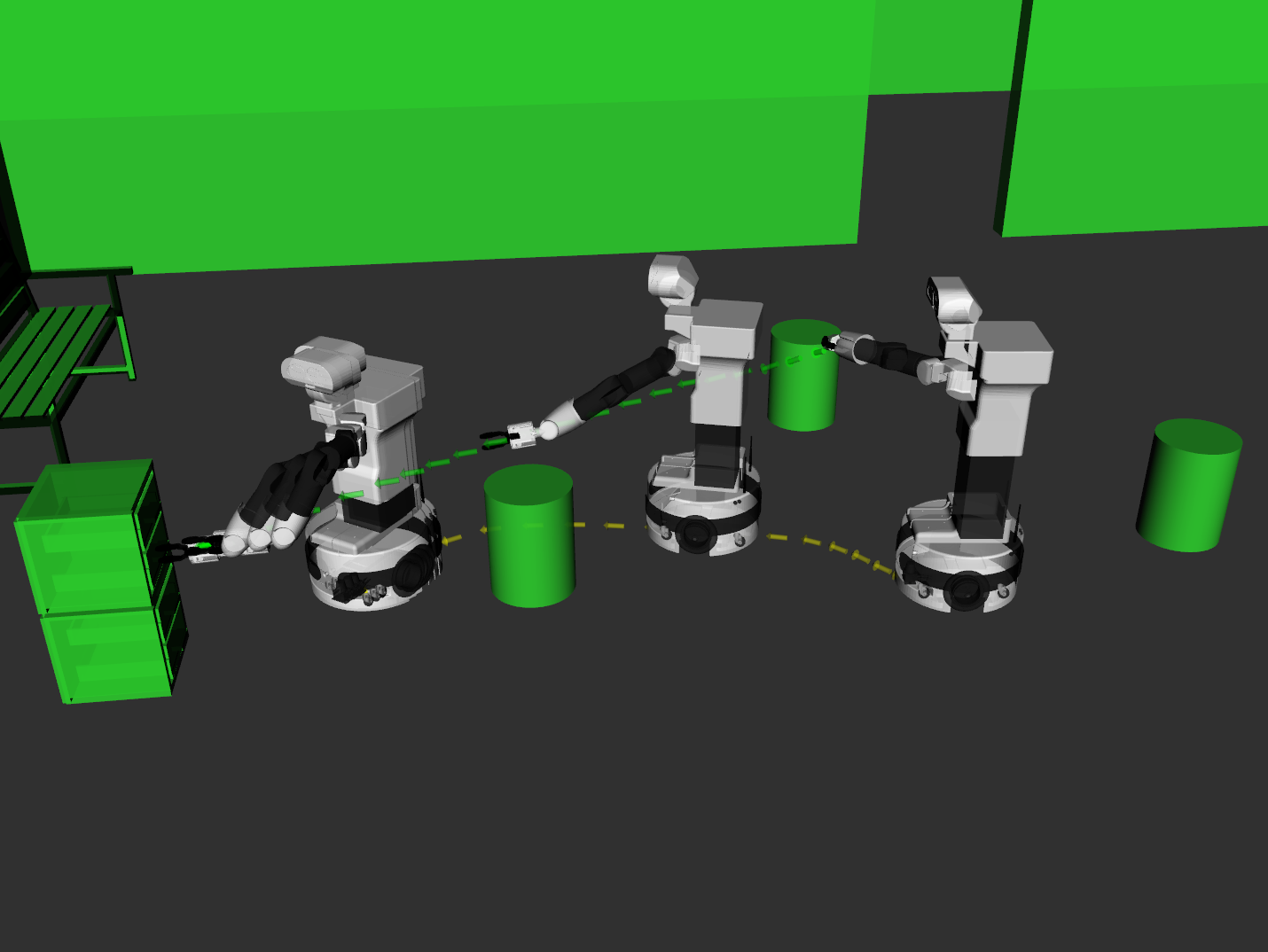}\\
	    \includegraphics[width=0.2\textwidth,trim={0.0cm 0.0cm 0.0cm 0.0cm},clip,angle=0]{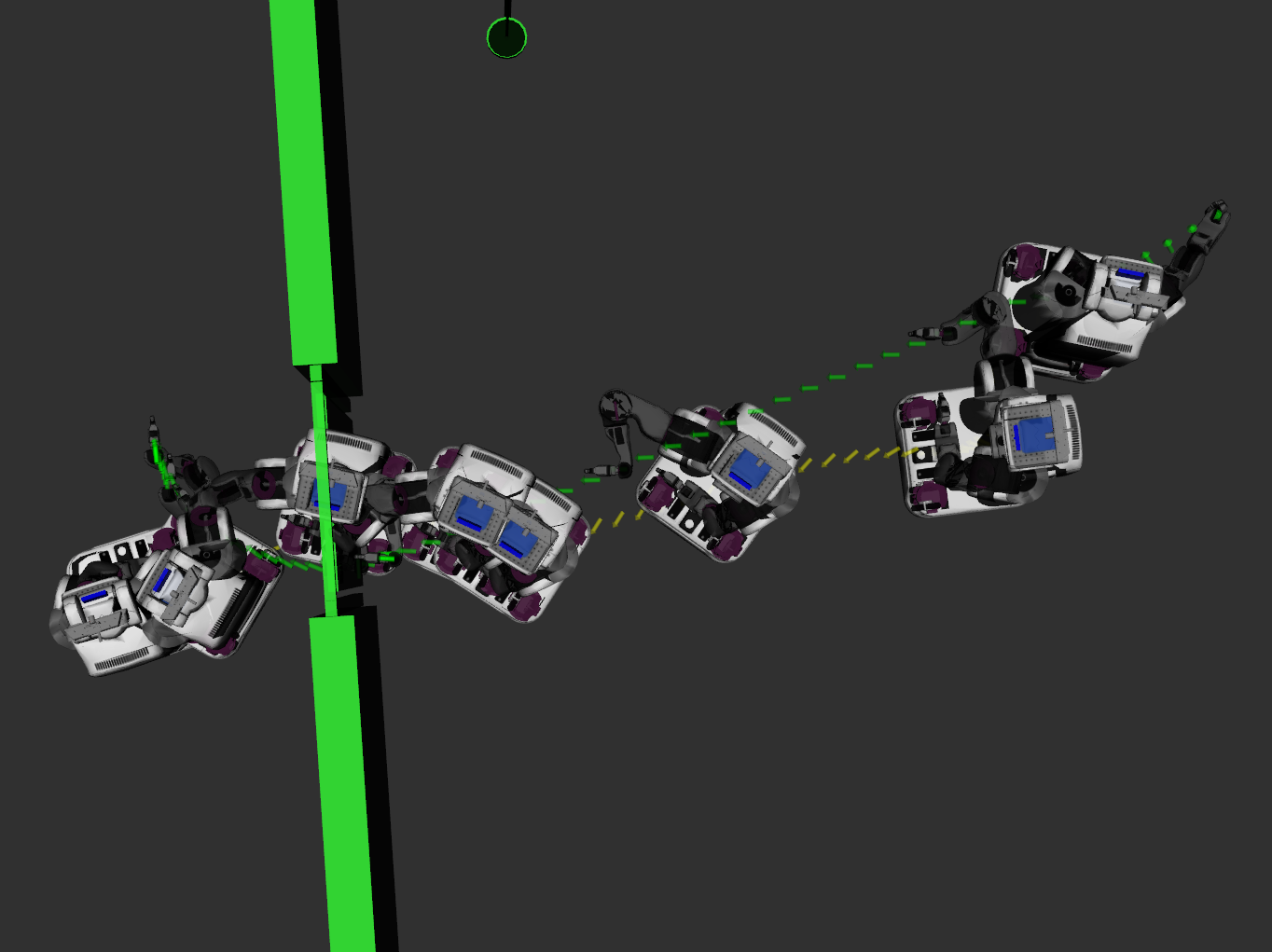} & 
	    \includegraphics[width=0.2\textwidth,trim={0.0cm 0.0cm 0.0cm 0.0cm},clip,angle=0]{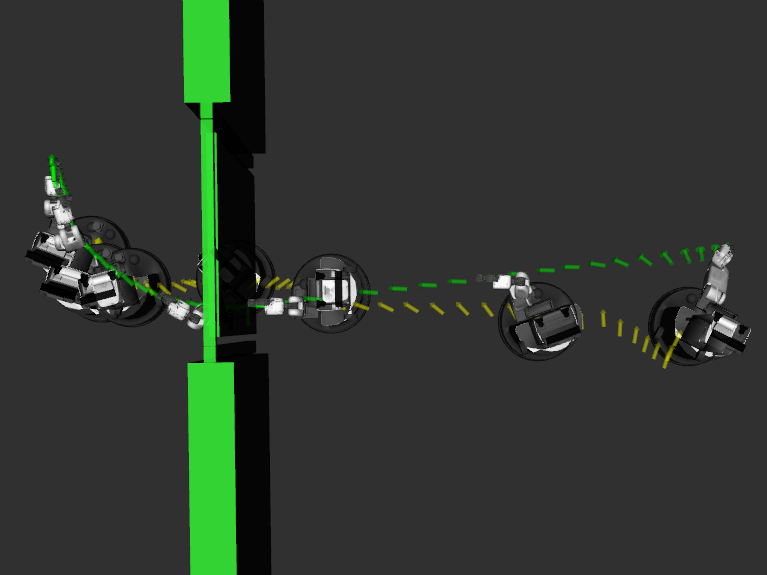} & 
	    \includegraphics[width=0.2\textwidth,trim={0.0cm 0.0cm 0.0cm 0.0cm},clip,angle=0]{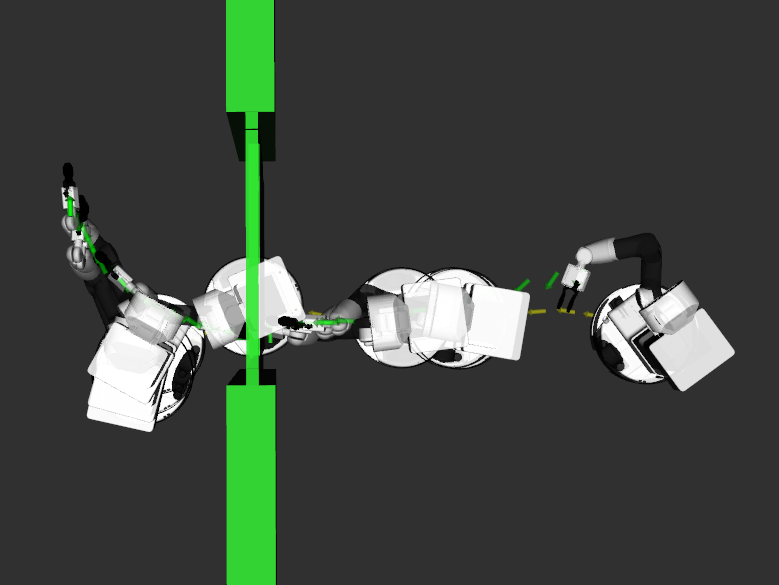}\\
	    \includegraphics[width=0.2\textwidth,trim={0.0cm 0.0cm 0.0cm 0.0cm},clip,angle=0]{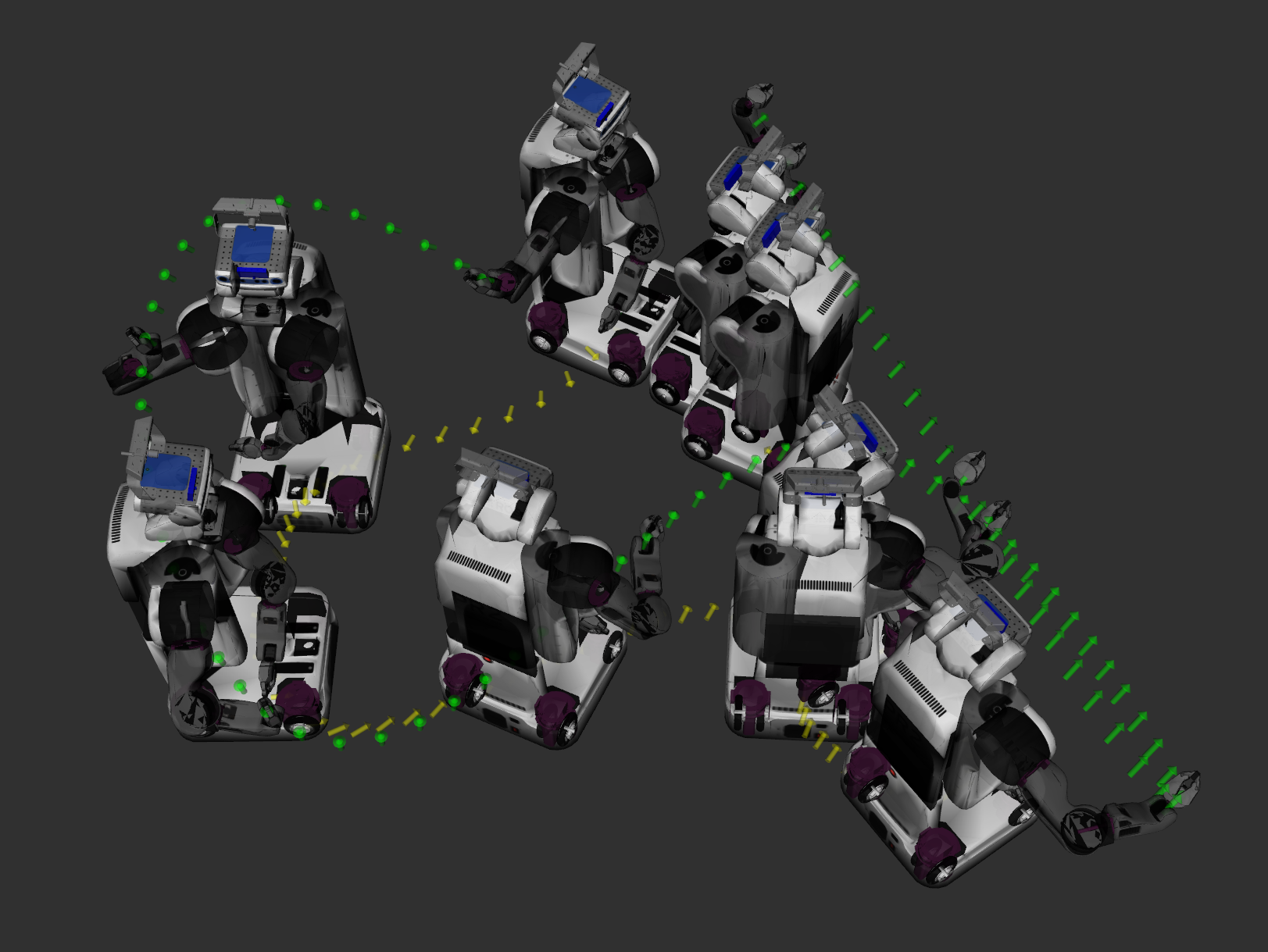} & 
	    \includegraphics[width=0.2\textwidth,trim={0.0cm 0.0cm 0.0cm 0.0cm},clip,angle=0]{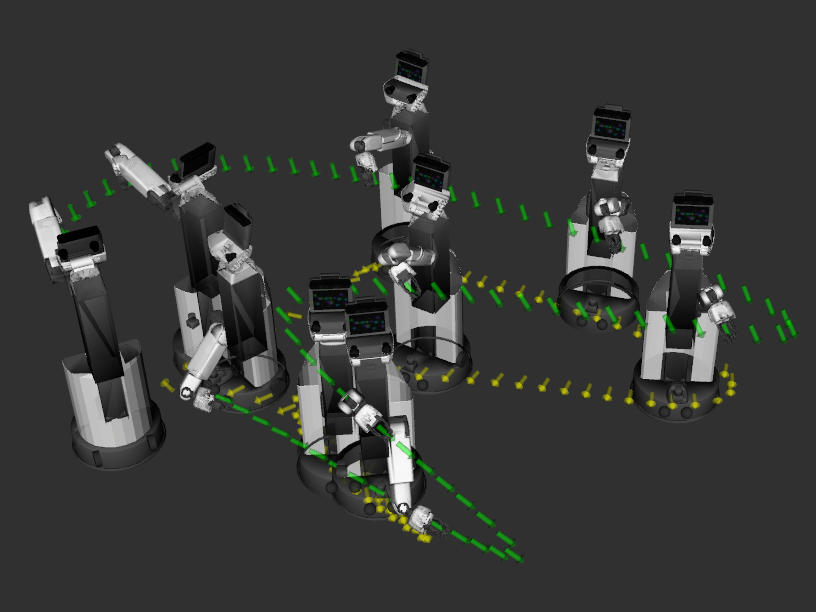} &
	    \includegraphics[width=0.2\textwidth,trim={0.0cm 0.0cm 0.0cm 0.0cm},clip,angle=0]{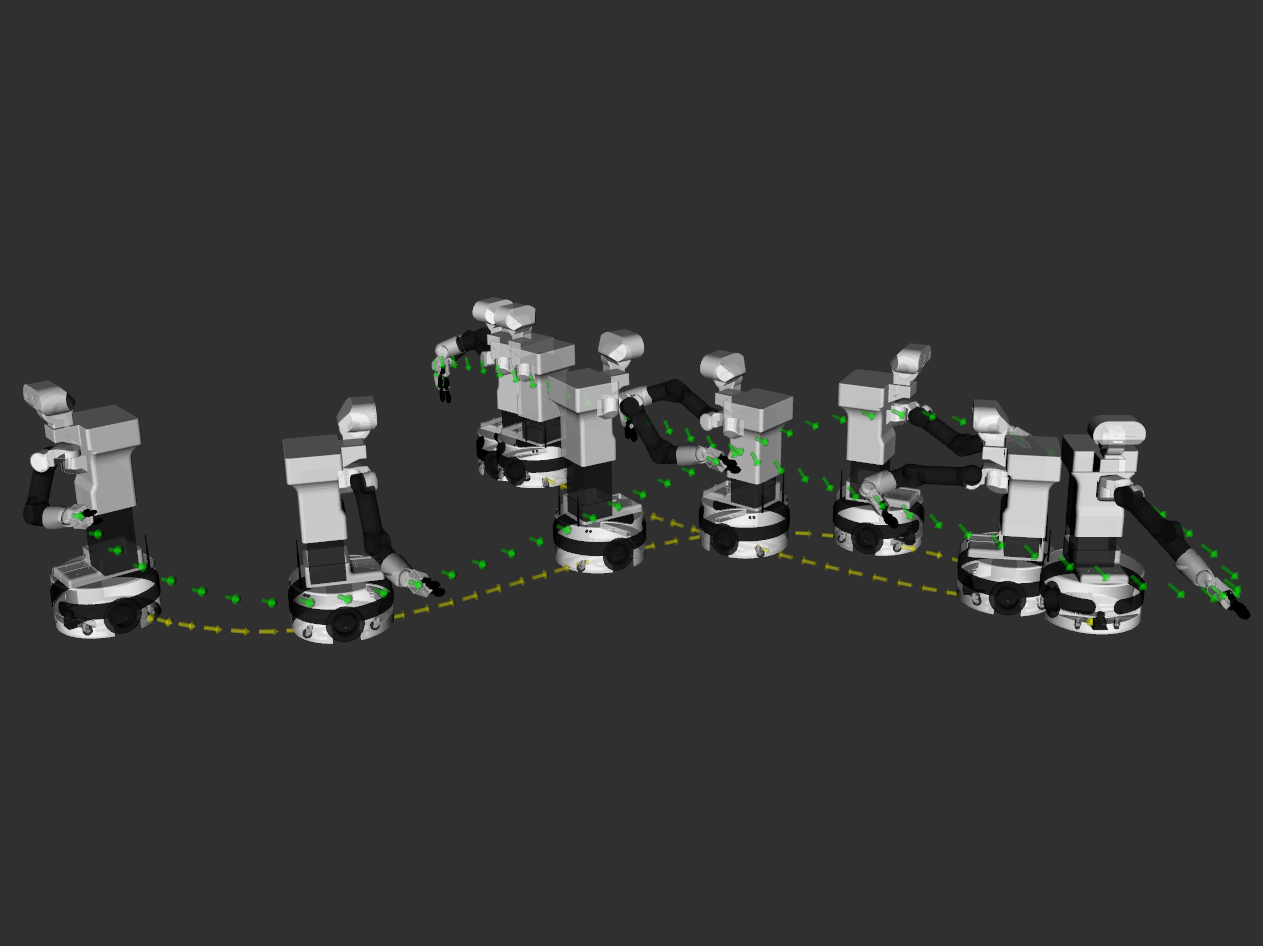}\\
	    \includegraphics[width=0.2\textwidth,trim={0.0cm 0.0cm 0.0cm 0.0cm},clip,angle=0]{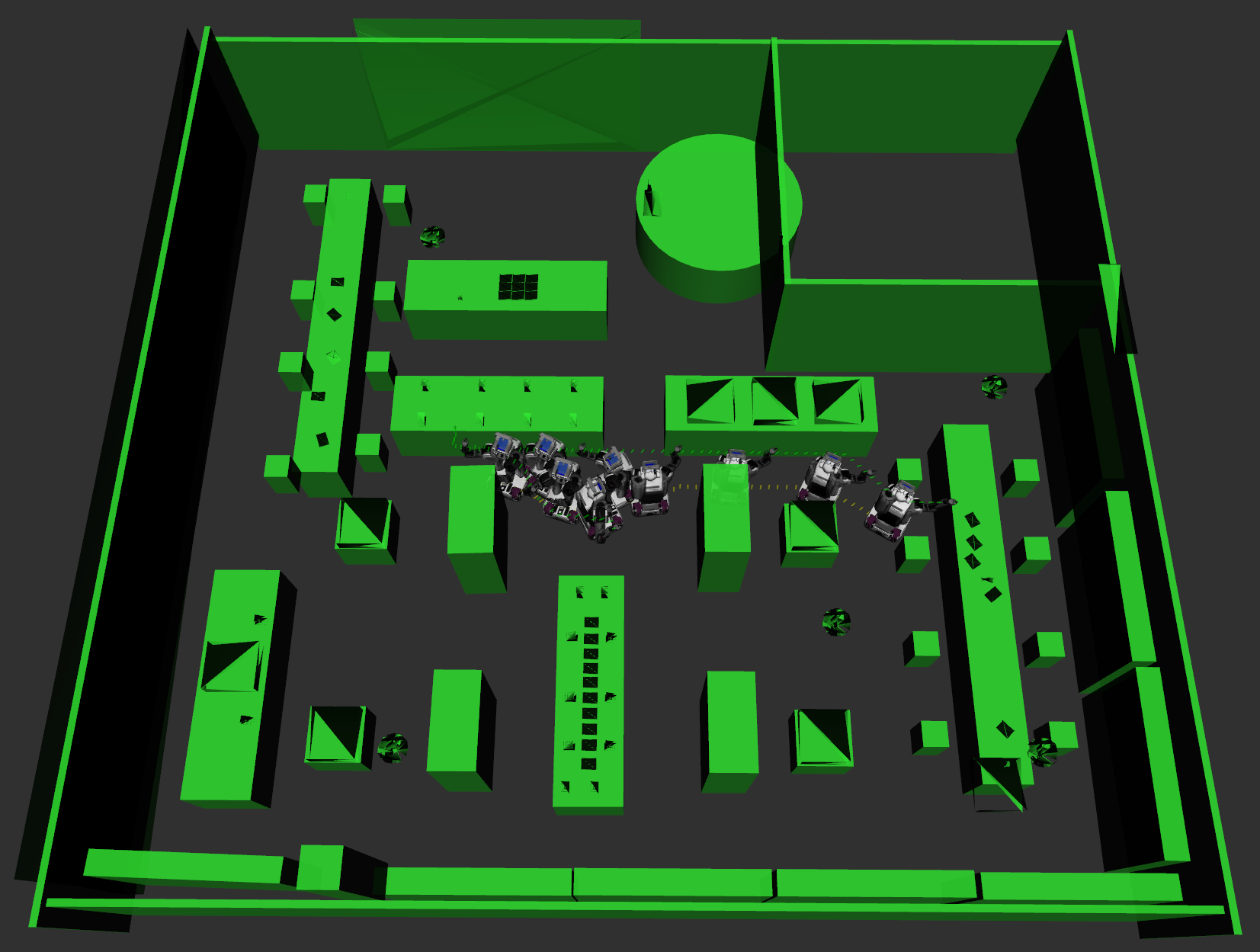} &
	    \includegraphics[width=0.2\textwidth,trim={0.0cm 0.0cm 0.0cm 0.0cm},clip,angle=0]{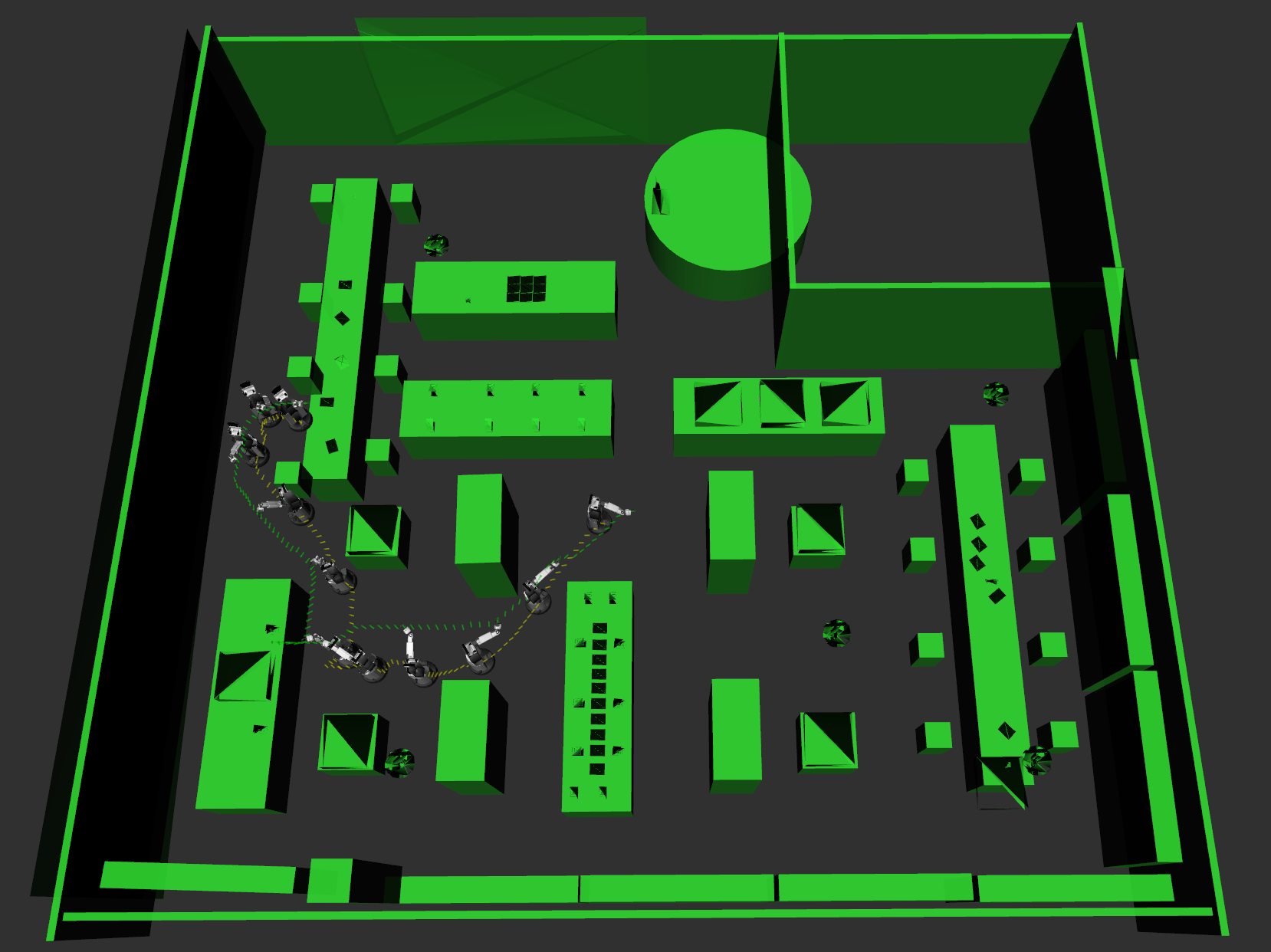} &
	    \includegraphics[width=0.2\textwidth,trim={0.0cm 0.0cm 0.0cm 0.0cm},clip,angle=0]{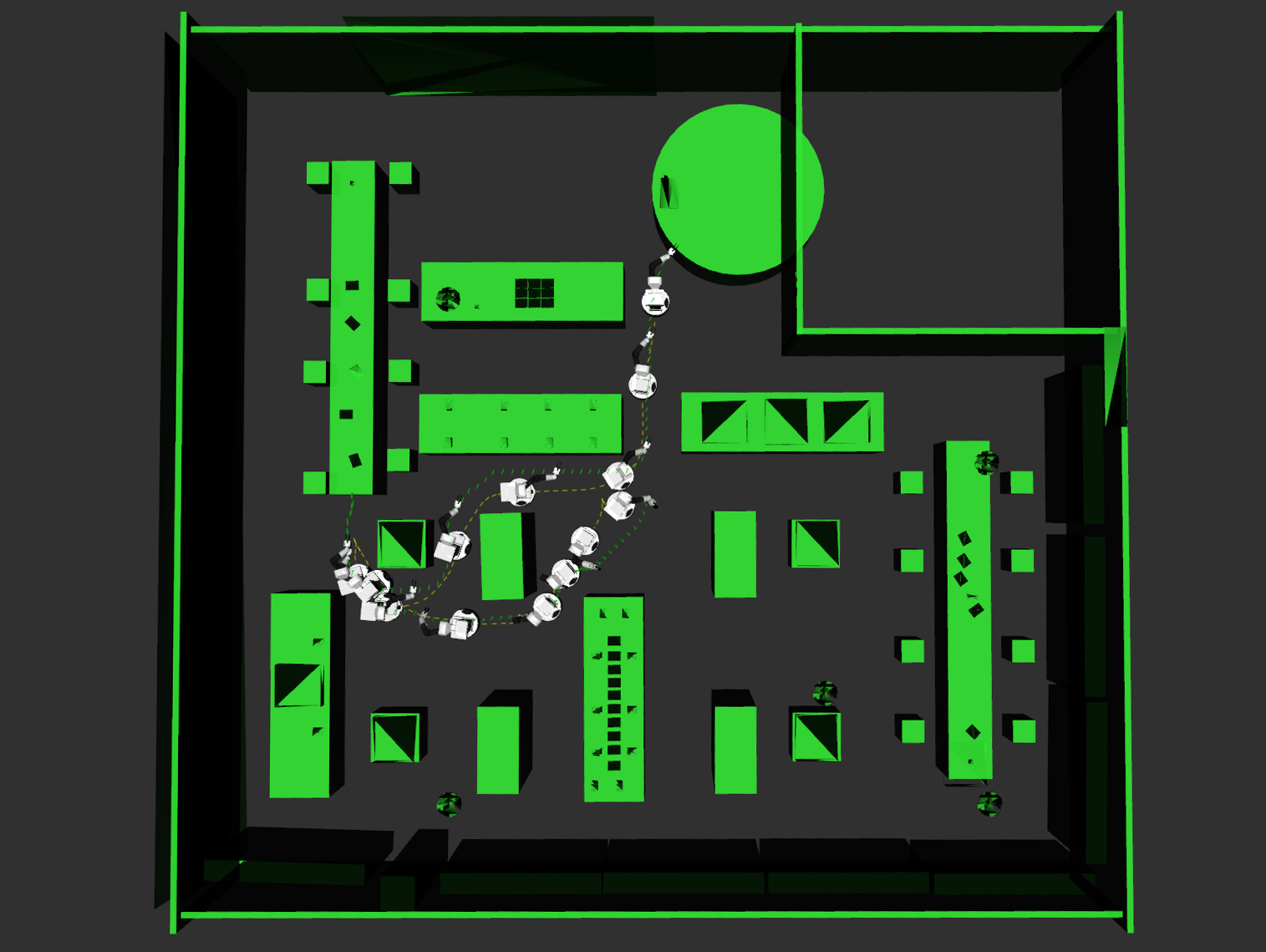}\\
	    \end{tabular}
        }
	\caption{Example motions produced by \ours on the different tasks and robots. The base and ee-trajectories are marked in yellow and green, respectively. From left to right: PR2, HSR, TIAGo. From top to bottom: p\&p, cabinet, drawer, door, spline and bookstore. }
  	\label{fig:trajectories}
\end{figure*}
\setlength{\tabcolsep}{6pt}
\renewcommand{\arraystretch}{1}
\begin{figure*}
    \footnotesize
	\centering
  		\includegraphics[width=0.75\textwidth,trim={0.0cm 0.0cm 0.0cm 0.0cm},clip,angle =0]{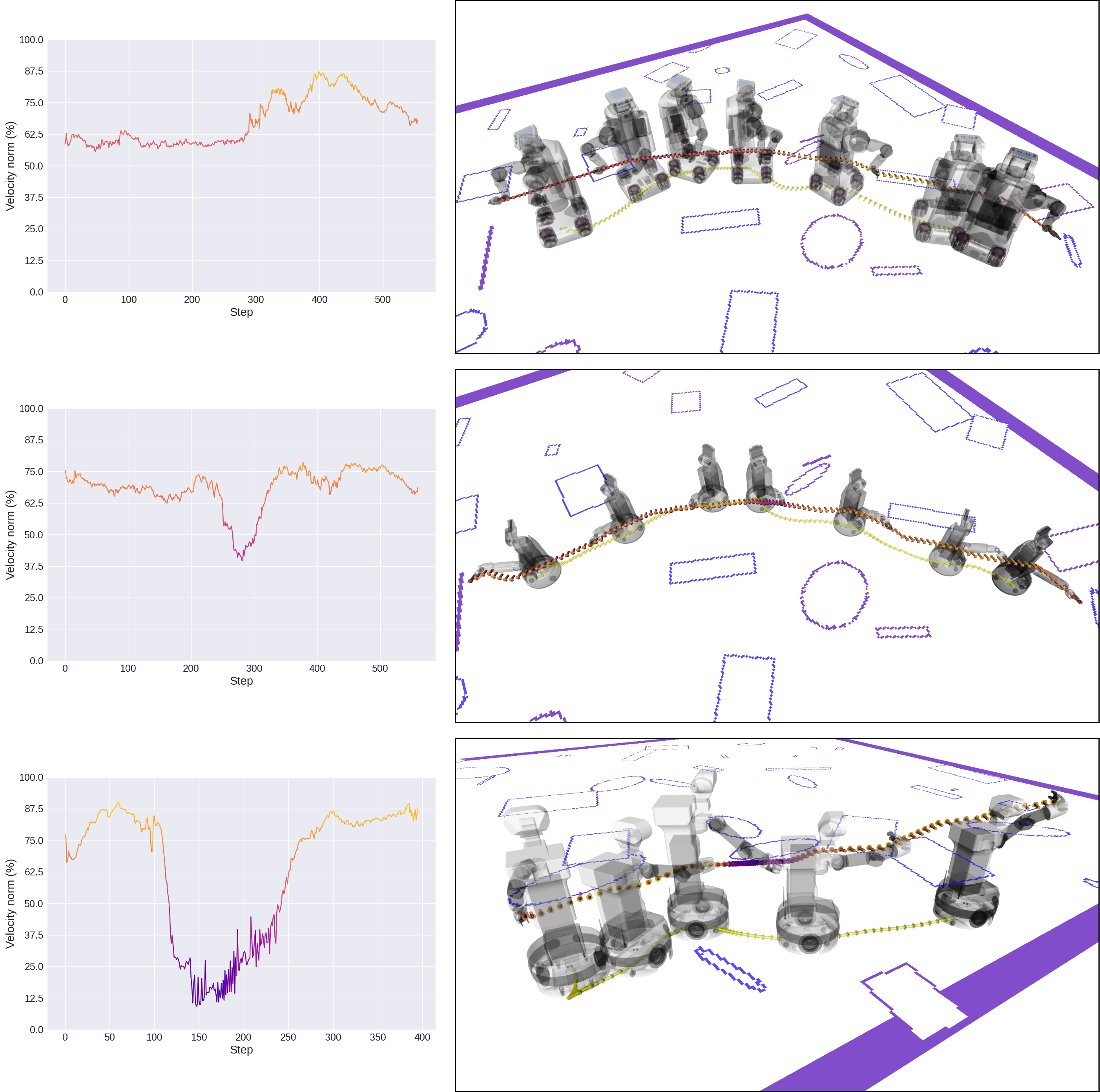}
	\caption{\myworries{Analysis of the learned norm of the end-effector velocities in the rnd obstacle task. Left: velocity norm \myworriestwo{as percentage of the max. allowed velocity} over environment steps. Right: corresponding trajectory. Marker colors indicate the norm of the end-effector velocity in the same scale as the plots on the left. From top to bottom: PR2, HSR, TIAGo.}}
  	\label{fig:vel_norm}
\end{figure*}

With these motions, we extend the suite of mobile manipulation tasks introduced by \cite{honerkamp2021learning}, while also removing all restrictions on the initial start pose. The task objects are situated in a virtual room and the robot spawns in a random pose and configuration within an area in the center of the room. To evaluate the obstacle avoidance capabilities, we place random obstacles in the path of the agent. We allow the end-effector motions to directly pass over these obstacles, leading to challenging poses and decisions on how to position the base. The task setup and start area are shown in \figref{fig:train_task}.

We first evaluate the agents in the analytical environment, abstracting from low-level execution and sensors. The results are shown in \tabref{tab:analytical}. MPC solves a significant amount of episodes across all the robots. In particular, it performs very well on the obstacle-free spline task\myworriestwo{, but} it frequently gets stuck around obstacles, only making progress once time progresses and the desired end-effector pose moves far enough away for the tracking cost term to dominate. This results in a violation of one of the constraints: either large deviations from the desired pose, base collisions, or joint limits as it is unable to trade them off optimally in these situations. Due to the difficulties in balancing these soft constraints, in the door opening task, the MPC controller frequently drives over the door frame with the PR2. The E2E model is able to solve a reasonable number of tasks on the HSR. \myworriestwo{However,} as the action space increases with the 7-DoF arms of the PR2 and TIAGo, it is no longer able to succeed in any of the tasks. This matches the fact that recent reinforcement learning-based mobile manipulation approaches have to restrict the action space to learn. In contrast, our hybrid approach scales effectively with the configuration space of the robot and is able to use the full flexibility of the PR2, achieving the best results on this platform with far above 90\% success on many tasks. We achieve similar success rates on the HSR, demonstrating the agent's ability to use the base to complement its limited arm. The most difficult platform is the TIAGo. While our approach does very well on p\&p and the imitation learning tasks, performance is significantly lower on the rnd obstacle and spline tasks. We hypothesize that the differential drive induces a much harder exploration problem. In particular, it is much harder to move from one mode of operation to another: e.g. in front of an obstacle, the base has to decide to either turn left or right. Once it has made a decision, it cannot easily change the path without violating the end-effector motion.  As such it becomes much harder to escape local optima in the behavior policy. To better learn long-horizon policies for this robot, we start training with a lower base control frequency of \SI{0.0125}{\hertz} and then linearly increase it to \SI{10}{\hertz} over the course of training. Across all tasks and robots, our approach significantly outperforms the other methods.
Also note that, given the procedural generation of the environment, it may not be possible to achieve a perfect score on the rnd obstacle task, as some end-effector motions may not be possible to fulfill kinematically for some of the robots.

Qualitatively, we find that the agent produces reasonable behavior such as seeking robust poses in the center of its workspace or moving backward until reaching an open space to turn. Examples are shown in \figref{fig:trajectories} and in the accompanying video.
\myworries{\figref{fig:vel_norm} shows examples of the learned end-effector velocity norm. We find that the policy learns more complex strategies than simply reducing velocities when close to an obstacle and varies its behavior with the obstacles, the current configuration of the robot, and the desired next end-effector poses. \myworriestwo{The bottom subfigure shows the TIAGo agent strongly reduce the velocity as the end-effector passes over an obstacle while the base has to take a much longer way around the left of the obstacle.} 
Furthermore, the changes in velocity are consistent over time and do not immediately jump from one value to another and the agent learns to use the full range of velocities, not only the extrema of the available range.}


\subsection{Ablation: Individual Components}
\begin{table*}
    \centering
    \caption{Success rates for experiments in the analytical environment \textit{without} obstacles.}
    \begin{threeparttable}
    \begin{tabularx}{\textwidth}{cl|YY|YY|YYYY|YY|Y}
    \toprule
           & & \multicolumn{2}{c|}{Linear} & \multicolumn{2}{c|}{$A^*$-slerp} & \multicolumn{4}{c|}{Imitation} & \multicolumn{2}{c|}{Spline} & \multicolumn{1}{c}{Avg}\\
    \cmidrule{3-13}
      & Agent & \multicolumn{2}{c|}{rnd goal} & \multicolumn{2}{c|}{p\&p}  & \multicolumn{2}{c}{cabinet} & \multicolumn{2}{c|}{drawer}  & \multicolumn{2}{c|}{spline} & \\
      &       & success & w/jumps & success & w/jumps & success & w/jumps & success & w/jumps & success & w/jumps & success \\
    \midrule
        \parbox[t]{1mm}{\multirow{5}{*}{\rotatebox[origin=c]{90}{PR2}}}
        & LKF                        &  ~0          & 80   & \phantom{00}0           & \phantom{0}46 &   ~0          & ~86                & \phantom{00}0          & ~64 & ~0           & 42 & ~0.0\\ 
        & + bioik, l2, jumps         &  38          & 94   & \phantom{00}8           & \phantom{0}22 &  ~36          & ~82                & \phantom{0}22          & ~68 & ~0           & 92 & 20.8 \\
        & + learn-torso              &  \textbf{92} & 92   & \phantom{0}50           & \phantom{0}50 &  ~72          & ~74                & \phantom{0}74          & ~74 & 82           & 82 & 74.0 \\
        & + ee-velocity              &  90          & 90   & \phantom{0}72           & \phantom{0}72 &  ~82          & ~82                & \phantom{0}98          & ~98 & 66           & 66 & 81.6 \\
        & + collision: \textbf{\ours}&  88          & 88   & \textbf{100}            & 100           &  ~98          & ~98                & \textbf{100}           & 100 & \textbf{72}  & 72 & \textbf{91.6} \\
        \cmidrule{1-13}
        \parbox[t]{1mm}{\multirow{5}{*}{\rotatebox[origin=c]{90}{HSR}}}
        & LKF                        & \textbf{72} & 74 & \phantom{0}90          & \phantom{0}92     & 76            & 84 & 80          & \phantom{0}84          & \phantom{0}64          & 80 & 76.4 \\
        & + bioik, l2, jumps         & 66          & 72 & \phantom{0}94          & \phantom{0}96     & 82            & 92 & 78          & \phantom{0}84          & \phantom{0}82          & 92 & 80.4 \\
        & + ee-velocity              & \textbf{72} & 74 & \phantom{0}90          & \phantom{0}92     & 90            & 92 & 86          & \phantom{0}92          & \phantom{0}80          & 94 & 83.6 \\
        & + collision: \textbf{\ours}& \textbf{72} & 72 & \textbf{\phantom{0}98} & \phantom{0}98     & \textbf{98}   & 98 & \textbf{92} & \phantom{0}96          & \phantom{0}\textbf{86} & 96 & \textbf{89.2} \\
        \cmidrule{1-13}
        \parbox[t]{1mm}{\multirow{5}{*}{\rotatebox[origin=c]{90}{Tiago}}}
        & LKF                        & 14          & 50 & \phantom{00}4          & \phantom{0}76     & ~4            & 80 & ~0           & \phantom{0} 72        & \phantom{00}0          & ~4 & ~4.4 \\
        & + bioik, l2, jumps         & 48          & 76 & \phantom{0}42          & \phantom{0}74     & 36            & 92 & 44           & \phantom{0} 92        & \phantom{0}20          & 70 & 38.0 \\
        & + learn-torso              & 64          & 64 & \textbf{\phantom{0}96} & \phantom{0}96     & 80            & 80 & 90           & \phantom{0} 90        & \phantom{0}48          & 48 & 75.6 \\
        & + ee-velocity              & 66          & 66 & \phantom{0}86          & \phantom{0}86     & \textbf{94}   & 94 & \textbf{92}  & \phantom{0} 92        & \phantom{0}50          & 50 & 77.6 \\
        & + collision: \textbf{\ours}& \textbf{76} & 76 & \phantom{0}90          & \phantom{0}90     & 88            & 88 & 86           & \phantom{0} 86        & \phantom{0}\textbf{56} & 56 & \textbf{79.2} \\
    \bottomrule
    \end{tabularx}
  \begin{tablenotes}[para,flushleft]
       \footnotesize      
       Notes: LKF denotes the previous Learning Kinematic Feasibility approach \cite{honerkamp2021learning}. Each row adds parts of our contributions to the row above it until we arrive at our proposed model. \textit{w/jumps} denotes success rates ignoring configuration jumps. Evaluated over a single seed.
     \end{tablenotes}
   \end{threeparttable}
    \label{tab:lkf_comparison}
\end{table*}

To evaluate the impact of our contributions, we perform extensive comparisons with the original LKF approach. As it cannot avoid obstacles, we train these approaches on a random goal reaching task in an empty map and compare the ability to follow the motions on a subset of the previous tasks \textit{without obstacles}. Note that results are not directly comparable to those reported in \cite{honerkamp2021learning}, as our task definitions are generally more challenging as they remove restrictions on initial poses and include the torso joint of the PR2. We report both the same success rates as before as well as a success rate that does not penalize configuration jumps as used in \cite{honerkamp2021learning}, which we label \textit{w/jumps}.

The results are shown in \tabref{tab:lkf_comparison}. Row by row we add our contributions. We first switch to the BioIk solver with minimal displacement regularisation and the generalized L2 kinematic feasibility reward function. This largely increases the success rate by removing most configuration jumps. For the PR2 and TIAGo, we then additionally learn the torso lift velocities, which have a similarly large impact on the success rate. This is because most jumps occurred in the torso lift joints which move much slower than the arm joints. Furthermore, for many goals, it is important to plan ahead and to start moving the torso early on in the episode to achieve large height differences.
Next, we also learn to control the velocity of the end-effector motions. This slightly increases success rates, even in the absence of obstacles. We found the impact of this term even more important in the presence of obstacles as the base has to maneuver much more. Lastly, we add the occupancy map and train in the new procedurally generated training task. With this we arrive at our approach, which achieves the best average performance on all robots, highlighting the importance of all components. We hypothesize that this last increase stems from the fact that the new obstacle task largely increases the difficulty and variety of the motions during training. \myworries{This demonstrates the importance of each individual contribution to arrive at a method that largely increases success rates and robustness over previous work.}

\subsection{Scaling to Human-Centered Maps}

To further evaluate the generalization to complex objects with unseen geometries and the ability to navigate narrow human-centered environments, we evaluate the agents in a realistic bookstore environment~\cite{awsbookstore} shown in \figref{fig:train_task}. We define a set of ten feasible object locations across the whole map and draw random pairs of locations to pick up an object from and place it down. The robot starts in the center of the map, again in a random start configuration. The resulting tasks are very long horizon, taking on average 1,300-1,500 steps at a control frequency of \SI{10}{\hertz} and require passing through passages as narrow as \SI{0.85}{\meter}. The results are reported in \tabref{tab:analytical}. Our approach is the only method to successfully complete this task on all robots. While success rates are lower than in p\&p, both the PR2 and HSR succeed in the majority of episodes. Success rates are somewhat lower on the TIAGo with 42.0\%. Difficulties of this task are on one hand the sheer length of each task. On the other hand, the robot has to repeatedly pass through very narrow passages. This is particularly challenging as the agent at the same time has to strictly adhere to the end-effector motions. As our focus is the learned base behavior, we do not adapt the end-effector orientations to the environment, even though the pick \& place task does not require specific motions while carrying the object. This highlights a large potential to further improve the success rates by learning or adapting the end-effector motions to the specific task. We leave this for future work. 



\subsection{Dynamic Obstacles}

To evaluate the reactiveness of the trained agents, we construct two tasks with dynamic obstacles. The obstacles move with a random velocity between \SI{0.1}{\meter\per\second} and \SI{0.15}{\meter\per\second}. We re-plan the end-effector motion at every time step in MPC-fashion, which easily runs in real-time as all the complexities are shifted to the RL agent. This is again a zero-shot transfer: the agent has never seen moving obstacles during training. In the \textit{dynamic obstacles} task, we uniformly spawn 16 dynamic obstacles in an empty room and draw a random goal for the end-effector, as shown in \figref{fig:train_task}.

We exclude goals on the very edges of the robot's workspace where its maneuverability is very low, see "Restr. height" in \tabref{tab:constraints}. The \textit{dynamic p\&p} task uses the same setup as p\&p, but we replace the static obstacles with three dynamic obstacles. The results are included in \tabref{tab:analytical}. We find that our agent is able to very quickly react to these obstacles. \myworries{We attribute this capability to the fact that the agent has learned a robust policy that prefers to keep some distance to obstacles whenever this is possible. As a result, it tries to do the same whenever a dynamic obstacle approaches. This robustness may be a result of SAC's entropy maximization and the action noise that we apply during training.} Success rates for the PR2 and HSR are close to those of static obstacle tasks. The TIAGo has more difficulties: the differential drive does not allow it to easily move away from obstacles that move into the side of its base. As it has never seen dynamic obstacles, it has no incentive to anticipate these situations and to proactively place its base accordingly. Additional fine-tuning on dynamic obstacles might be able to induce it with such a sense of foresight. While the MPC approach is able to react to dynamic obstacles, its success rate drops further down to 30-64\% versus the 50.4-88\% of our approach.


\subsection{Comparison to Planners}
\begin{table*}
    \centering
    \caption{Evaluation of the planner baselines.}
    \begin{threeparttable}
    \begin{tabularx}{\textwidth}{cl|YYY|YYY|YYY}
    \toprule
      &       & \multicolumn{3}{c|}{rnd obstacle} & \multicolumn{3}{c|}{p\&p} & \multicolumn{3}{c}{bookstore p\&p} \\
      \cmidrule{3-11}
      & Agent & success & planning time & rel ee path & success & planning time & rel ee path & success & planning time & rel ee path \\
    \midrule
        \parbox[t]{1mm}{\multirow{4}{*}{\rotatebox[origin=c]{90}{PR2}}}
        & RRTConnect               & 74.0          & ~3.1 sec & 1.25 & ~78.0           & ~0.9 sec & 1.04 & 22.0           & 119.6 sec & 0.97 \\
        & RRTConnect (upright)     & --            & --       & --   & ~72.0           & ~3.3 sec & 0.98 & 22.0           & 124.6 sec & 0.97 \\
        & Bi2RRT*                  & 62.5          & 41.5 sec & 1.38 & ~84.0           & ~8.6 sec & 1.17 & ~8.0           & 115.3 sec & 1.12 \\
        & \textbf{\ours}            & \textbf{86.0} & ~0.1 sec & 1.00 & \textbf{~97.6}  & ~0.1 sec & 1.00 & \textbf{70.0}  & \phantom{00}0.1 sec  & 1.00 \\
    \midrule
        \parbox[t]{1mm}{\multirow{3}{*}{\rotatebox[origin=c]{90}{HSR}}}
        & RRTConnect               & \textbf{82.0} & 0.5 sec  & 0.83 & \textbf{100.0}  & ~1.0 sec  & 0.99 & 42.0           & 12.2 sec & 0.89 \\
        & RRTConnect (upright)     & --            & --       & --   & \phantom{00}0.0 & n.d.      & n.d. & ~0.0           & n.d.     & n.d. \\
        & \textbf{\ours}            & 63.2          & ~0.1 sec & 1.00 & \phantom{0}92.0 & ~0.1 sec  & 1.00 & \textbf{68.0}  & \phantom{0}0.1 sec & 1.00 \\
    \bottomrule
    \end{tabularx}
\begin{tablenotes}[para,flushleft]
       \footnotesize      
       Notes: Motions for the planners are unconstrained while \ours has to follow the full end-effector motion. \textit{(upright)} adds a constraint to keep the object upright after picking it up. rel ee path denotes the length of the ee-path relative to \ours. Planning time and rel ee path are only computed over successful episodes.
     \end{tablenotes}
   \end{threeparttable}
    \label{tab:planners}
\end{table*}

Planning-based methods are very general and widely used for manipulation tasks. While the specification of arbitrary motion constraints such as defined by the imitation learning motions is difficult as discussed in \secref{sec:related}, these approaches are a powerful baseline for a subset of our tasks: goal-reaching and pick\&place in static environments. We compare with two sampling-based planners on the rnd obstacle, pick\&place, and bookstore tasks. We remove the end-effector motion constraints and instead solely specify the task with three goal poses: in front of the object, the grasp pose, and the location to place it down. Note that this is a significantly simpler task as the end-effector is completely unconstrained in -etween the goals. In the second evaluation, we incorporate an additional simple pose constraint to hold the object upright after grasping it that we label as \textit{(upright)}.

\tabref{tab:planners} compares the success rates, planning time, and the resulting length of the end-effector trajectory relative to our approach, which still has to fulfill all motion constraints. We find that they perform well in the obstacle and pick\&place tasks, even outperforming our approach on the HSR. At the same time, success rates decrease on the PR2. This can be attributed to the larger joint space and the larger base, making the spaces narrower. In contrast, our approach performs better on the PR2 and is able to make use of its flexibility, indicating good scaling with robot complexity. Already adding the simple upright constraint decreases performance on p\&p by 6ppt for the PR2. The more restricted HSR completely fails to sample successful paths under this constraint. Moving to the complex bookstore map, performance drops significantly below our approach with success rates of 22\% to 42\% versus 70\%. While planning times are quite low on the other tasks, they now increase to almost 2 minutes for the PR2.
Qualitatively, the planners can produce somewhat unnatural paths, such as waving the whole arm  when going from the goal directly in front of the pick object to the pick object. In contrast, our approach seeks out base poses that enable good reactiveness and that are robust to noise in the base movements.

These results highlight two aspects: on one hand the difficulty of the bookstore map, on the other hand, the general applicability of our approach and its beneficial scaling with the size of the configuration space: while having to follow much more restrictive motion constraints, it remains competitive with unrestricted state-of-the-art methods and even outperforms them as both robot and map complexity increase. At the same time, it is directly applicable to dynamic scenes, partially observable environments, and arbitrary end-effector motions as demonstrated in \secref{sec:arbitrary}.


\subsection{Executability}
\begin{table*}
    \centering
    \caption{Success rates for experiments in the gazebo environment.} 
    \begin{threeparttable}
    \begin{tabularx}{\textwidth}{cl|YYYYY|YYY|Y|Y}
    \toprule
           & & \multicolumn{5}{c|}{$A^*$-slerp} & \multicolumn{3}{c|}{Imitation Learning} & \multicolumn{1}{c|}{Spline} & \multicolumn{1}{c}{Avg} \\ 
    \cmidrule{3-12}
      & Agent                & rnd obstacle & p\&p & bookstore p\&p &  dynamic obstacle & dynamic p\&p & cabinet & drawer & door & spline & \\ 
    \midrule
        \parbox[t]{1mm}{\multirow{3}{*}{\rotatebox[origin=c]{90}{PR2}}}
        & MPC                & 26.0   & 12.0 & ~0.0   & 38.0 & ~2.0    & ~0.0 & ~2.0   &     ~0.0 & 18.0 & 10.9          \\ 
        & E2E                &   ~0.0 & ~0.0 &   ~0.0 & ~0.0 &  ~0.0   & ~0.0 &   ~0.0 &     ~0.0 & ~0.0 & ~0.0          \\ 
        & \textbf{\ours}      & 81.6   & 87.6 &   48.8 & 93.2 &  74.4   & 94.0 &   96.8 &     60.4 & 75.6 & \textbf{79.2} \\ 
        \cmidrule{1-12}
        \parbox[t]{1mm}{\multirow{3}{*}{\rotatebox[origin=c]{90}{HSR}}}
        & MPC                & 12.0   & 14.0 &   ~0.0 & 30.0 & ~2.0 & ~0.0 &   ~4.0 &     ~0.0 & 42.0 &              11.6 \\ 
        & E2E                & 29.6   & 27.2 &   19.6 & 27.2 & ~2.8 & 39.2 &   ~8.4 &     ~1.2 & ~7.2 & 18.0          \\ 
        & \textbf{\ours}      & 64.0   & 92.4 &   54.4 & 91.2 & 78.8 & 95.6 &   90.4 &     85.6 & 91.6 & \textbf{82.7} \\ 
        \cmidrule{1-12}
        \parbox[t]{1mm}{\multirow{3}{*}{\rotatebox[origin=c]{90}{Tiago}}}
        & MPC                & n.a.   & n.a. & n.a.   & n.a. & n.a. & n.a. &    n.a.& n.a.     & n.a. & n.a.          \\ 
        & E2E                & ~0.0   & ~0.0 &   ~0.0 & ~0.0 & ~0.0 & ~0.0 &   ~0.0 &     ~0.0 & ~0.0 & ~0.0          \\ 
        & \textbf{\ours}      & 56.0   & 85.2 &   54.8 & 71.2 & 51.2 & 78.8 &   82.4 &     30.0 & 56.0 & \textbf{62.8} \\ 
    \bottomrule
    \end{tabularx}
    \begin{tablenotes}[para,flushleft]
       \footnotesize      
       Notes: Evaluation of zero-shot transfer to the Gazebo physics simulator on unseen tasks from three different motions systems, an $A^*$-based system, an imitation system learned from human demonstrations and spline interpolation of random waypoints. The last column reports the average across all tasks. We evaluate all models on three different robotic platforms, the PR2, HSR, and TIAGo. MPC is not evaluated on the TIAGo as it does not posses velocity controllers.
     \end{tablenotes}
   \end{threeparttable}
    \label{tab:gazebo}
\end{table*}

To evaluate if the produced motions are readily executable, we transfer the learned policies to the Gazebo physics simulator. The agent's actions are sent to the robot's low-level base and arm controllers at an average frequency of around \SI{30}{\hertz} to \SI{70}{\hertz}. We construct an occupancy map by integrating the LiDAR and range sensor data into a binary costmap. For the dynamic obstacle tasks, the end-effector motions are generated based on an additional global costmap built from the robot's sensor data. The MPC baseline relies on velocity controllers for the whole robot. We use default velocity controllers of the PR2 and a pseudo velocity controller of the HSR, which underneath utilizes the position controllers. TIAGo does not provide any velocity controllers for the arm, as such we were unable to evaluate the MPC approach on this robot. While our agent can only rely on its sensors, we still provide the MPC approach with the ground-truth signed distance field to reduce complexity. For the dynamic obstacle tasks and the bookstore map, we inflate the local map passed to the RL agent by \SI{3}{\centi\meter} to allow the agent to more quickly react to the obstacles. For the narrow door opening task, we scale the actions of the PR2 by a factor of 0.5.

The results for all tasks are shown in \tabref{tab:gazebo}. In contrast to the analytical environment, the agents have to generalize to unseen physics and low-level controllers, asynchronous execution as well as noisy and partial occupancy maps. Furthermore, any arm collision is now recorded as a failure. Nonetheless, the performance of our approach closely matches the analytical environment with small drops of 9ppt, 0.3ppt, and 2.6ppt in average performance. Looking more closely, the differences stem almost exclusively from the door opening (PR2, TIAGo) and the bookstore (PR2) tasks. In the door opening task we find that the imprecisions induced by the low-level controllers and asynchronous execution cause the PR2 to slightly touch the door frame in a number of episodes. As this is a very high-precision task for its large base, unseen physics and accelerations can quickly have an impact. For most episodes, success depends on only a few centimeters. Failures on the TIAGo stem largely from collisions between the elbow and the door frame. The robot commonly approaches the door with its arm in one of two configurations. In one of these configurations, the elbow is unable to avoid the door frame, leading to several failures. As the agent does not take into account arm collisions during training, the learned policy has no means yet to avoid this failure mode. We leave full 3D-collision avoidance to future work. Similar to the door task, the drop in performance in the bookstore map for the PR2 can be attributed to a larger number of collisions in narrow pathways due to the differences in physics and accelerations compared to the training environment as well as a small number of arm collisions with bookshelves. Overall, performance remains high across the large majority of tasks. This is in contrast with the MPC approach which is significantly impacted by these factors. The approach struggles with unseen physics and imperfect execution of motions, leading to collisions and large deviations from the desired motions on both robots. Once the error terms become too large, the approach is unable to recover good behavior, resulting in large and abrupt constraint violations.


\subsection{Real-World Experiments}

\setlength{\tabcolsep}{1pt}
\renewcommand{\arraystretch}{1}
\begin{figure*}
	\centering
	\footnotesize
	{\setlength{\fboxsep}{0pt}%
  \setlength{\fboxrule}{0pt}%
	\resizebox*{!}{.9\textheight}{%
  	\begin{tabular}{ p{1.5cm}cc }
  	    & PR2 & HSR \\
  		Floorplan & \fbox{\includegraphics[width=0.5\columnwidth,trim={0cm 0cm 0cm 0cm},clip,angle =0,valign=c]{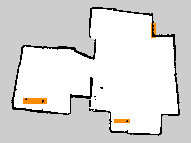}} &
  		\fbox{\includegraphics[width=0.5\columnwidth,trim={0cm 0cm 0cm 0cm},clip,angle =0,valign=c]{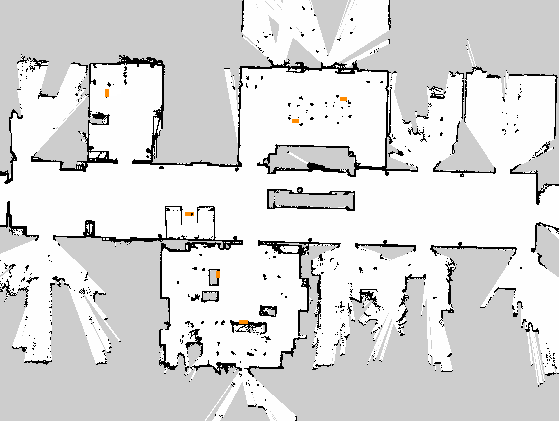}} \\
  		Dynamic P\&P & \fbox{\includegraphics[width=0.5\columnwidth,trim={0cm 0cm 0cm 0cm},clip,angle =0,valign=c]{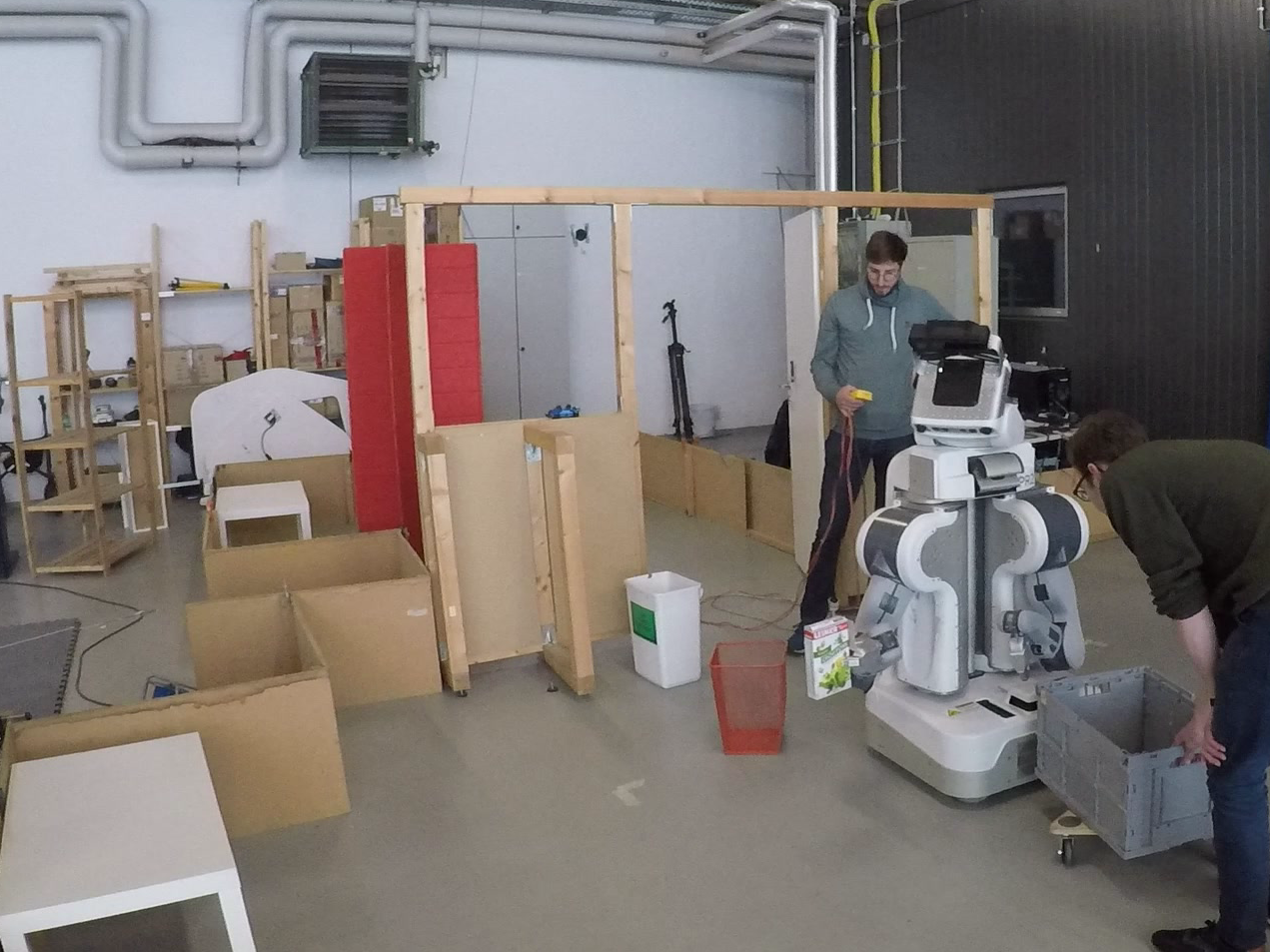}} & \fbox{\includegraphics[width=0.5\columnwidth,trim={0cm 0cm 0cm 0cm},clip,angle =0,valign=c]{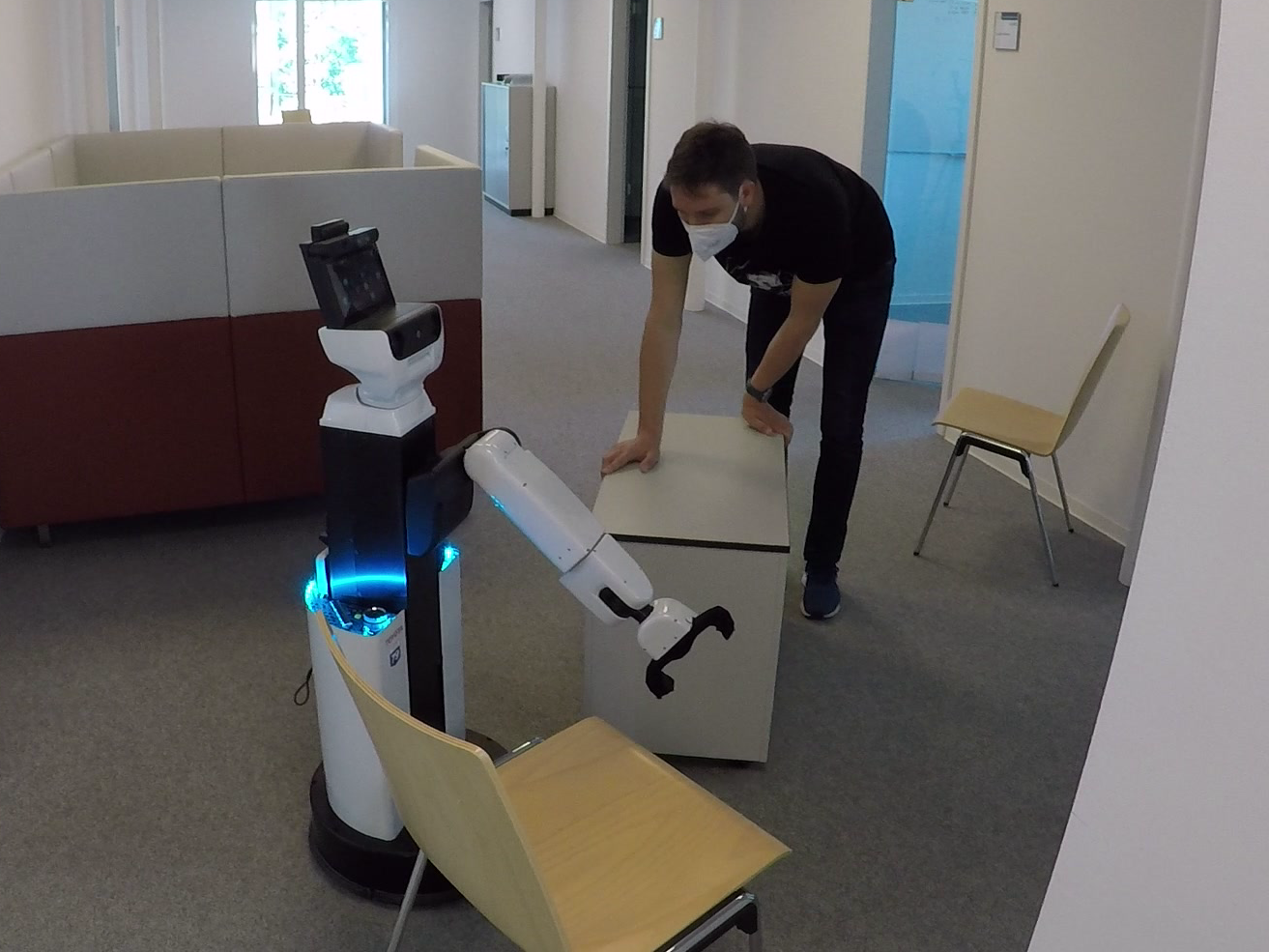}}\\
		Door & \fbox{\includegraphics[width=0.5\columnwidth,trim={0cm 0cm 0cm 0cm},clip,angle =0,valign=c]{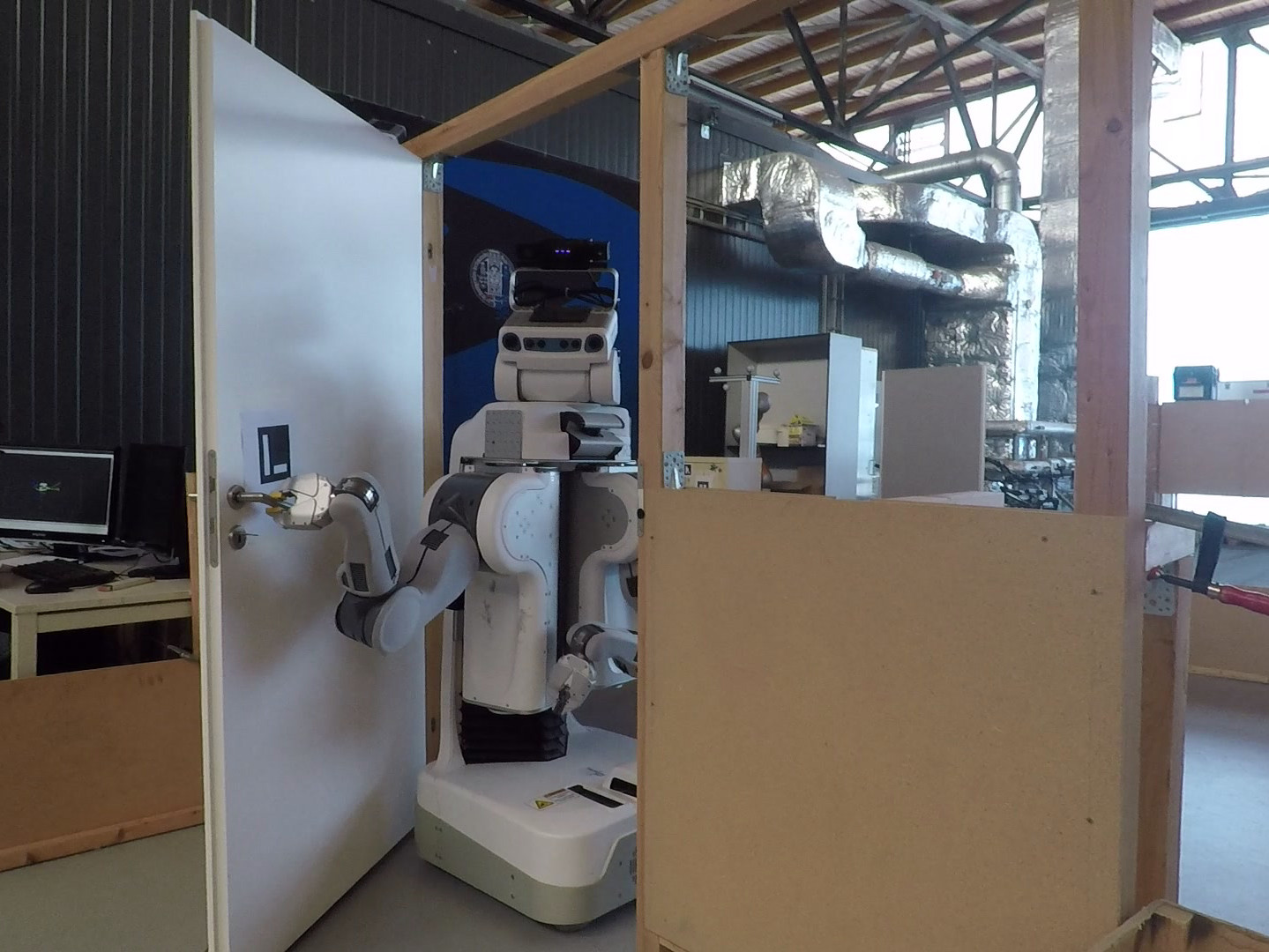}} &
		  \fbox{\includegraphics[width=0.5\columnwidth,trim={0cm 0cm 0cm 0cm},clip,angle =0,valign=c]{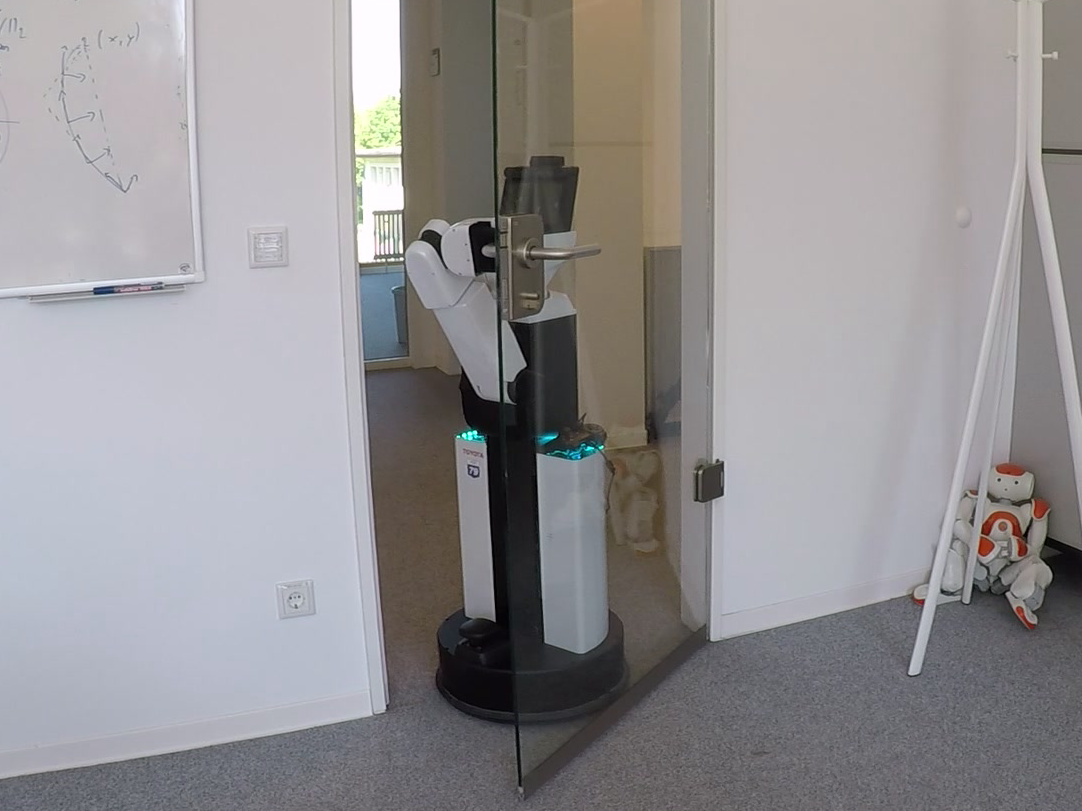}}\\
  		Cabinet & \fbox{\includegraphics[width=0.5\columnwidth,trim={0cm 0cm 0cm 0cm},clip,angle =0,valign=c]{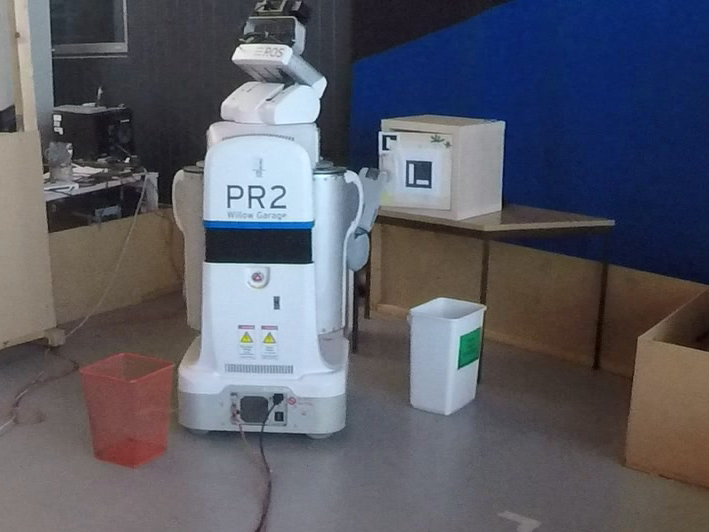}} & \fbox{\includegraphics[width=0.5\columnwidth,trim={0cm 0cm 0cm 0cm},clip,angle =0,valign=c]{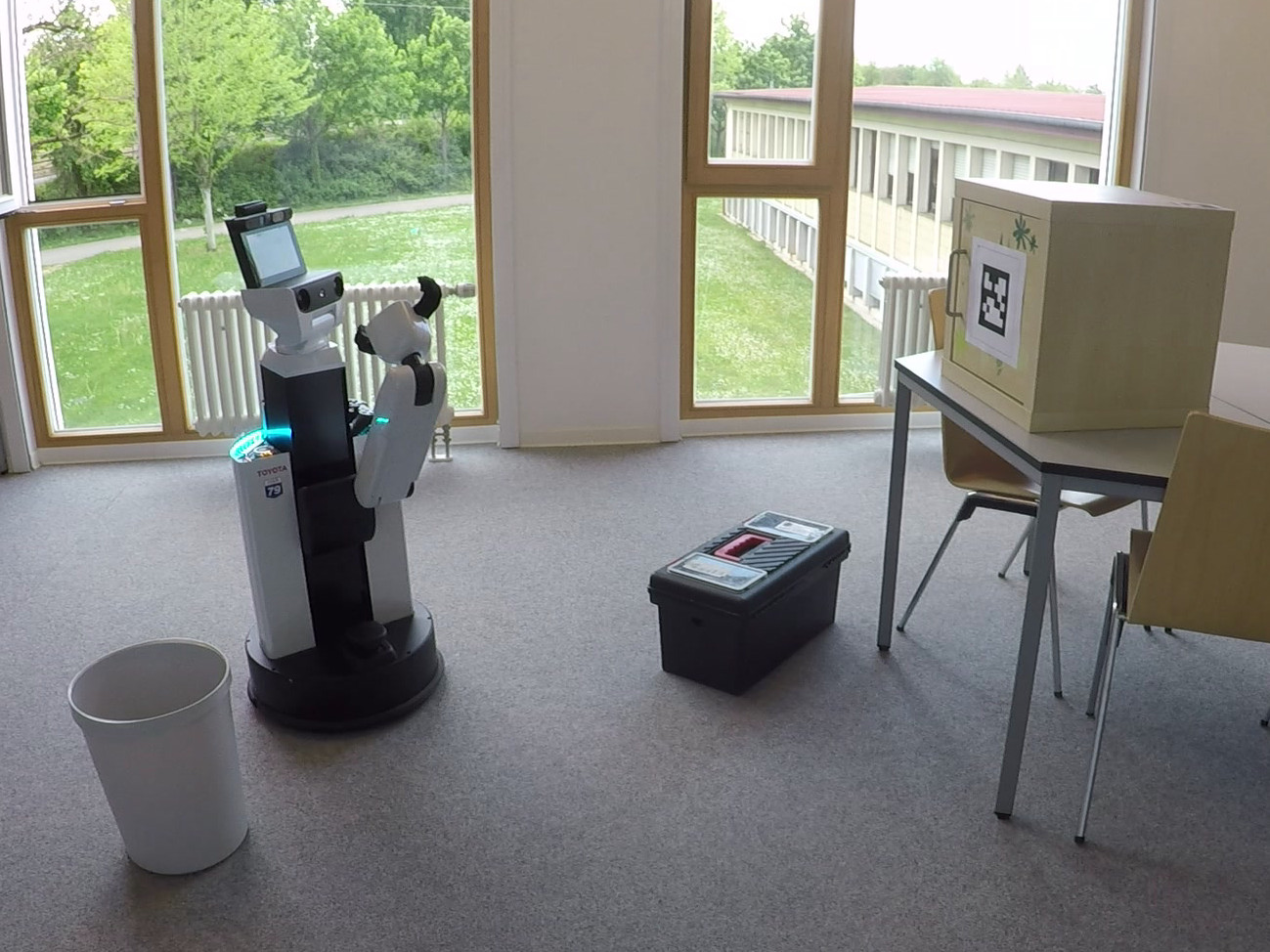}}\\
  		Drawer & \fbox{\includegraphics[width=0.5\columnwidth,trim={0cm 0cm 0cm 0cm},clip,angle =0,valign=c]{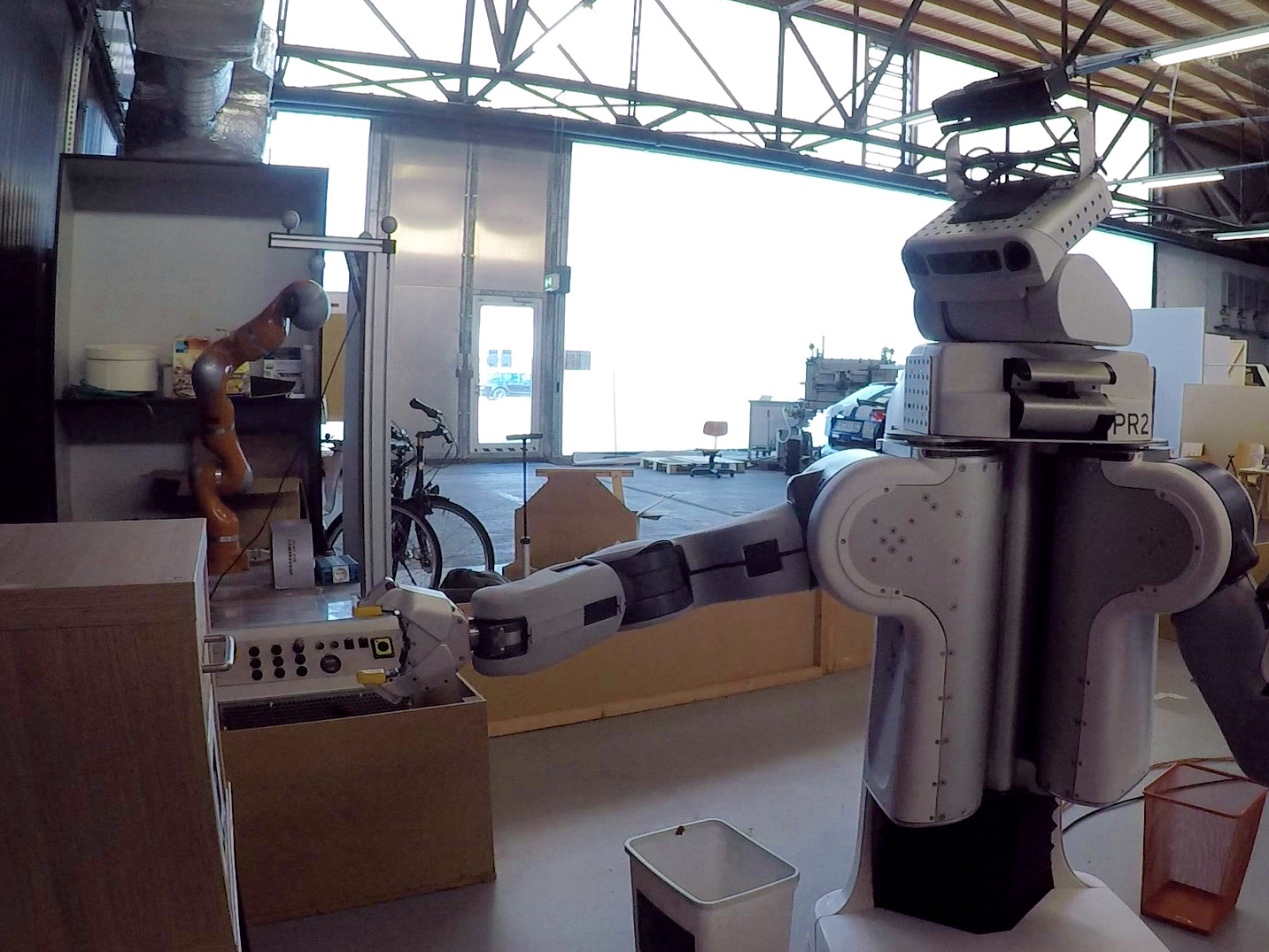}} & \fbox{\includegraphics[width=0.5\columnwidth,trim={0cm 0cm 0cm 0cm},clip,angle =0,valign=c]{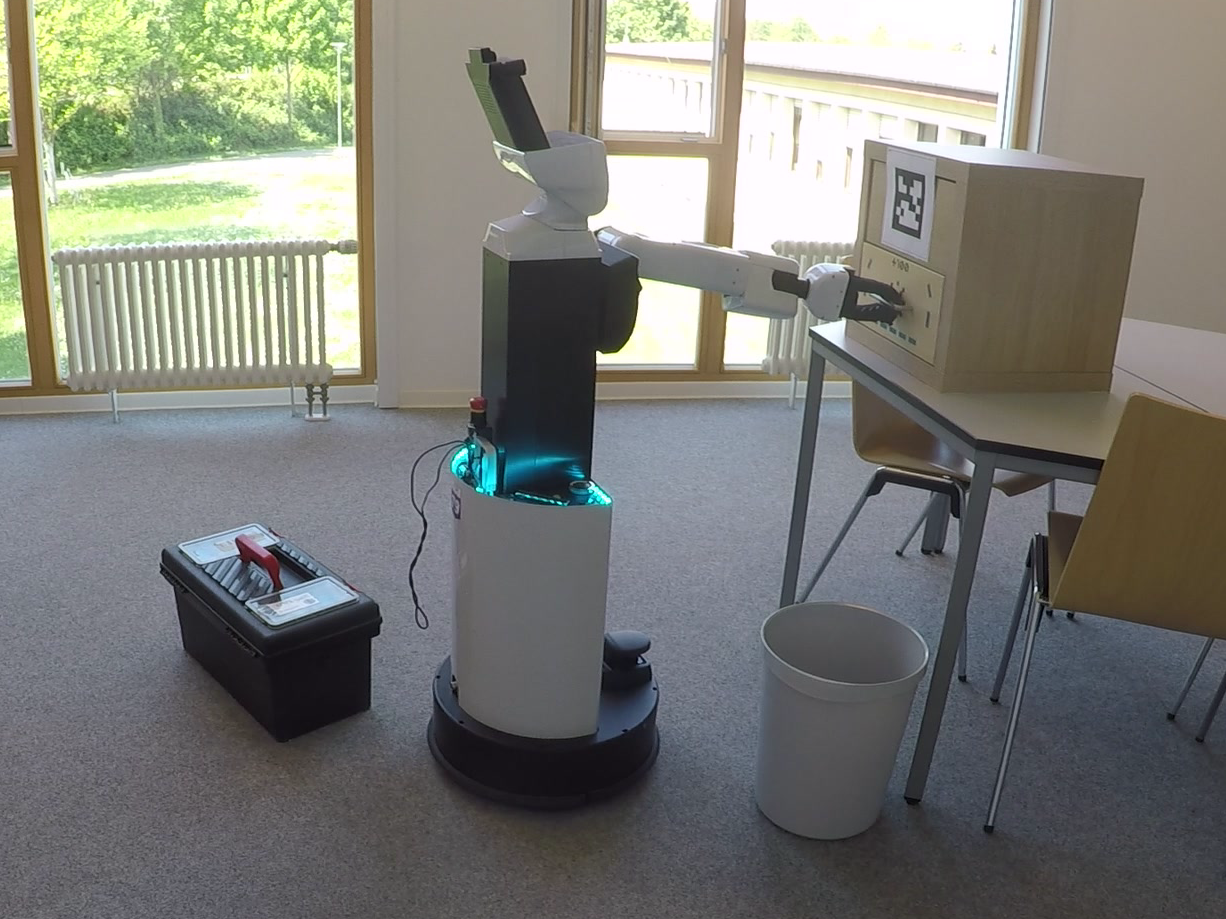}}\\
	\end{tabular}
	}}
	\caption{Real world experiments on the PR2 (left) and HSR (right) robots. From top to bottom: environment map, pick\&place while avoiding static and dynamic obstacles (table locations marked in orange), opening and driving through a door, opening a cabinet and opening a drawer with static obstacles constraining the base.}
  	\label{fig:real-world}
\end{figure*}
\setlength{\tabcolsep}{6pt}
\renewcommand{\arraystretch}{1}

We transfer the agents to the real world and evaluate them on the HSR and PR2 robots. For the PR2, we construct an environment consisting of two rooms connected by a door in the hall it is located in. We directly evaluate the HSR in our office building, a common human-centered environment not adapted to the robot's capabilities. Maps of both environments are shown in \figref{fig:real-world}. Due to their low success rates in the analytical environment and the gazebo environment, both the MPC and E2E baselines pose an increased risk of damaging the robot if executed in the real world. We, therefore, refrained from running them in our real-world experiments.

We mark goal poses for the end-effector with AR markers and use the robots' head cameras to detect them. If the marker is initially out of sight, the robot receives an initial guess of the target pose. To easily specify goals, we use a pre-recorded map and Adaptive Monte Carlo Localization (AMCL) for localization. We also give the end-effector planners access to this map, in the form of a static layer in the sensed global costmap. For each episode, we move the robot into a random start pose and draw random initial joint values. We scale the agents' actions by a factor of 0.5 to ensure the safe execution of the motions. For the PR2, we also inflate the local occupancy map by \SI{1}{\centi\meter} for the door opening task and \SI{3}{\centi\meter} for all the other tasks. Results for all the tasks are reported in \tabref{tab:real_world}. The experiments are shown in \figref{fig:real-world} and in the accompanying video.

\begin{table}
    \centering
    \caption{Success rates in the real world experiments. 
    } 
    \begin{threeparttable}
    \begin{tabularx}{\columnwidth}{cl|YY|YYY|Y|Y}
    \toprule
           & & \multicolumn{2}{c|}{$A^*$} & \multicolumn{3}{c|}{Imitation Learning} & \multicolumn{1}{c|}{Spline} & \multicolumn{1}{c}{Total}\\
    \cmidrule{3-9}
      & Metric               & p\&p static & p\&p dynamic & cabinet & drawer & door     & spline \\
    \midrule
        \parbox[t]{0.5mm}{\multirow{6}{*}{\rotatebox[origin=c]{90}{PR2}}} 
        & \textbf{Success}   &  \textbf{12} & \textbf{12} & \textbf{28}  &  \textbf{25} &  \textbf{23} & \textbf{24} & \textbf{124}\\
        & Base coll.     &  ~2      & ~2    & ~0  &  ~0  &  ~0  &  ~0 & \phantom{00}4 \\
        & Arm coll.      &  ~0      & ~0    & ~2  &  ~0  &  ~3  &  ~0 & \phantom{00}5 \\
        & Grasp fail     &  ~1      & ~1    & ~0  &  ~3  &  ~4  &  ~0 & \phantom{00}9 \\
        & IK fail        &  ~0      & ~0    & ~0  &  ~2  &  ~0  &  ~6 & \phantom{00}8 \\
        & Nr. episodes   &  15      & 15    & 30  &  30  &  30  &  30 & 150\\
        \midrule
        \parbox[t]{0.5mm}{\multirow{7}{*}{\rotatebox[origin=c]{90}{HSR}}}
        & \textbf{Success}   & \textbf{11} & \textbf{12} & \textbf{23} & \textbf{24}    &  \textbf{24} & \textbf{16} & \textbf{110}\\
        & Base coll.     & ~0       &  ~2   & ~0   & ~0  &  ~1 & ~0 & \phantom{00}3 \\
        & Arm coll.      & ~1       &  ~0   & ~0   & ~0  &  ~1 & ~0 & \phantom{00}2 \\
        & Grasp fail     & ~0       &  ~0   & ~6   & ~3  &  ~0 & ~0 & \phantom{00}9 \\
        & IK fail        & ~3       &  ~1   & ~1   & ~2  &  ~4 & ~4 & \phantom{0}15 \\
        & Safety stop    & ~0       &  ~0   & ~0   & ~1  &  ~0 & ~0 & \phantom{00}1 \\
        & Nr. episodes   &  15      & 15    & 30   & 30  &  30 & 20 & 140\\
    \bottomrule
    \end{tabularx}
      \begin{tablenotes}[para,flushleft]
       \footnotesize      
       Notes: The PR2 is evaluated in an environment consisting of two rooms connected by a door. The HSR is directly evaluated in our offices. Each cell denotes the number of episodes. The last column sums over all tasks.
     \end{tablenotes}
   \end{threeparttable}
    \label{tab:real_world}
\end{table}

{\parskip=5pt\noindent\textit{Pick\&Place}:} We define positions for several tables, marked orange in \figref{fig:real-world}, as possible pick up and place locations, then draw random pairs of these locations for each episode. We then randomly arrange 2 to 4 obstacles within the map. These obstacles include a large wall segment with a curved star footprint, bins, and a chair for the PR2, and chairs and roll-containers for the HSR. In the second experiment, on top of these static obstacles, we incorporate dynamic obstacles in the form of humans and wheeled objects moved by the authors. For the HSR, we use the \textit{A$^*$-fwd} planner. We also inflate the local map by \SI{1.5}{\centi\meter} and include the map as a static layer in the local map, as the robot does not have any sensors in the back. For the dynamic obstacles, we increase the global map inflation from \SI{0.4}{\meter} to \SI{0.5}{\meter} for both robots to obtain smoother trajectories for the end-effector motion generated by the \textit{A$^*$-}planner.

Both robots achieve high success rates of 76.6\% -- 80\% for both the static and dynamic obstacles. They demonstrate well-planned movements around obstacles and behavior such as backing out of confined starting poses if necessary. When confronted with dynamic obstacles, they react quickly and are able to evade obstacles that move directly into their base without breaking the end-effector motion. Secondly, the learned behavior is robust to frequently changing end-effector motions as the $A^*$-planner adapts to dynamic changes in the shortest path.

{\parskip=5pt\noindent\textit{Spline}:} For the spline interpolation, we draw waypoints from within a single room of the environment. Both robots are able to follow these motions with success rates of 80\%. The main source of failures is too large deviations from the desired end-effector motion when having to follow motions at very large heights (PR2) or unusual orientations (HSR). As an additional difficulty, the HSR's gripper frequently moves so low as to be detected as an obstacle by its base LiDAR. However, the learned policy proved robust to this disturbance.

{\parskip=5pt\noindent\textit{Cabinet and drawer}:} For the imitation system motions, we construct tasks with the same objects as in the simulation. As the imitation learning system does not incorporate obstacle avoidance into the end-effector motions, we restrict obstacles to objects with a low height, such that the motions can pass over them and choose start poses in the same room as the target object. We rearrange two obstacles every 3 to 5 episodes, deliberately placing at least one of them such that it constrains the opening motion. The PR2 achieves very high success rates of 93.3\% and 83.3\% on the cabinet and drawer tasks respectively. The HSR succeeds in 76.6\% and 80\% of all the episodes. Its main failure source is the precision of the grasp, grasping slightly in front of the handle, thereby not opening the object. In one episode, the HSR joints hit a safety limit while being in contact with the drawer which led the controllers to be stopped by the software.

{\parskip=5pt\noindent\textit{Door}:} For the PR2, we use the same door and motions as in the simulation. For the HSR, we use a door within our office, with an even smaller frame width of \SI{0.83}{\meter}. As the door's opening radius differs, the learned imitation motions do not directly apply to it. Instead, we sample eight waypoints and interpolate them with the spline interpolator to construct an opening motion for the end-effector. As its arm is not strong enough to press down the handle, we do not lock the door's spring for the HSR. The PR2 succeeds in 76.6\% of all episodes and performs better than in Gazebo. In general, the real hardware seems more reactive than in the simulator. The HSR achieves success rates of 80\%. Failure cases for the PR2 include collisions of the elbow with the door itself, slipping off the handle, and not being able to unlock the door lock. The HSR in a few cases maneuvers to the right of the handle, such that it is not able to pass through the door frame without moving the non-continuous arm roll joints into another configuration (which would require violating the motion constraints). In these situations, it prefers to stay in place until it terminates because the deviations to the desired motion become too large, rather than to produce unsafe behavior such as collisions. The HSR agent proved robust to weird and irregular occupancy maps produced by the glass door.

Across all the tasks, we observe that the learned behaviors directly transfer to the real world with average success rates of 82.6\% and 78.6\% on the PR2 and HSR respectively. The agents learned to efficiently avoid obstacles and cause very few base collisions. They rather fail the arm motions than drive into an obstacle. The produced motions are robust to noise such as localization error, which could be significant whenever the number of unmapped obstacles are large. Although the overall motions are smooth, at higher speeds the arm motions can sometimes become a bit shaky. This has several reasons: the low-level base and arm controllers are independent and as such cannot take into account each other's errors. Secondly, the hardware introduces several additional noise sources that are not present in the simulation. These include asynchronous control, localization errors, calibration errors, unseen obstacle geometries, and artifacts in the occupancy maps. We found that the acceleration regularization introduced in \secref{sec:approach} is essential for smoothness. Finetuning on the real system is a promising avenue to further increase the precision and the feasible velocities on the real systems. The HSR's main failure source are the IK failures. A number of these occurred due to conflicts between the head and arm. Unlike in training in the real world, the head moved to focus the camera on the target object. Thereby leading to different self-collision constraints.

\setlength{\tabcolsep}{1pt}
\renewcommand{\arraystretch}{1}
\begin{figure*}
	\centering
    \footnotesize
    \setlength{\tabcolsep}{0.05cm}
    {\renewcommand{\arraystretch}{1}
    \begin{tabular}{cc}
        {(a)} & {(b)} \\
  		\includegraphics[width=0.4\textwidth,trim={0.0cm 0.0cm 0.0cm 0.0cm},clip,angle=0]{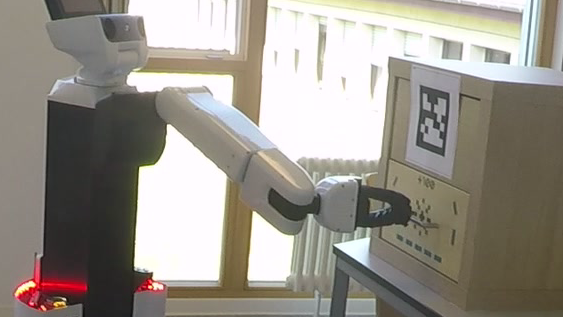} & \includegraphics[width=0.4\textwidth,trim={0.0cm 0.0cm 0.0cm 0.0cm},clip,angle=0]{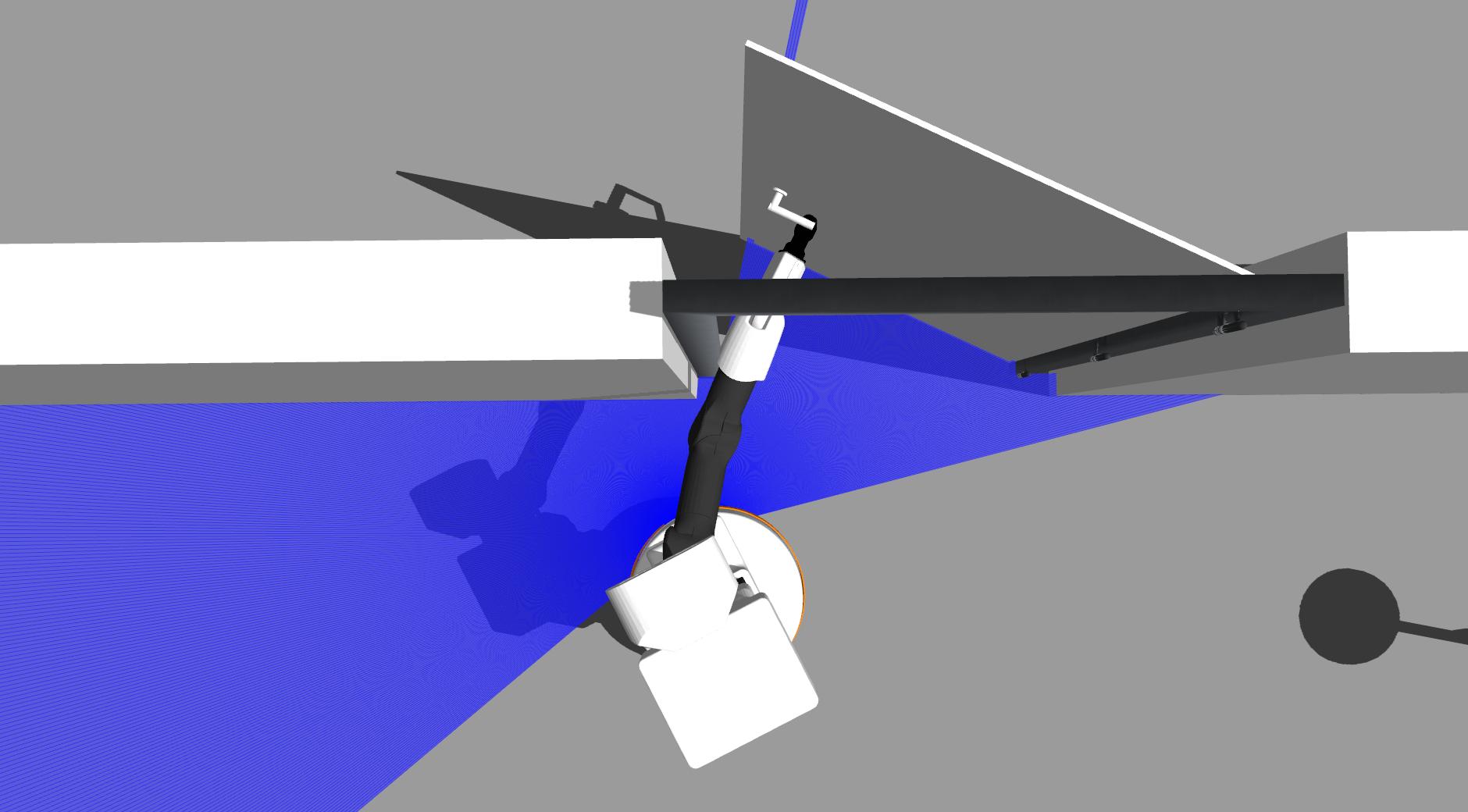} \\ \includegraphics[width=0.4\textwidth,trim={0.0cm 0.0cm 0.0cm 0.0cm},clip,angle=0]{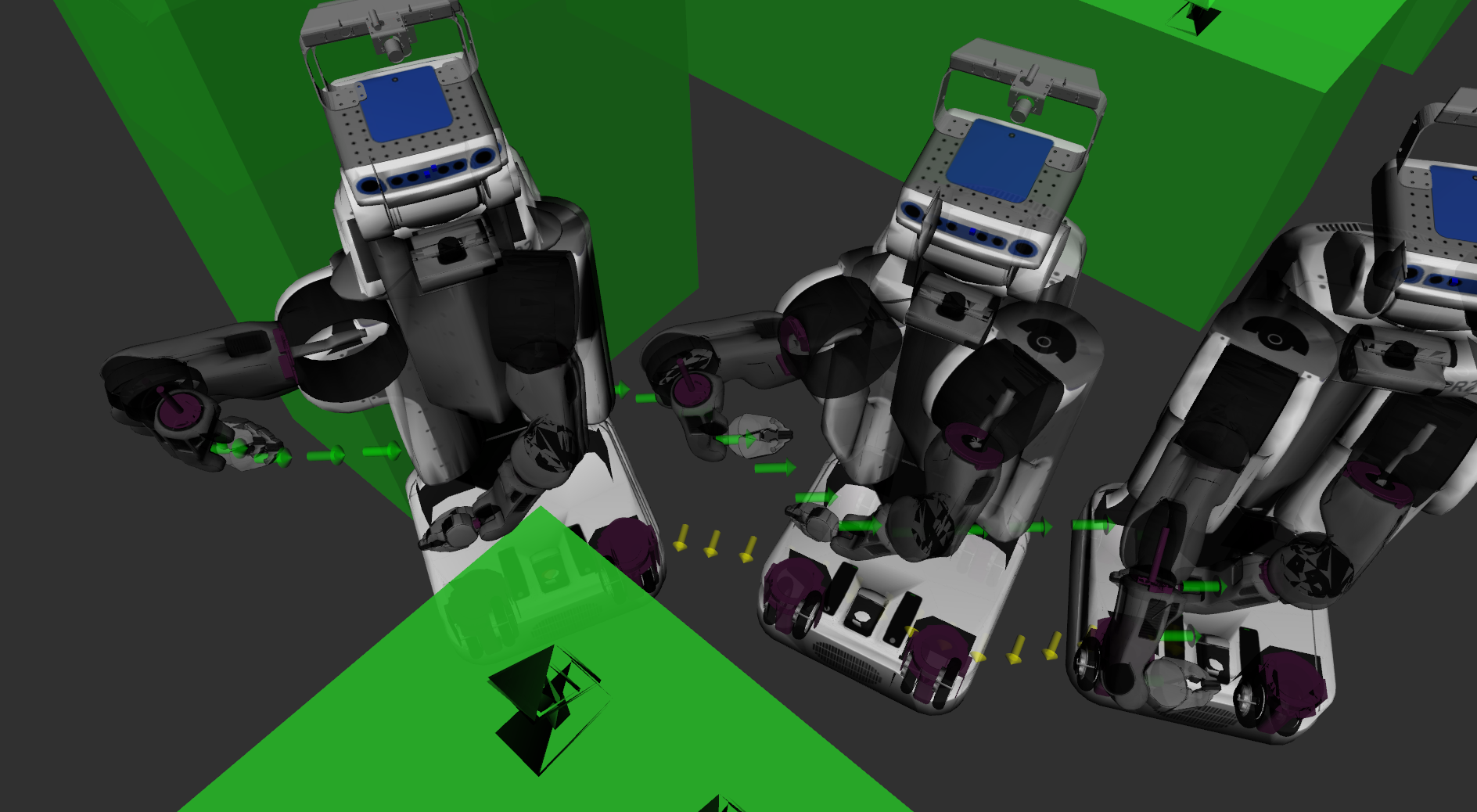} &
        \includegraphics[width=0.4\textwidth,trim={0.0cm 0.0cm 0.0cm 0.0cm},clip,angle=0]{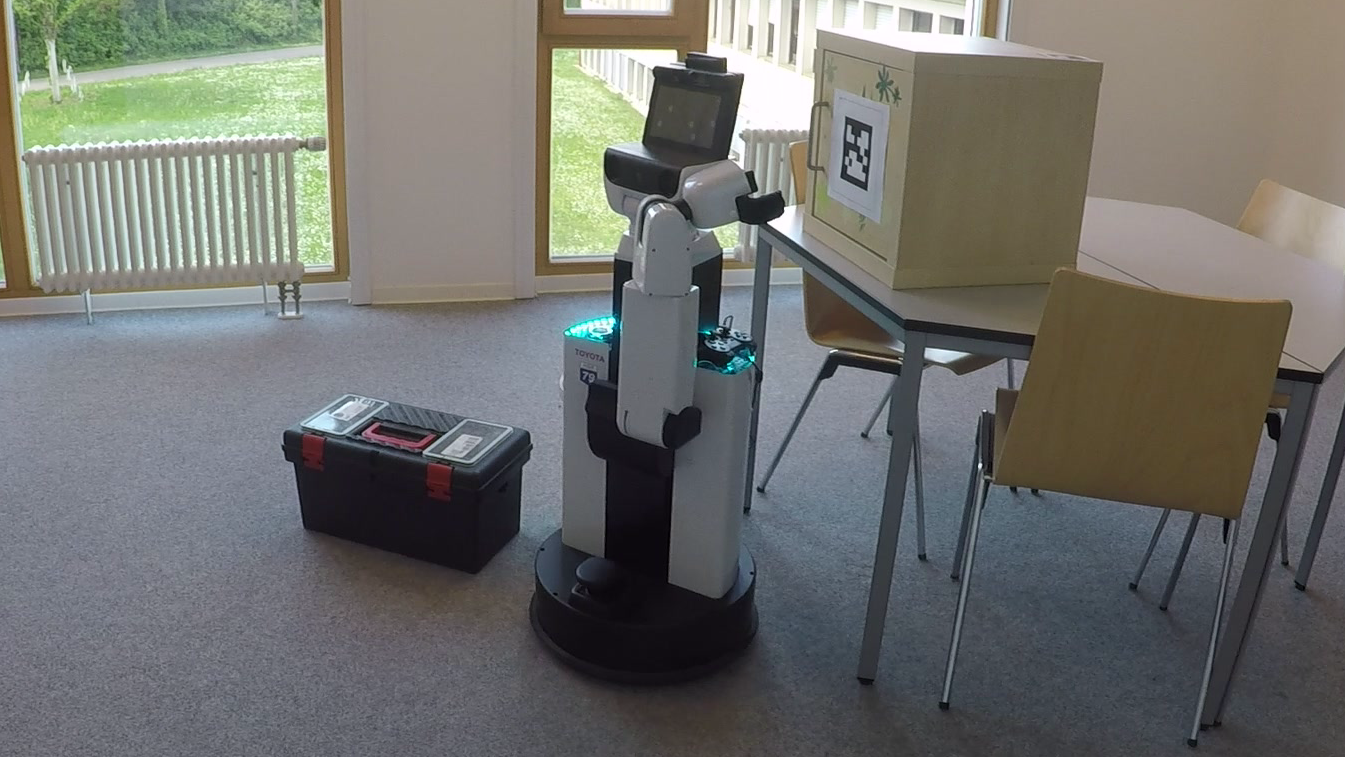} \\
        {(c)} & {(d)}
    \end{tabular}}
	\caption{\myworries{Examples of failure cases. (a) The independence between base and arm controllers together with noise in localization and actuation's can impact the end-effector precision which in some cases leads to grasping failures. (b) Collision avoidance focuses on the end-effector and the robot base but does not take into account collisions of the robot arm. Here the TIAGo's wrist collides with the door frame. (c) Heuristics to generate pick\&place motions can produce overly restrictive motions. \myworriestwo{In the situation shown here, the interpolation results in a desired end-effector pose that is oriented in the opposite of the driving direction (pointing to the right while the path to the goal is further to the left). This leads} the base to collide with an obstacle at the left while trying to adhere to this orientation. (d) The agent does not consider head configurations outside of the training distribution. As the HSR's head turns to refocus the cabinet, this leads to a self-collision with the arm.}}
  	\label{fig:failures}
\end{figure*}
\setlength{\tabcolsep}{6pt}
\renewcommand{\arraystretch}{1}
\begin{figure}
    \footnotesize
	\centering
  		\includegraphics[width=1.0\columnwidth,trim={0.0cm 0.0cm 0.0cm 0.0cm},clip,angle =0]{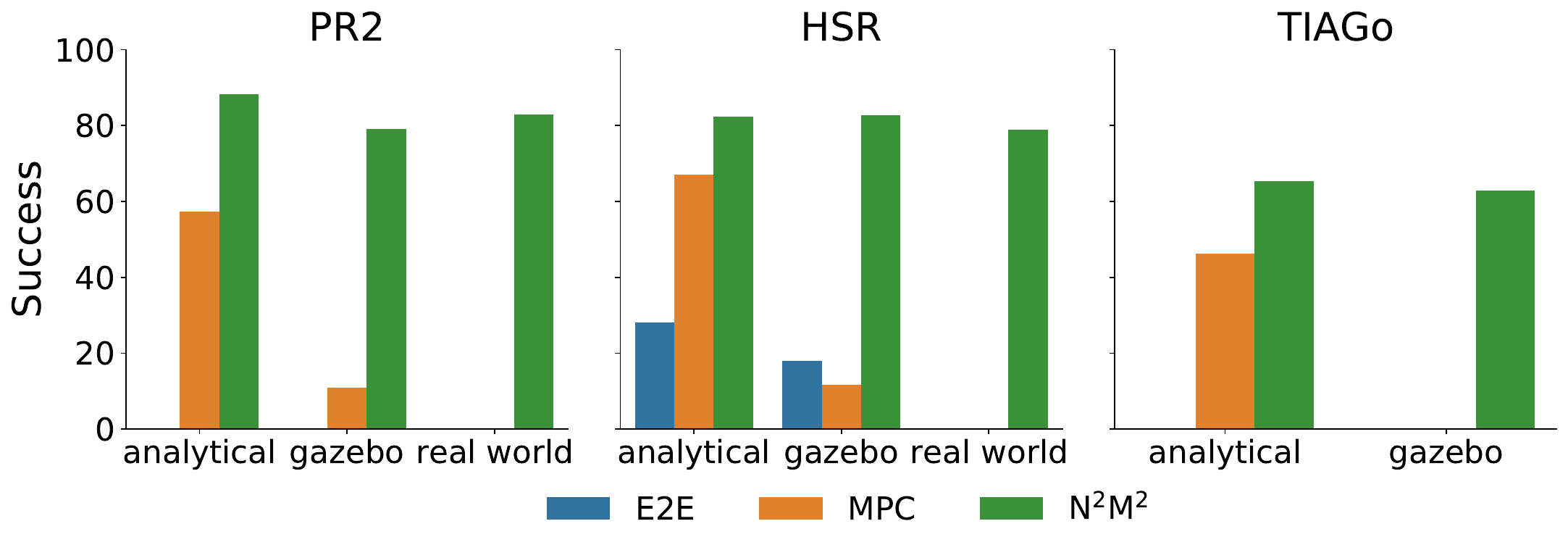}
  		\vspace{-0.5cm}
	\caption{Average success rates across different robots and environments. Given the low success rates in the analytical environment and the simulated gazebo environment leading to an increased risk of damaging the robot, we did not evaluate the MPC and E2E baselines in the real world.}
  	\label{fig:avg_success}
\end{figure}

\myworries{\subsection{Limitations}\label{sec:limitations}
While the approach achieves high success rates across all tasks and environments, we identify a number of failure cases and limitations to address in the future. \tabref{tab:real_world} provides a quantitative decomposition of the failures in the real world. \figref{fig:failures} shows examples of failure cases.\\
{\noindent\textit{High-precision navigation:}} While the approach is able to consistently navigate real-world maps, centimeter-exact navigation such as in the narrow door frame and passages of the bookstore map can still be a challenge. We identify two causes for this: first, as we perform zero-shot transfer to the real world, the agent is unaware of noise in components such as the low-level controllers or localization. Finetuning in the real world should enable the agent to better take these uncertainties into account. Secondly, the resolution of the obstacle map is \SI{2.5}{\cm}, which leaves the agent to perceive up to \SI{1.25}{\cm} of free space on each side as occupied. For high-precision tasks, the agent may benefit from further increasing this resolution.\\
{\noindent\textit{Independent controllers:}} We rely on out-of-the-box low-level controllers of the different robotic platforms. At the moment neither of these offers whole-body controllers. Instead, we use the independent base and arm controllers. These cannot take into account each other's error terms, sometimes leading to deviations from the desired end-effector pose or grasp failures e.g. when the base rotates while the arm is moving (\figref{fig:failures}~(a)). The development of joint low-level controllers is a promising avenue to further stabilize and increase the precision of the end-effector motions.\\
{\noindent\textit{Arm collisions:}} We present an efficient and well-generalizing obstacle avoidance in which the end-effector motion generator is responsible for collision-free motions for the end-effector, while the RL agent avoids base collisions. However, in certain scenarios, it is still possible for the arm to collide with objects in the environment, as the RL agent does not take into account 3D obstacle avoidance (\figref{fig:failures}~(b)). We plan to investigate full collision avoidance for the arm using RGB-D data or voxel maps in future work.\\
{\noindent\textit{Overly restrictive end-effector motions:}} Certain tasks such as pick\&place do not require the robot to follow exact end-effector positions or orientations for large parts of the task. The motions we use in this work use effective heuristics and interpolation. While this is simple and efficient for many cases, it can lead to overly difficult end-effector motions for example in the narrow passages of the bookstore map (\figref{fig:failures}~(c)). In the future, this may be addressed by developing methods that allow the agent to deviate further from the end-effector motions where unproblematic or by learning end-effector motions.\\
{\noindent\textit{Head collisions:}} In the real world we control the robots' head cameras to focus on the target objects. This moves the head into unseen configurations that the agent is unaware of. In the case of the HSR, this causes a number of self-collisions with the arm (\figref{fig:failures}~(d)). Conditioning on the full robot state and randomizing the head configurations during training is a promising avenue to alleviate this problem.\\
\subsection{Generality and Applicability}
{\noindent\textit{Solvable tasks:}} 
Our method is in principle able to provide kinematically feasible navigation for any task for which a motion generator for the end-effector is available. 
This applies to a wide variety of mobile manipulation tasks, as we demonstrate with a large range of such tasks in our experiments. This modularity enables us to reuse knowledge and generalize to unseen tasks, in contrast to previous learning based methods that have to retrain from scratch on each new task. \myworriestwo{However,} certain tasks can currently not be solved in this manner, for example tasks that require specific joint configurations, such as pushing open a door with the elbow joint.\\
{\noindent\textit{Robotic platforms:}} The approach is directly applicable to any mobile robotic system that can be decomposed into base and manipulator. In our experiments we demonstrate this on holonomic and non-holonomic wheeled drives, without any platform-specific hyper-parameter tuning. But the same formulation could also be applied to different morphologies such as quadrupeds. Main requirements are that the manipulator is not needed in the navigation motions and the availability of a low-level controller interface to translate velocity or position commands from the agent to the actuators. 
We provide pre-trained agents for the PR2, HSR and the TIAGo on the project website\footnote{\url{http://mobile-rl.cs.uni-freiburg.de}}. We also provide the methods to train agents for other robots, requiring only an implementation of the corresponding environment.
}

\section{Conclusion}\label{sec:conclusion}
We introduced \ours, which extends the formulation of kinematic feasibility for mobile manipulation to complex unstructured environments. We generalized its objective function and extended the agent's control to the velocity of the end-effector motions and prevent configuration jumps by learning the torso joints and introducing a regularization to the inverse kinematics. We then introduced a procedurally generated training environment that uses strong randomization and simple elements to produce diverse scenarios. The result is a powerful approach that can successfully act in unseen human-centered real-world environments. In extensive experiments across a variety of robots, physics, and environments, we demonstrated that this approach successfully generalizes in a zero-shot manner to novel tasks, unseen objects and geometries, and dynamic obstacles. By leveraging a hybrid combination of inverse kinematics and reinforcement learning, the agent solves tasks with a vast continuous configuration space in which previous state-of-the-art approaches struggle. Our method outperforms these approaches both in the analytical environment and in the transfer to unseen environments on all robotic platforms. \figref{fig:avg_success} summarizes the success rates across the different robots and environments.

In this work, we have focused on achieving arbitrary motions and used simple robot-agnostic end-effector motion generators. We purposefully abstracted from optimizing the motions or goals themselves to demonstrate the system's capabilities to achieve challenging motions. In the future, we plan to jointly or iteratively optimize the robot's behaviors and the generated end-effector motions. A particularly exciting direction is to incorporate this work into hierarchical approaches that learn to produce motions or subgoals for the end-effector to achieve high-level goals. Such an approach will benefit from the ability to abstract from complex base behavior to reason in a much simpler space of end-effector motions. This can be done both within a learning-based paradigm or on the motion planning level of task-and-motion planning-based pipelines. 

Further work includes the incorporation of partially observable and 3D obstacle avoidance for the robot arm. The flexibility of the RL approach means that this can be incorporated based on voxel maps or directly learned from camera inputs. The development of joint low-level controllers and finetuning of the learned policies in the real world are promising avenues for further improvements in precision and velocities. Lastly, we find that the current reinforcement learning methods still face difficulties to explore certain high-dimensional continuous action spaces, as exhibited by the TIAGo robot. Methods to alleviate this problem will be important for robotics. A particularly interesting direction for mobile manipulation is the combination of value learning methods with efficient Monte-Carlo rollouts, combining the best of MPC and learning-based approaches.

\section*{Acknowledgements}
We thank Adrian R\"ofer and Niklas Wetzel for their feedback on the initial draft of the paper.

\ifCLASSOPTIONcaptionsoff
  \newpage
\fi


\bibliographystyle{IEEEtran}
\bibliography{references.bib}

\begin{IEEEbiography}[{\includegraphics[width=1in,height=1.25in,clip,keepaspectratio]{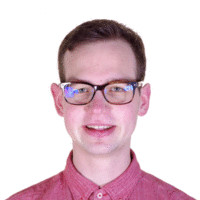}}]{Daniel Honerkamp} is a Ph.D.~student in the Robot Learning Lab headed by Abhinav Valada. He received his M.S.~degree in Computational Statistics and Machine Learning from University College London in 2018. His research focuses on Reinforcement Learning and Autonomous Decision Making for Mobile Manipulation and Embodied Agents.
\end{IEEEbiography}

\begin{IEEEbiography}[{\includegraphics[width=1in,height=1.25in,clip,keepaspectratio]{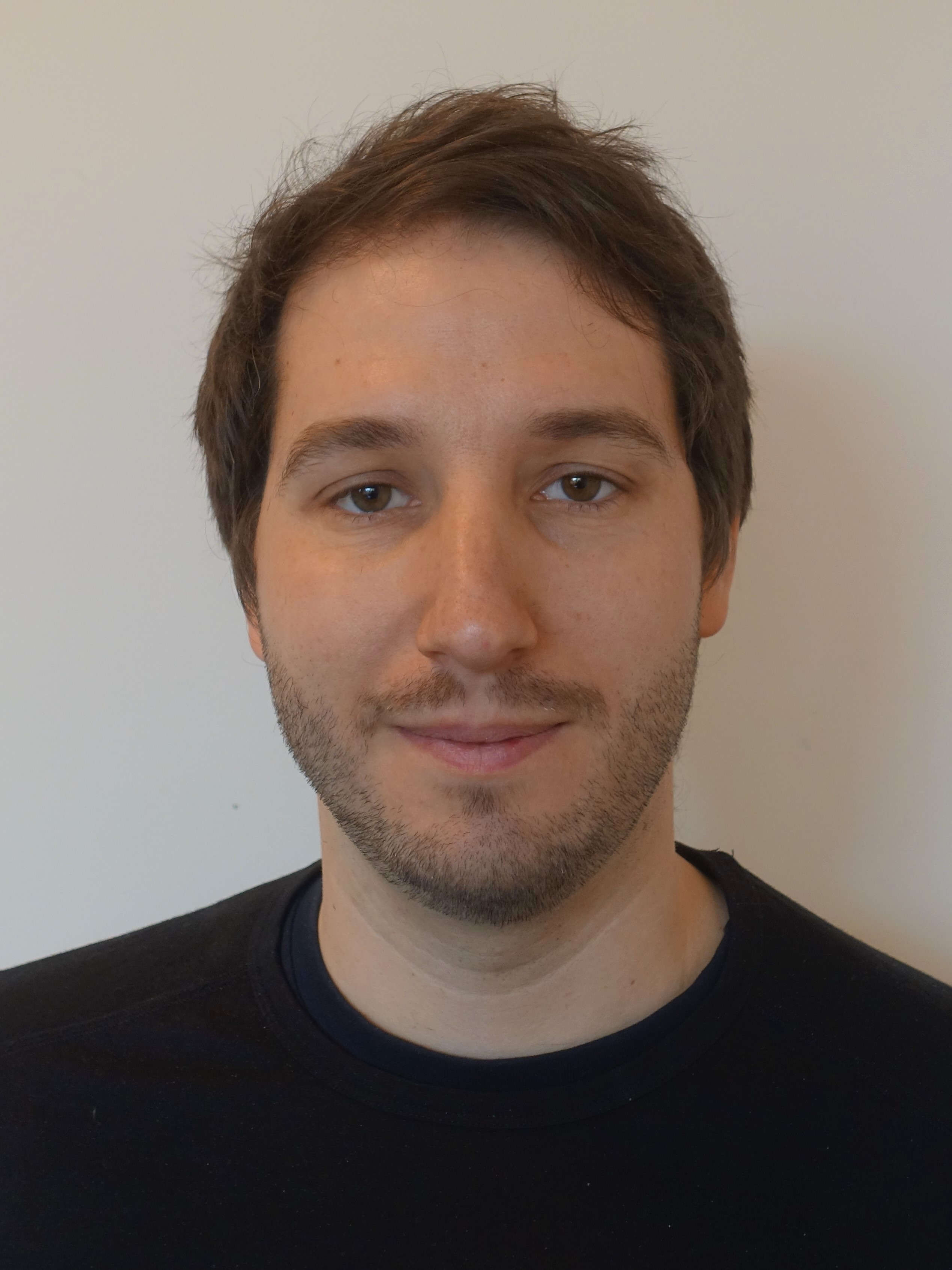}}]{Tim Welschehold} is a postdoctoral researcher in the Autonomous Intelligent Systems group headed by Wolfram Burgard. He received his Ph.D.~degree in Computer Science from the University of Freiburg in 2020 and his Diploma in Physics in 2013. He is a member of the BrainLinks-BrainTools Center. His research revolves around Reinforcement Learning, Imitation Learning and Representation Learning in robotics with a focus on Mobile Manipulation and Dynamical Systems.
\end{IEEEbiography}

\begin{IEEEbiography}[{\includegraphics[width=1in,height=1.25in,clip,keepaspectratio]{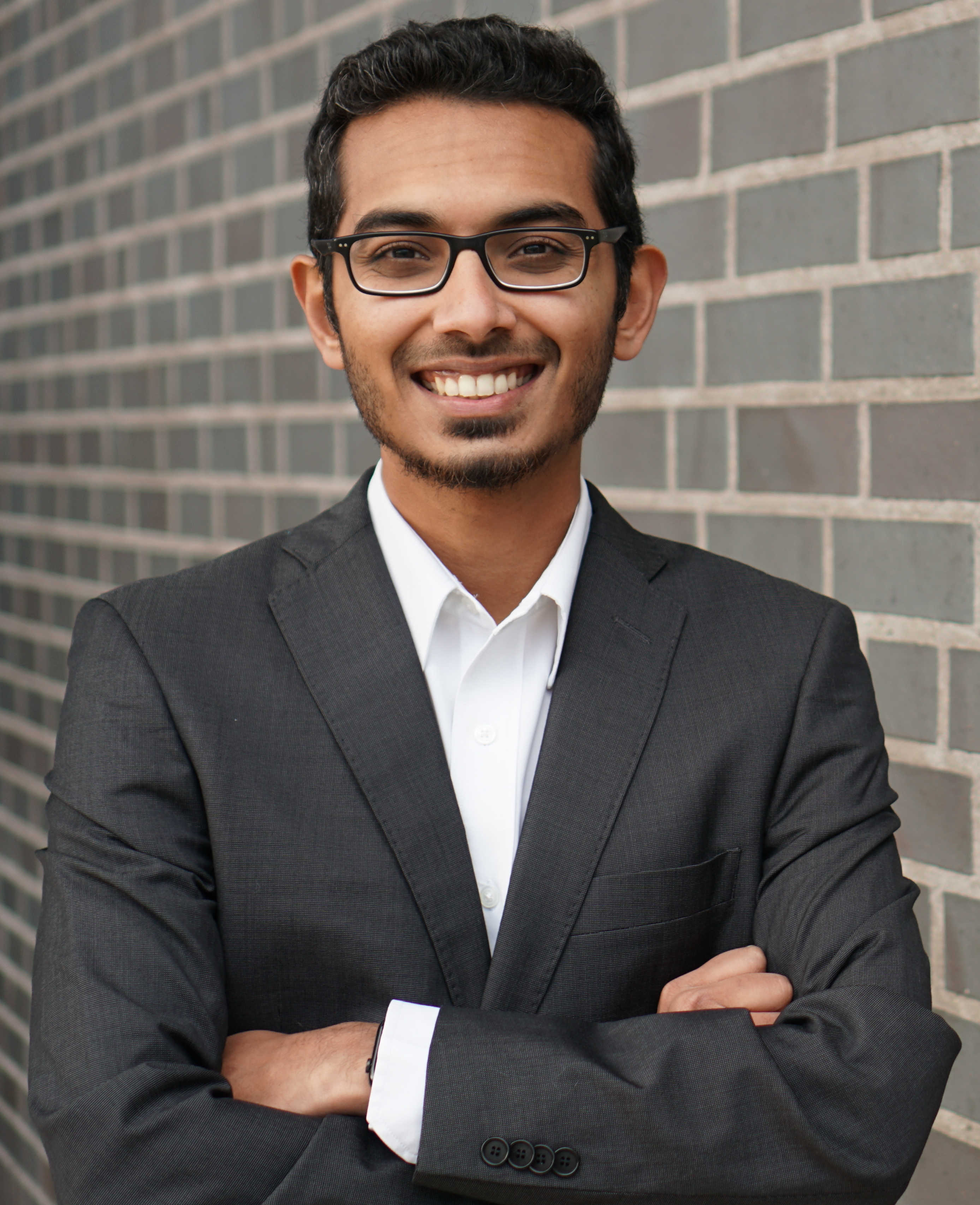}}]{Abhinav Valada}
is an Assistant Professor and Director of the Robot Learning Lab at the University of Freiburg. He is a member of the Department of Computer Science, a principal investigator at the BrainLinks-BrainTools Center, and a core faculty in the European Laboratory for Learning and Intelligent Systems (ELLIS) unit in Freiburg. He received his Ph.D.~in Computer Science from the University of Freiburg and his M.S.~degree in Robotics from Carnegie Mellon University. His research lies at the intersection of robotics, machine learning, and computer vision with a focus on tackling fundamental robot perception, state estimation, and control problems using learning approaches in order to enable robots to reliably operate in complex and diverse domains.
\end{IEEEbiography}




\end{document}